\documentclass[sigconf]{aamas}

\usepackage{balance} 
\usepackage{algorithm,algpseudocode}
\usepackage{subcaption}   
\usepackage{xcolor}

\usepackage{hyperref}

\makeatletter
\gdef\@copyrightpermission{
  \begin{minipage}{0.2\columnwidth}
   \href{https://creativecommons.org/licenses/by/4.0/}{\includegraphics[width=0.90\textwidth]{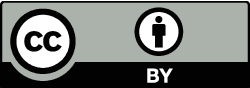}}
  \end{minipage}\hfill
  \begin{minipage}{0.8\columnwidth}
   \href{https://creativecommons.org/licenses/by/4.0/}{This work is licensed under a Creative Commons Attribution International 4.0 License.}
  \end{minipage}
  \vspace{5pt}
}
\makeatother

\setcopyright{ifaamas}
\acmConference[AAMAS '25]{Proc.\@ of the 24th International Conference
on Autonomous Agents and Multiagent Systems (AAMAS 2025)}{May 19 -- 23, 2025}
{Detroit, Michigan, USA}{Y.~Vorobeychik, S.~Das, A.~Nowé  (eds.)}
\copyrightyear{2025}
\acmYear{2025}
\acmDOI{}
\acmPrice{}
\acmISBN{}

\acmSubmissionID{520}

\title{\centering Boosting Robustness in Preference-Based Reinforcement Learning with Dynamic Sparsity}

\subtitle{Extended Abstract}

\author{Calarina Muslimani}
\affiliation{
  \institution{University of Alberta}
  \country{Canada}}
\email{musliman@ualberta.ca}

\author{Bram Grooten}
\affiliation{
  \institution{Eindhoven University of Technology}
  \country{Netherlands}}
\email{b.j.grooten@tue.nl}
\author{Deepak R.S. Mamillapalli}
\affiliation{
  \institution{University of Alberta}
  \country{Canada}}
\email{mamillap@ualberta.ca}
\author{Mykola Pechenizkiy}
\affiliation{
  \institution{Eindhoven University of Technology }
  \country{Netherlands}}
\email{m.pechenizkiy@tue.nl}

\author{Decebal C. Mocanu}
\affiliation{
  \institution{University of Luxembourg}
  \country{Luxembourg}}
\email{decebal.mocanu@uni.lu}
\author{Matthew E. Taylor}
\affiliation{
  \institution{University of Alberta}
  \institution{Alberta Machine Intelligence Institute}
  \country{Canada}}
\email{matthew.e.taylor@ualberta.ca}

\begin{abstract}
To integrate into human-centered environments, autonomous agents must learn from and adapt to humans in their native settings. Preference-based reinforcement learning (PbRL) can enable this by learning reward functions from human preferences.
However, humans live in a world full of diverse information, most of which is irrelevant to completing any particular task. 
It then becomes essential that agents learn to focus on the subset of task-relevant state features.
To that end, this work proposes R2N (Robust-to-Noise), the first PbRL algorithm that leverages principles of dynamic sparse training to learn robust reward models that can focus on task-relevant features.
In experiments with a simulated teacher, we demonstrate that R2N can adapt the sparse connectivity of its neural networks to focus on task-relevant features, enabling R2N to significantly outperform several sparse training and PbRL algorithms across simulated robotic environments. We open-source our code at the following link: \href{https://github.com/cmuslima/R2N}{\url{https://github.com/cmuslima/R2N}}.
\end{abstract}


\newcommand{\BibTeX}{\rm B\kern-.05em{\sc i\kern-.025em b}\kern-.08em\TeX}

\begin{document}


\pagestyle{fancy}
\fancyhead{}


\maketitle 


\section{Introduction}
\label{sec:introduction}

Recent advances in reinforcement learning (RL) are bringing us closer to a future where RL agents aid humans in their daily lives \citep{Dwivedi_2021, chatgpt4, geminiteam2024gemini}. 
Preference-based RL (PbRL) is a promising paradigm that allows RL agents to leverage human preferences to adapt their behavior to better align with human intentions \citep{christiano2017deep, pebble}. However, to effectively integrate agents into human-centered environments, autonomous agents should be able to learn from humans in their natural settings.
Unfortunately, human environments are inherently noisy. 
For example, a household robot should be able to effectively learn from human preferences despite receiving a continuous stream of information regarding various household chores. 
Suppose a household robot is tasked with learning to clean a toy room, and a human provides the robot with preferences on how the room should be cleaned. In this scenario, the robot might receive distracting information such as sounds from children playing, colors and shapes of various toys, room temperature, etc. Only a certain subset of the robot's perceptions is relevant to the task. Identifying this subset of task-relevant features can boost performance and sample efficiency.
However, there has been little attention in PbRL on how to learn in such noisy environments. 
More often, research in PbRL focuses on hand-engineered environments that contain only task-relevant features \citep{pebble, surf, hejna2023few, rune, sdp}. There has been little attention in PbRL on how to learn in noisy environments where identifying all relevant features can be difficult and time-consuming. This lack of attention is problematic as we consider more real-world settings, like the household robot example. In such settings,  recent algorithmic improvements for PbRL may not be helpful in environments containing irrelevant features. This paper asks the question:
\begin{center}
How can autonomous agents learn from human preferences in \textit{noisy environments} with many irrelevant features?
\end{center}

To study this, we consider the Extremely Noisy Environment (ENE) problem setting \citep{anf}. Given a traditional RL environment, an ENE expands the size of the state space by adding features of random noise sampled from $\mathcal{N}(0,1)$. These new features are task-irrelevant and the agent does not have any information on the utility of each feature. Learning becomes quite difficult without knowing which features are task-relevant and which are noise. 
While previous work in dynamic sparse training (DST) has shown promising results in filtering out irrelevant features in reinforcement learning \citep{set, rigl, ghada, anf}, its application to preference-based reinforcement learning remains unexplored. 
To that end, we present R2N, a \emph{robust-to-noise} PbRL algorithm that leverages principles of dynamic sparse training to learn a robust reward model, effectively filtering out irrelevant features while reinforcing neural network connections to pertinent ones. Importantly, this filtering and policy learning is done solely from preferences and not from a ground truth reward function or other background knowledge.

This paper's core contributions are as follows:
\begin{itemize}
    \item We are the first to consider the Extremely Noisy Environment setting for preference-based reinforcement learning. 
    \item We propose \textit{R2N}, a noise-robust PbRL algorithm that enables its learned reward model and RL agent to focus on relevant environmental features through dynamic sparse training. 
    \item We demonstrate that R2N can maintain competitive performance, but often outperform four sparse training algorithms in five DeepMind Control environments \citep{dm_control} with high added noise.
    \item We integrate R2N with three state-of-the-art PbRL algorithms, leading to significant performance improvements. This demonstrates the versatility of R2N and its ability to enhance multiple PbRL methods.

\end{itemize}

The primary goal of this paper is to highlight the importance of continued research in PbRL for noisy environments, as it is necessary for PbRL to be effective in real-world settings. We demonstrate the potential of dynamic sparse training as a promising avenue towards achieving this goal.

\section{Related work}

\paragraph{Human-in-the-Loop RL} Human-in-the-loop RL consists of a growing set of methods that allow an RL agent to leverage human feedback to improve its behavior. Various types of human feedback have been considered, including demonstrations, action advice, scalar feedback, and preferences. By learning from demonstrations, teachers provide $\langle$state, action$\rangle$ trajectories of the desired agent behavior \citep{abbeel2004apprenticeship, argall2009survey,brys2015reinforcement}. Similarly, in the action advising setting, a teacher suggests actions for the RL agent to take \citep{torrey2013teaching,frazier2019improving}. 
As an alternative, other approaches consider scalar and preference-based feedback. In learning from scalar feedback, teachers provide a scalar rating of an agent's behavior \citep{knox2009interactively,griffith2013policy,loftin2016learning,macglashan2017interactive,deeptamer, rating_based_RL}. In PbRL, human feedback is even simpler, only requiring a preference between two recorded segments of the agent's behavior. The preferences are often used to learn an estimate of the true reward function, which the RL agent then maximizes \citep{christiano2017deep, pebble}. To reduce the number of human preferences necessary, works have considered a variety of techniques, including data augmentation \citep{surf}, uncertainty-based exploration \citep{rune}, meta-learning \citep{hejna2023few}, semi-supervised learning \citep{surf}, and pre-training with sub-optimal data \citep{sdp}. Other work has focused on leveraging preferences without explicitly modeling a reward function \citep{hejna2023inverse, dpo}. 
PbRL has been popularized in recent years, particularly due to its success in improving large language models \citep{chatgpt4, ziegler2020finetuning}.

\paragraph{Sparsity in Neural Networks} 

Sparsity provides a means to reduce the parameter count of neural networks without decreasing performance or representational power \citep{set, rigl, bellec2018deep, dettmers2019sparse,  yuan2021mest, wortsman2019discovering, atashgahi2022brain}. Dynamic sparse training is a subfield of sparse training, with methods such as SET \citep{set} and RigL \citep{rigl}, that start from a random sparse neural network and improve its topology over time.
In both algorithms, the neural network is randomly pruned at initialization up to a certain sparsity level $s \in (0,1)$. 
During training, the network is periodically updated by pruning and growing connections in each sparse layer. More specifically, both SET and RigL drop a fraction of connections with the lowest weight magnitude. However, to grow new weights, SET selects random locations, whereas RigL grows new weights in locations where the gradient magnitude is the highest.
This means RigL needs to compute the gradient of all weights (including the inactive ones) during a sparse topology update. 
RigL has been shown to outperform SET in supervised learning tasks \citep{rigl}, but in RL, there seems to be no significant difference \cite{graesser2022state}.
The fact that DST methods are dynamic (i.e., able to update the network's topology over time) is crucial. 
In reinforcement learning \citet{ghada} applied DST, successfully improving performance using only \textasciitilde$50\%$ of the weights.  
In noisy RL environments, \citet{anf} further showed that specifically sparsifying the input layer improves the robustness of RL algorithms. \citet{graesser2022state} also provided a large overview of the state of sparse training in a diverse set of RL environments, algorithms, and sparsity levels.

Other forms of dynamic sparse training can include DropConnect \cite{dropconnect} and L1 regularization \cite{l1reg}. 
DropConnect introduces dynamic sparsity into a model by randomly setting a subset of weights to zero during each forward pass. In contrast, 
L1 regularization adds a term to the model's loss function that is proportional to the sum of the absolute value of the model's weights. This induces sparsity by driving the weights of less important features to zero, effectively performing feature selection.
Unlike DST approaches, \emph{static} sparse training prunes a set of weights at initialization to a fixed sparsity level, and this sparsity pattern remains fixed throughout training. However, static sparse training has generally been found to be less effective \cite{set, graesser2022state}.

\section{Background}
\label{sec:background}
In reinforcement learning, an RL agent interacts with an environment to maximize the expected cumulative (discounted) reward. This interaction process is modeled as a Markov Decision Process (MDP) consisting of $\langle \mathcal{S}, \mathcal{A}, T, r, \gamma \rangle$. At each interaction step $t$, the agent receives a state $s_t \in \mathcal{S}$ and takes an action $a_t \in \mathcal{A}$ according to its policy $\pi(s|a)$.
The environment then provides a reward $r_{t+1} = r(s_{t},a_{t})$ and transitions to the next state $s_{t+1}$ according to the transitions dynamics $T(s_{t+1}|s_{t},a_{t})$. The agent attempts to learn a policy that maximizes the discounted return $G= \sum_{k=0}^{\infty}\gamma^{k}r_{t+k+1}$.

This work assumes an MDP$\setminus$R setting, where access to the environmental reward function is not provided. The goal is to learn a good policy while simultaneously learning a proper estimate of the reward function from human preferences. 
\paragraph{Preference-based Reinforcement Learning}

PbRL considers trajectory segments $\sigma$, where each segment consists of a sequence of states and actions
\{$s_t, a_t, s_{t+1}, a_{t+1},..., s_{t+k}, a_{t+k}$\}, where $k$ is the trajectory segment length. 
 Two segments, $\sigma^0$ and $\sigma^1$, are then compared by a teacher. If the teacher prefers segment $\sigma^1$ over segment $\sigma^0$, then the target label $y=1$, and if the converse is true, $y=0$. If both segments are equally preferred, then $y=0.5$.
As feedback is provided, it is stored as tuples ($\sigma^0, \sigma^1, y$) in a dataset $D$. Then, we follow the Bradley-Terry model \citep{bradley_terry_model} to define a preference predictor $P_{\theta}$ using the reward function estimator $\hat{r}_\theta$:
\begin{equation}\label{eq:bradley_terry}
P_{\theta}(\sigma^1 > \sigma^0) = \frac{\text{exp}\left(\sum_{t}\hat{r}_\theta(s^{1}_{t}, a^{1}_{t})\right)}{\sum_{i \in \{0,1\}}\text{exp}\left(\sum_{t}\hat{r}_\theta(s^{i}_{t}, a^{i}_{t})\right)}
\end{equation}

Intuitively, if segment $\sigma^i$ is preferred over segment $\sigma^j$ in Equation \eqref{eq:bradley_terry}, then the cumulative predicted reward for $\sigma^i$ should be greater than for $\sigma^j$. 
To train the reward function, we can use supervised learning where the teacher provides the labels $y$. We can then update $\hat{r}_\theta$ through $P_{\theta}$ by minimizing the standard binary cross-entropy objective. 

\begin{multline}\label{eq:cross_entropy}
L^{CE}(\theta, D) = -\ E_{(\sigma^0,\sigma^1, y)\sim D} \bigl[(1-y)\:\text{log}P_{\theta}(\sigma^0 > \sigma^1)\\ 
+\; y\:\text{log}P_{\theta}(\sigma^1 > \sigma^0) \bigr]
\end{multline}

In Equation \eqref{eq:cross_entropy}, the loss increases as the predicted probability that $\sigma^1 > \sigma^0 $ diverges from the true label (e.g., $y=1$). This loss drives the reward model to update its weights to output a greater predicted total reward for $\sigma^1$ than for $\sigma^0$. The learned reward function, $\hat{r}_\theta$, is then used in place of the environmental reward function in the typical reinforcement learning interaction loop.

\paragraph{Noise in Human-in-the-loop RL}

RL agents can experience different types of noise while interacting and learning from humans: measurement error and distracting features. 
Measurement error results from uncertainty in perception. For example, if a human is providing feedback to an RL agent, they might be unsure or indecisive about what feedback to provide. In the preference learning literature, some work has studied this by incorporating ``imperfect’’ simulated teachers that provide random preference orderings to some percentage of their total preference queries \citep{bpref}.  However, this type of noise is outside the scope of this work. 
\paragraph{Problem Setting}
We study the setting in which noise is classified as a distracting feature in an environment. For example, consider the household cleaning robot that receives preferences on cleaning styles. 
 In this task, the robot can receive excess information about the house that is not necessary for the cleaning task. Therefore, the robot needs to learn to filter through the noise to focus solely on task-relevant features.
We consider the Extremely Noisy Environment (ENE) \citep{anf} to study this setting. In an ENE, the state space of a regular RL environment is increased by concatenating a large number of irrelevant features. More specifically, an ENE enlarges the state space such that a certain fraction $n_{f} \in [0,1)$ of the total state space is random noise.
These irrelevant features are produced by sampling i.i.d. from $\mathcal{N}(0,1)$. We use this setting in our experiments unless stated otherwise. 
The PbRL algorithms must identify the most relevant features to (1) learn a robust reward function and (2) learn an adequate policy.

\begin{figure*}[t]
  \centering
  \includegraphics[scale=0.24]{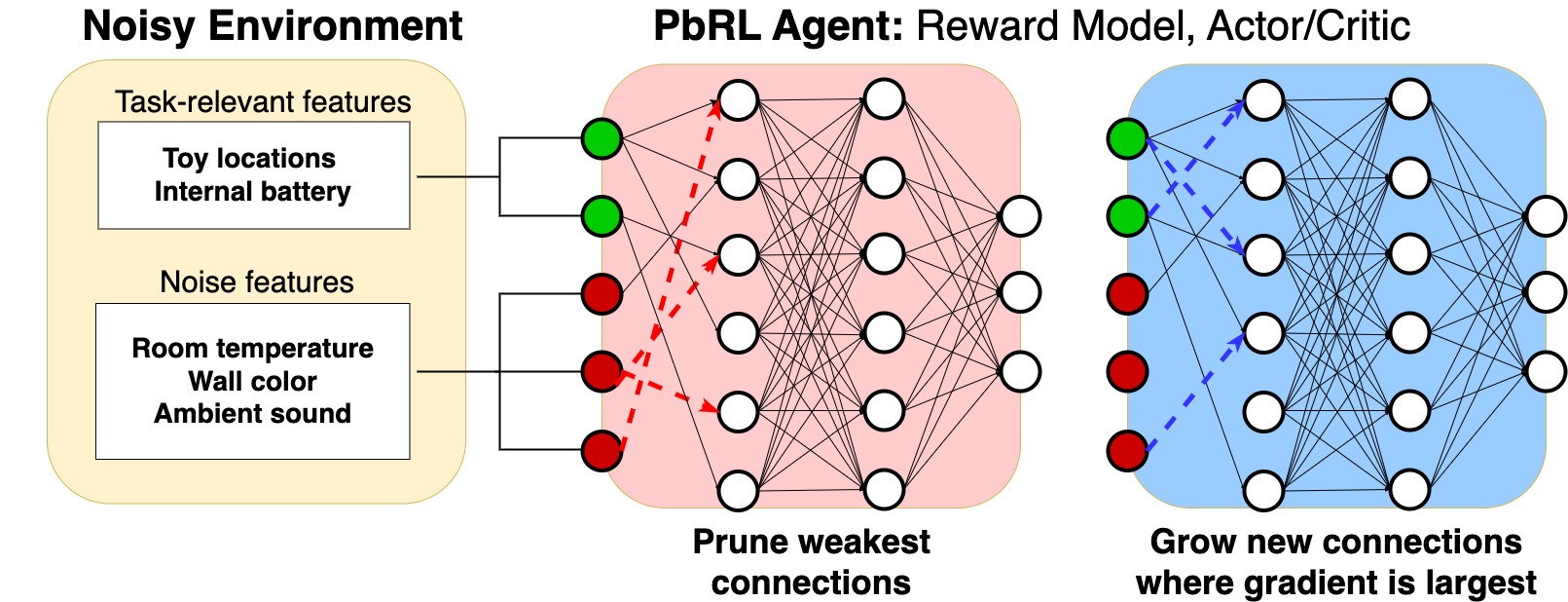}
  \caption{
Overview of R2N. Consider the example of a household robot tasked with cleaning a toy room from human preferences. The robot observes many features, although only a few are relevant to the task. 
R2N learns to connect with the input neurons that provide useful information by \emph{continually} pruning and growing new connections. }
  \label{fig:r2n_fig}
\end{figure*}

\begin{figure*}[h!]
  \centering
 \subfloat[Cartpole-swingup, Noise Fraction = 0.90]{\includegraphics[width=.45\textwidth]{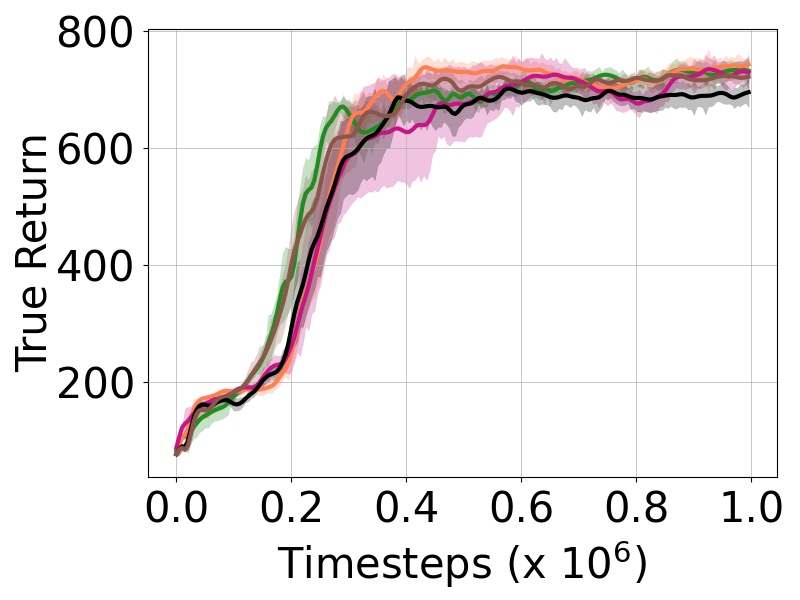}\label{fig:cartpole_feat_baselines}}
  \hfill
    \subfloat[Walker-walk, Noise Fraction = 0.90]{\includegraphics[width=.45\textwidth]{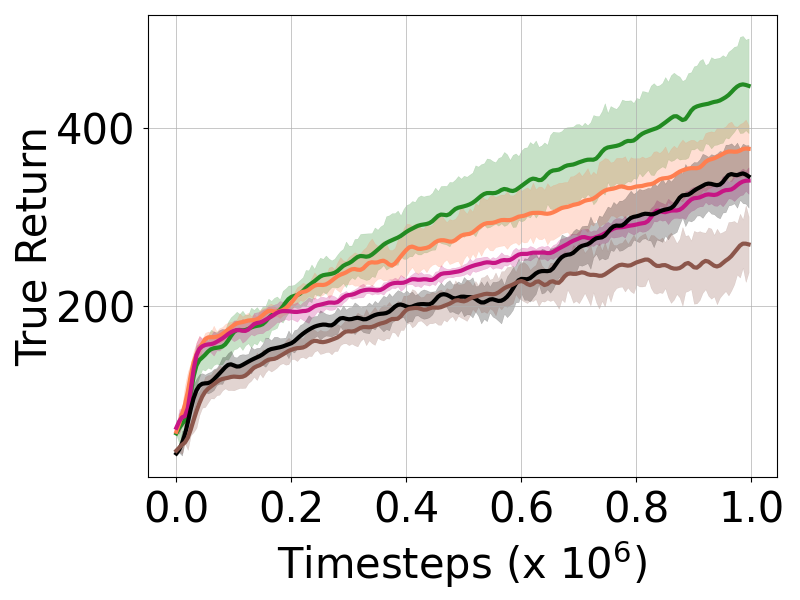}\label{fig:walker_walk_feat_baselines}}
  \hfill
    \subfloat[Quadruped-walk, Noise Fraction = 0.70]{\includegraphics[width=.45\textwidth]{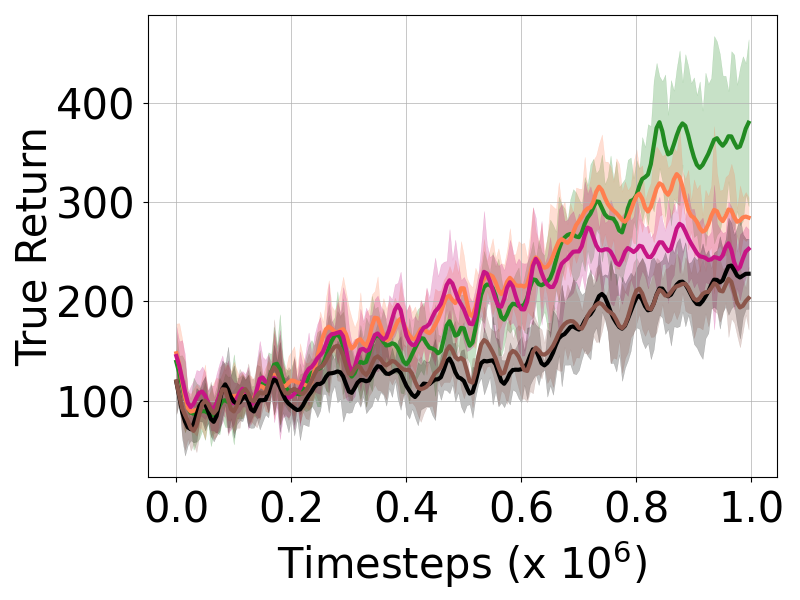}\label{fig:quadruped_walk_feat_baselines}}
  \hfill
  \subfloat[Cheetah-run, Noise Fraction = 0.90]{\includegraphics[width=.45\textwidth]{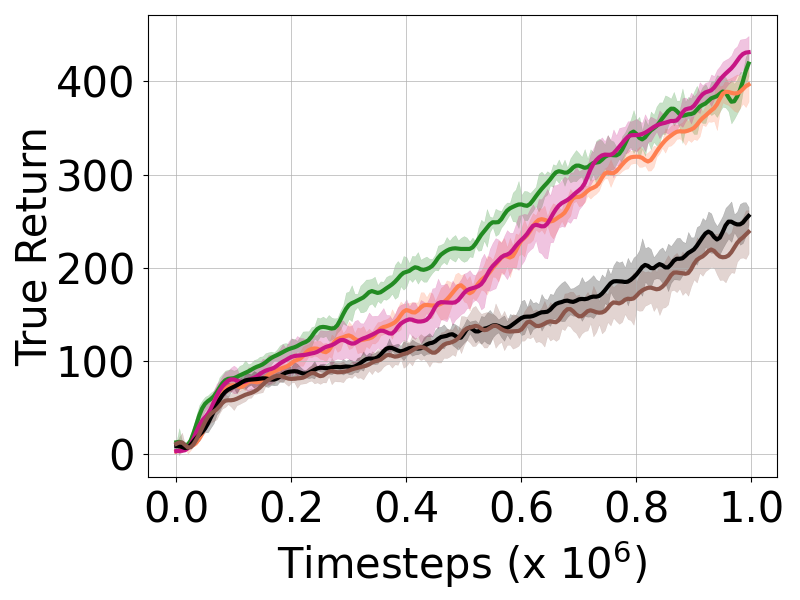}\label{fig:cheetah_run_feat_baselines}}
  \hfill
  \subfloat[Humanoid-stand, Noise Fraction = 0.70]{\includegraphics[width=.45\textwidth]{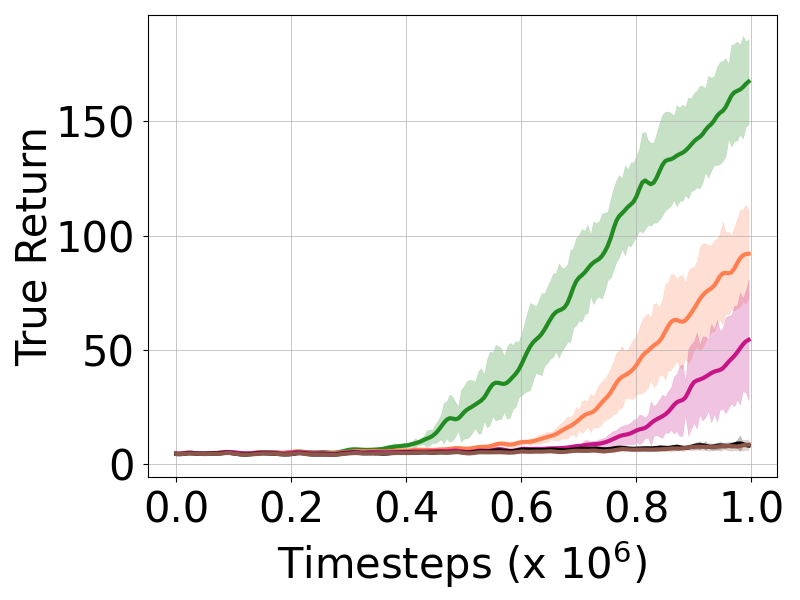}\label{fig:humanoid_stand_baselines}}
  \hfill
    \includegraphics[width=1.0\textwidth]{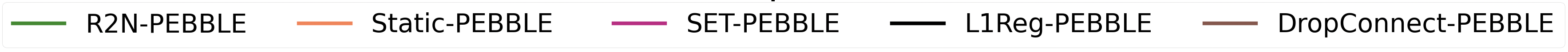}
  \captionsetup{skip=0.3em}
  \caption{These learning curves evaluate the effectiveness of R2N against various dynamic and static sparse training algorithms. R2N (green curves) maintains competitive or greater performance in all environments. Solid lines represent the mean, and shaded regions indicate the standard error across five runs.
  }
  
  \label{fig:main_results_comparing_sparsity}
\end{figure*}

\begin{figure*}[h!]
  \centering
 \subfloat[Cartpole-swingup, Noise Fraction = 0.90]{\includegraphics[width=.43\textwidth]{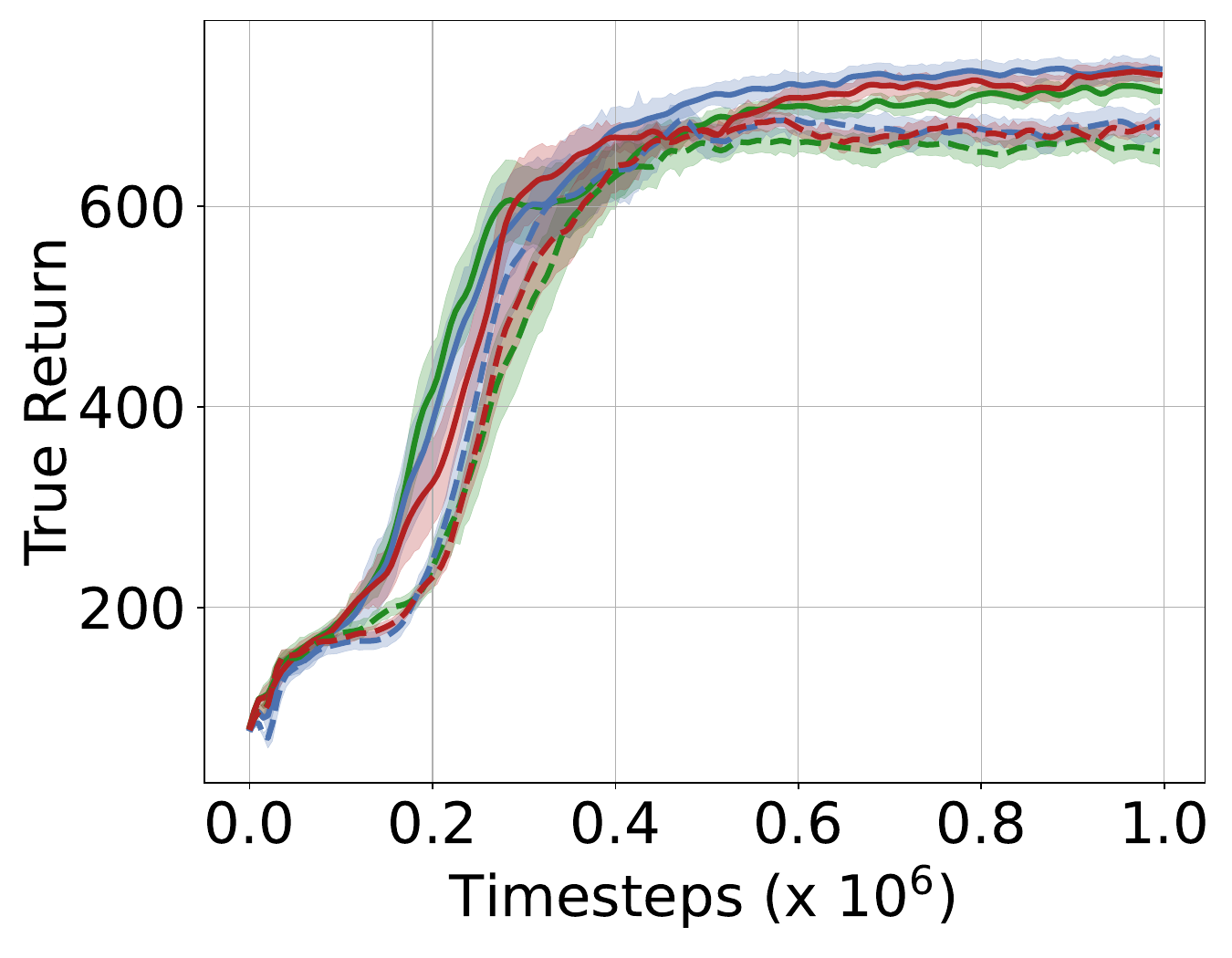}\label{fig:cartpole}}
  \hfill
    \subfloat[Walker-walk, Noise Fraction = 0.90]{\includegraphics[width=.43\textwidth]{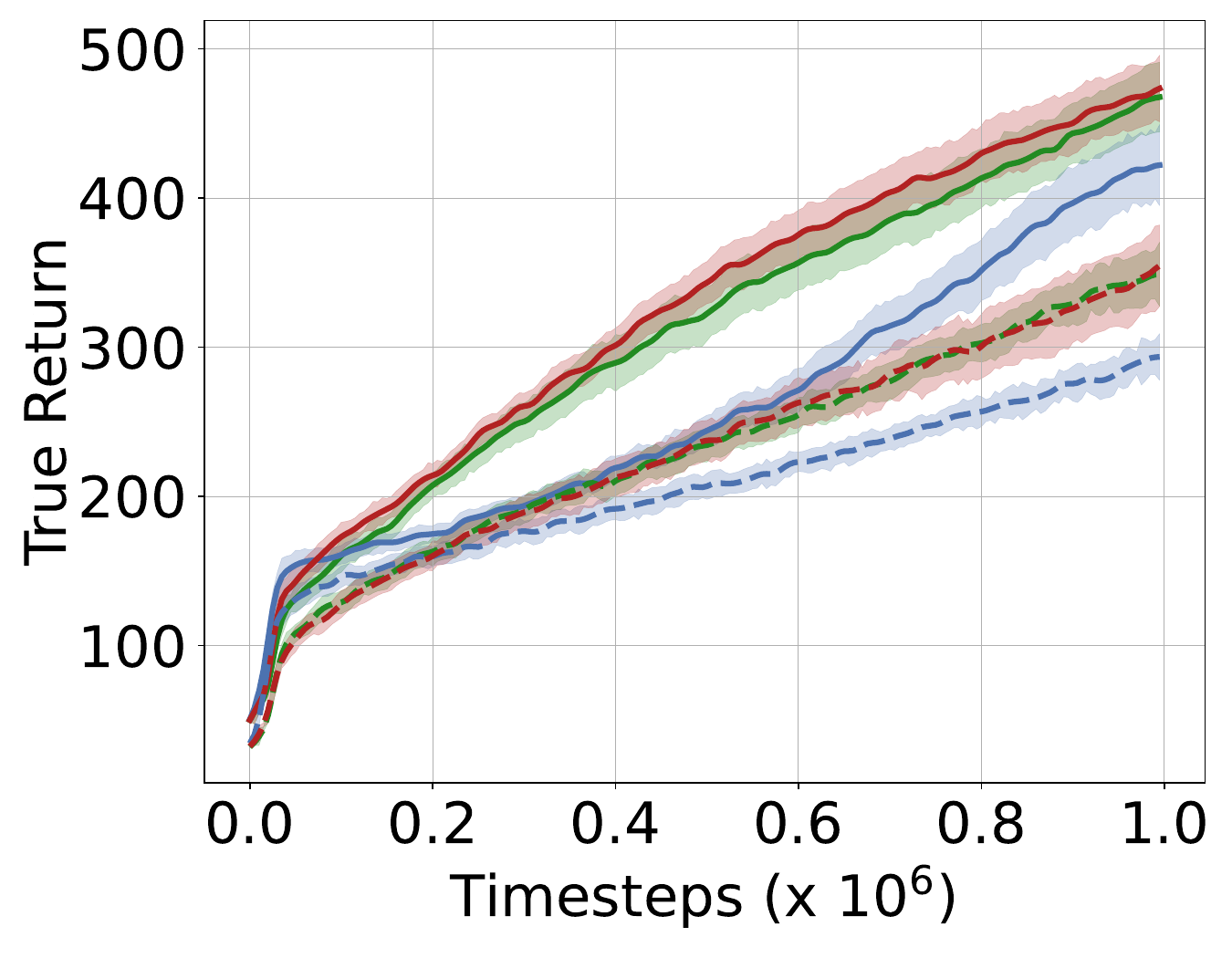}\label{fig:walker_walk}}
  \hfill
    \subfloat[Quadruped-walk, Noise Fraction = 0.70]{\includegraphics[width=.43\textwidth]{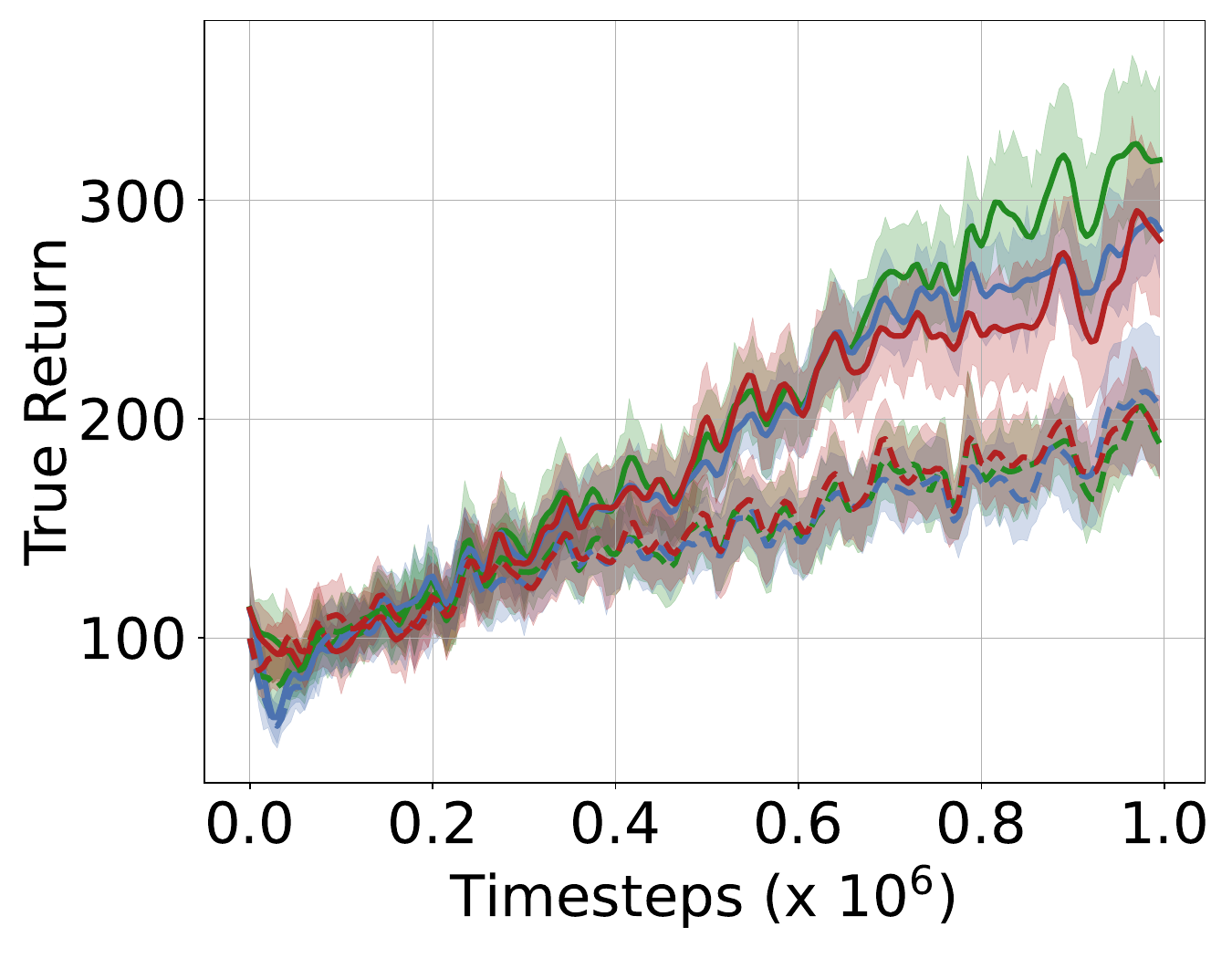}\label{fig:quadruped_walk}}
  \hfill
  \subfloat[Cheetah-run, Noise Fraction = 0.90]{\includegraphics[width=.43\textwidth]{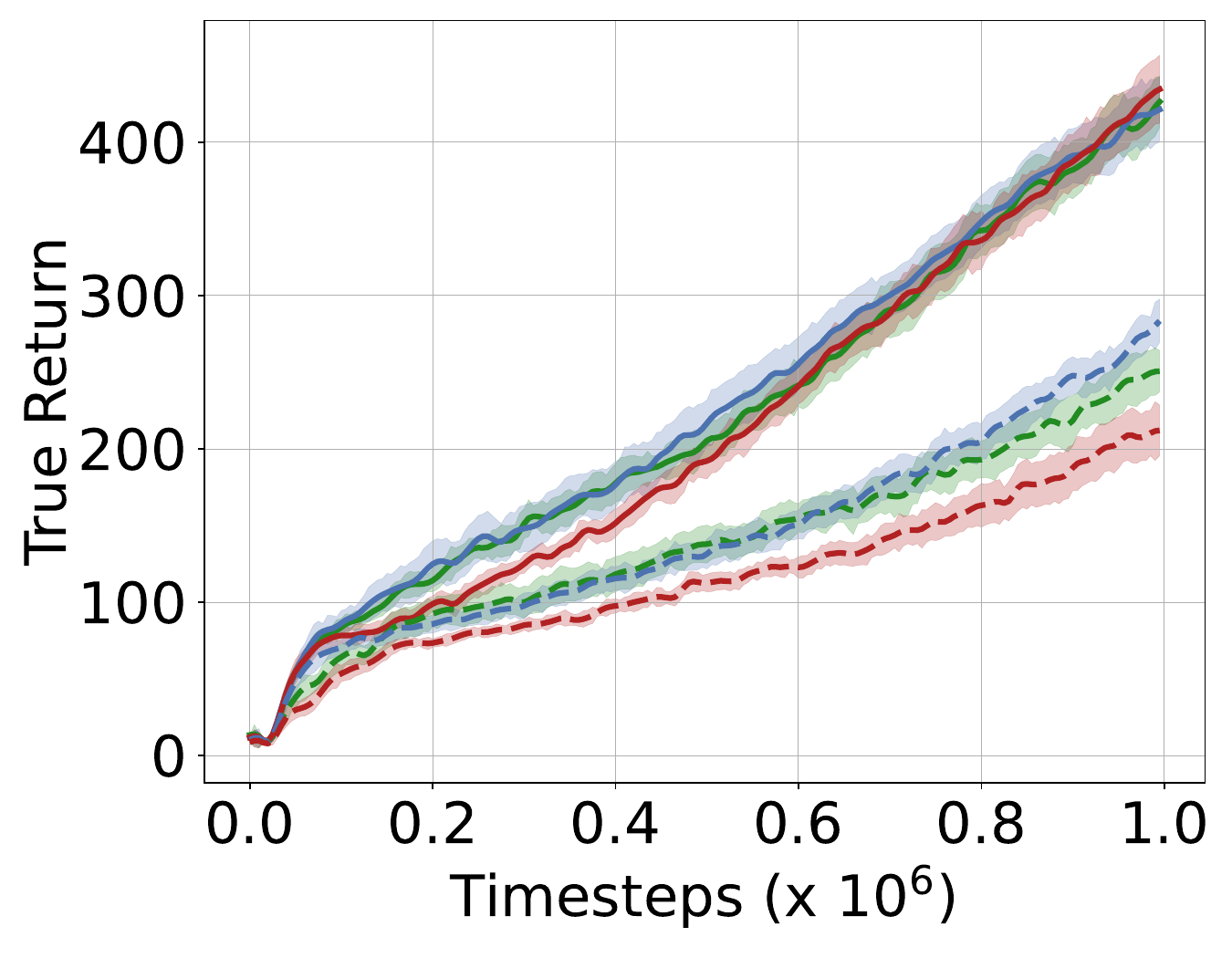}\label{fig:cheetah_run}}
  \hfill
  \subfloat[Humanoid-stand, Noise Fraction = 0.70]{\includegraphics[width=.43\textwidth]{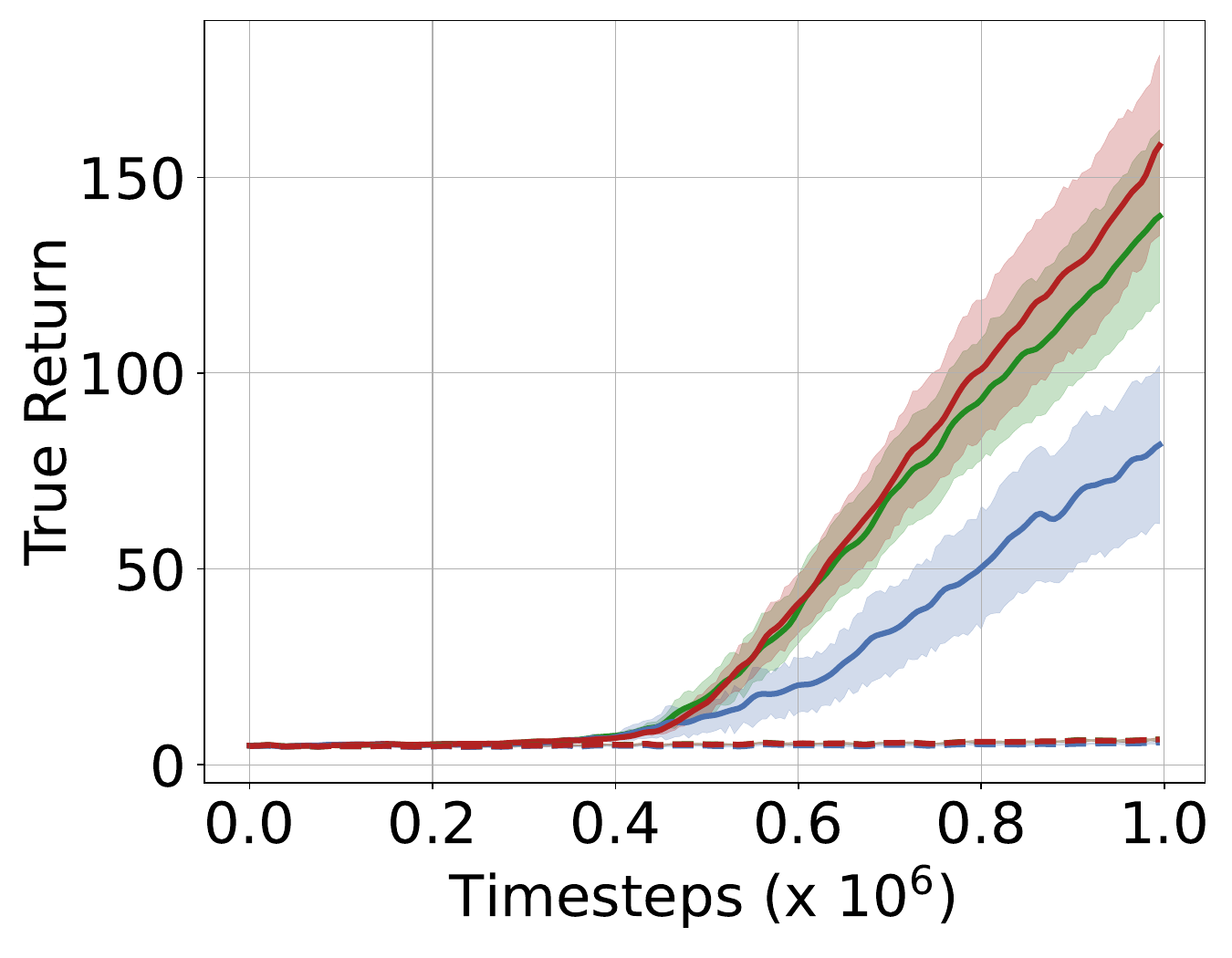}\label{fig:humanoid_stand}}
  \hfill
    \includegraphics[width=1.0\textwidth]{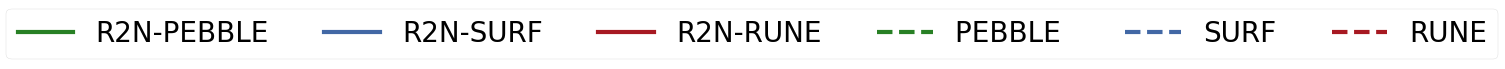}
  \captionsetup{skip=0.3em}
  \caption{These learning curves demonstrate that the use of R2N (solid curves) consistently outperforms the tested PbRL algorithms (dotted curves) in terms of final return and learning efficiency. 
  Solid lines and shaded regions represent the mean and standard error across 14 runs.
  }
  
  \label{fig:main_results_comparing_pbrl}
\end{figure*}

\section{Robust-to-Noise Preference Learning}\label{sec:method}
In this section, we introduce \textit{R2N}, a preference-based RL algorithm specifically designed to handle noisy environments (see Figure \ref{fig:r2n_fig}).  
The goal of R2N is to learn useful reward functions from feedback in environments that contain large numbers of task-irrelevant features (e.g., noise features). To do so, R2N applies dynamic sparse training techniques to PbRL algorithms for the learned reward model to focus on relevant environment features. Algorithm \ref{alg:R2N} outlines how DST is applied to the reward learning module in a PbRL algorithm, with novel components highlighted in blue.
R2N consists of two primary steps. First, at initialization, R2N randomly prunes the input layer of the reward model to a pre-defined sparsity level $s^R$ (see line 2 in Algorithm \ref{alg:R2N}). 
This is an important step, as prior works have shown that sparse neural networks can outperform their dense counterparts in both the supervised learning and RL settings \citep{set, rigl, ghada}. 
Second, 
after every $\Delta T^R$ weight updates in the training loop, we prune the weakest active connections in the reward model's input layer (see line 7 in Algorithm \ref{alg:R2N}). 
The strength of a connection is defined by the absolute value of its weight. After dropping a certain fraction $d^{R}_f \in (0,1)$ of the overall active connections, R2N grows the same number of connections in new locations (see line 8 in Algorithm \ref{alg:R2N}). Growing new connections ensures we maintain the same sparsity level throughout training. To choose which inactive connections to grow, we use the RigL algorithm \citep{rigl}, which activates connections with the highest gradient magnitude. 
As training the reward model is done via supervised learning, RigL is a more suitable DST candidate due to its demonstrated advantage over SET in supervised learning tasks \citep{rigl}.
This step enables the reward model to concentrate on the most pertinent features during training (see Figure \ref{fig:cheetah_run_feats} and Appendix \ref{sec:nn_connectivity}). We also repeat this procedure for the input layers of the actor and critic networks of the RL agent, as done in \citet{anf}. 

A key distinction between R2N and previous work on noisy settings is that R2N is specifically designed for PbRL. R2N is the first to apply DST to the reward learning module. 
Prior work focuses solely on the deep RL setting, only applying DST to the RL agent's networks \cite{anf}. 
This difference in setting further resulted in variations in the underlying DST algorithms. R2N purposely uses RigL as the reward models are trained via supervised learning, whereas \citet{anf} is based on the SET algorithm. Figure \ref{fig:main_results_comparing_sparsity} 
provides support for this design choice by demonstrating the advantages of using RigL over SET for preference learning. We further illustrate
the importance of R2N leveraging DST in both reward learning and RL modules for PbRL algorithms in Figure \ref{fig:dst_ablation_cheetah_run} and Appendix \ref{sec:dst_comp_study}. We observe that applying DST to only the reward module or only the RL module leads to significantly lower performance compared to R2N, which applies DST in both.

\begin{algorithm}
\caption{R2N}\label{alg:R2N}
 \textbf{Require}: 
\textcolor{blue}{Reward model sparsity level $s^R$, topology update period $\Delta T^R$,  drop fraction $d^{R}_{f}$}

\textbf{Require}: Set of collected preference data $D$ 

\begin{algorithmic}[1]
\State Initialize the reward model $\hat{r}_\theta$
\State \textcolor{blue}{Randomly prune the input layer of the reward model to sparsity level $s^R$} 



    \For{gradient step $t = 1 \ldots T$} \hfill $\triangleright$ Reward learning
        \State Sample minibatch $B \{ (\sigma^0, \sigma^1, y)^{i}\}^{b}_{i=1} \sim D$
    \State Update $\hat{r}_\theta$ with $L^{CE}, B$ (equ. \ref{eq:cross_entropy}) 
    \If{\textcolor{blue}{$t \ \mathrm{mod} \ \Delta T^R == 0$}} \hfill $\triangleright$ Update reward model topology
    \State \textcolor{blue}{Prune a fraction $d^{R}_{f}$ of the smallest magnitude reward model weights $\theta$}
   \State \textcolor{blue}{Grow $d^{R}_{f}$ new reward model weights $\theta$ via RigL}
   \EndIf
\EndFor

\end{algorithmic}
\end{algorithm}



\section{Experiments}
\label{sec:experiments}
In this section, we outline the research questions we address in this work and the respective experimental design and empirical results.
\subsection{Research Questions}
We consider the five research questions listed below. 
\begin{itemize}
    \item[RQ 1:] Can R2N outperform other sparse training baselines in extremely noisy environments?
    \item[RQ 2:] Can the addition of R2N boost the performance of a variety of PbRL algorithms?
    \item[RQ 3:]  
    How does the amount of noise and feedback affect the performance of R2N?
    \item[RQ 4:] What is the effect of applying DST to the reward learning versus RL module in R2N? 
    \item[RQ 5:] Can R2N learn in environments where the noise features imitate the task-relevant features?
\end{itemize}

\subsection{Experimental Design}
We evaluate R2N in the DMControl Suite \citep{dm_control}, a commonly used PbRL benchmark. 
More specifically, we consider the following five tasks: Cartpole-swingup, Walker-walk, Quadruped-walk, Cheetah-run, and Humanoid-stand. 
\paragraph{Baselines}

To evaluate the effectiveness of R2N (RQ 1), we compare it to four sparse training baselines: SET \cite{set}, Static Sparse Training, L1 Regularization \cite{l1reg}, and DropConnect \cite{dropconnect}. 
Each baseline was integrated with the PbRL algorithm PEBBLE \cite{pebble}. 
SET and Static Sparse Training are integrated within both the reward learning and actor/critic modules. L1 Regularization and DropConnect are integrated only within the reward learning module. PEBBLE is a PbRL algorithm that uses unsupervised exploration for policy initialization.

Next, to address RQ 2 and analyze the usefulness of R2N across diverse PbRL algorithms, we integrated it with two additional state-of-the-art PbRL algorithms: SURF and RUNE. These algorithms build upon PEBBLE by applying semi-supervised learning and data augmentation (SURF) and uncertainty-based exploration (RUNE). This results in the following baselines: PEBBLE, SURF, and RUNE.
We further show the performance of SAC \citep{sac} and ANF-SAC \citep{anf} in Appendix \ref{sec:r2n_vs_rl_algs}. However, note that these algorithms serve as oracle baselines as they have access to the ground truth reward during training, unlike the PbRL algorithms.

\paragraph{Implementation Details.}
The primary R2N-specific hyperparameters are the (1) reward model input layer sparsity level $s^R$, (2) the reward model topology update period $\Delta T^R$, and (3) the reward model drop fraction $d^{R}_{f}$. After a grid search, we set $s^R = 80\%$, $\Delta T^R = 100 $, and $d^{R}_{f} = 0.2 $. As we also apply DST to the RL agent, we use the same sparsity level, topology update period, and drop fraction for the actor and critic networks. 
For any PbRL-specific hyperparameters, we use the default values.
As for the RL agent, all methods use the SAC  algorithm with the same neural network architecture and associated SAC hyperparameters. 
See Appendix \ref{sec:Implementation_Details} for full hyperparameter details for all baselines.
For the PbRL baselines, we use a simulated teacher that provides preferences between two trajectory segments according to the ground truth reward function. 
Although our future work will involve human teachers, simulated teachers have commonly been used for evaluation in prior works \citep{christiano2017deep, pebble,surf, rune, sdp, bpref} to reduce the time and expense of human subject studies.

\paragraph{Training and Evaluation.}
We train all algorithms for 1 million timesteps. For evaluation, we show average offline performance (i.e., freeze the policy and act greedily) over ten episodes using the ground truth reward function. We perform this evaluation every $5000$ timesteps. 
Results are averaged over 5 or 14 seeds (Figure \ref{fig:main_results_comparing_pbrl}) with shaded regions indicating the standard error. 
To test for significant differences in final performance and learning efficiency (e.g., area under the curve: AUC), we perform a one-tailed  Welch's $t$-test (equal variances not assumed) with a $p$-value significance threshold of $0.05$. We use this statistical test, as it was found to be more robust to violations in test assumptions compared to other parametric and non-parametric tests \citep{colas2022hitchhikers}. 
See Appendix \ref{sec:table_results_rq1}--\ref{sec:ablation_tables} for a summary of final performance and AUC across all experiments.

\section{DMControl Results}\label{sec:main_results}
To address RQ 1 and 2, we evaluate R2N in three DMControl environments with noise fraction $n_{f}= 0.90$: Cartpole-swingup, Cheetah-run, and  Walker-walk. 
Recall that this results in a $10\times$ expansion in the state space size from the original tasks. 
This significantly increases the size of the state space for the original tasks, from 5 (Cartpole-swingup), 17 (Cheetah-run), and 24 (Walker-walk) to 51, 171, and 241 respectively. 
As a result, we use larger preference budgets of $400$ (Cartpole-swingup), $1000$ (Cheetah-run), and $4000$ (Walker-walk) to compensate for the increased task difficulty. 
Next, we evaluate R2N in two DMControl environments with noise fraction $n_{f}= 0.70$: Quadruped-walk and Humanoid-stand.
To keep the state space size for all ENE variants comparable, we use a smaller noise fraction for Quadruped-walk and Humanoid-stand. This is necessary as the original state spaces of these environments are three times larger than those of the DMControl environments previously tested. 
This results in the state space size increasing from 68 to 260 and 67 to 224 for the Quadruped-walk and Humanoid-stand environments respectively. 
In this setting, we used preference budgets of $4000$ (Quadruped-walk) and $10000$ (Humanoid-stand).  

\paragraph{R2N versus Sparse Training Baselines}
In Figure \ref{fig:main_results_comparing_sparsity} and Appendix \ref{sec:table_results_rq1}, we evaluate RQ 1, examining the effectiveness of R2N compared to four sparse training baselines. We find that in all five environments, R2N-PEBBLE (green curves) is the only algorithm that \emph{consistently} achieves superior performance. In particular, R2N-PEBBLE significantly outperforms L1-Regularization (black curves) and DropConnect (brown curves) in terms of learning efficiency in four out of five environments ($p \le 0.034$). While Static-PEBBLE (orange curves) and SET-PEBBLE (pink curves) prove to be more competitive, R2N maintains significant performance improvement both in Humanoid-stand (Static and SET-PEBBLE; AUC and final return, $p \le 0.024$) and Cheetah-run (Static-PEBBLE; AUC, $p \le 0.018$).

\begin{figure*}[h!]
  \centering
 \subfloat[Noise Study]{\includegraphics[width=0.33\textwidth, ,  trim = 0cm 2.4cm 0cm 0cm, clip]{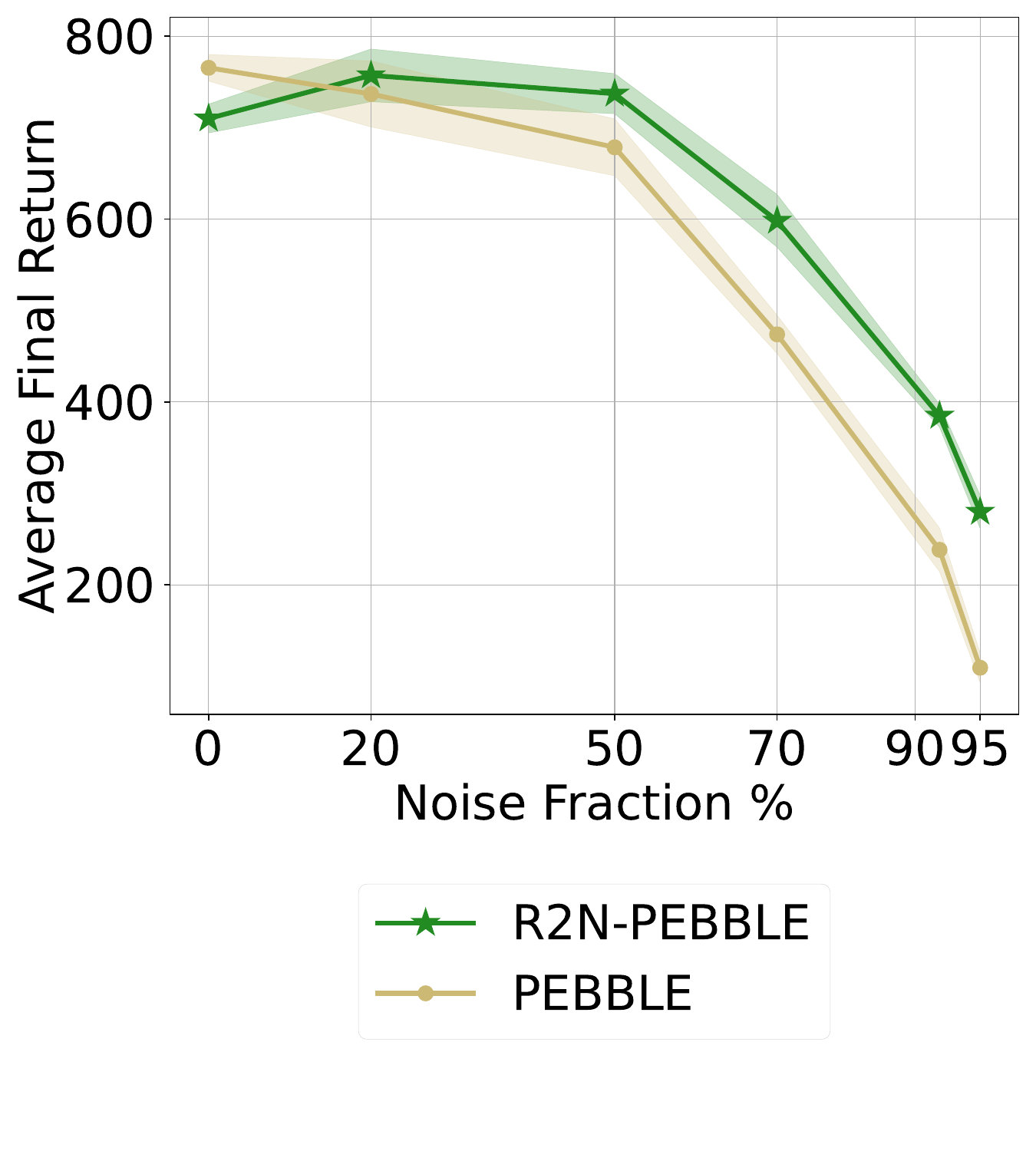}\label{fig:noise_ablation}}
  \hfill   
  \centering
 \subfloat[Feedback Study]{\includegraphics[width=0.33\textwidth,  trim = 0cm 2.4cm 0cm 0cm, clip]{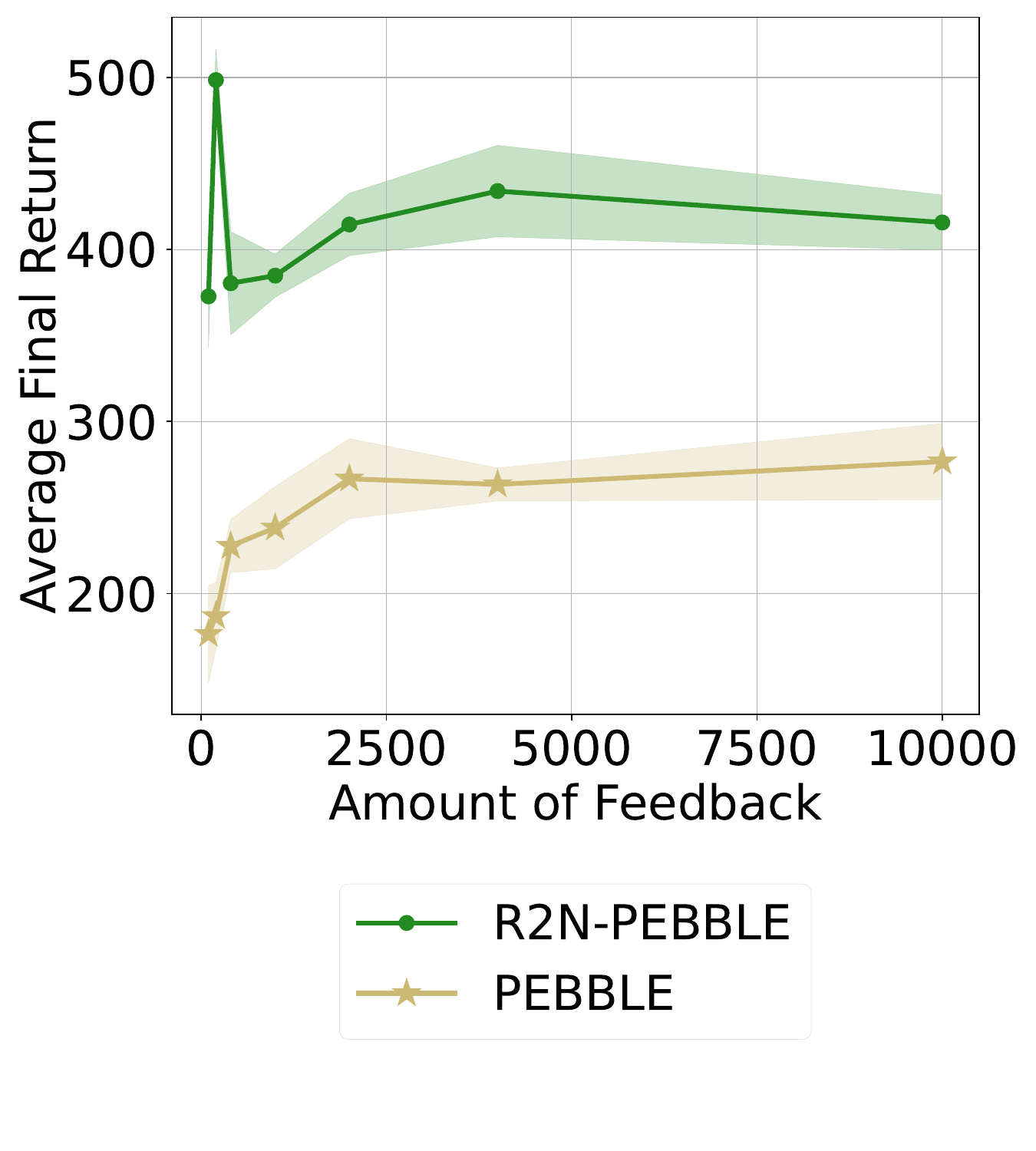}\label{fig:fb_ablation}}
  \hfill   
  \centering
 \subfloat[Neural Network Connections]{\includegraphics[width=0.33\textwidth,  trim = 0cm 2.5cm 0cm 0cm, clip]{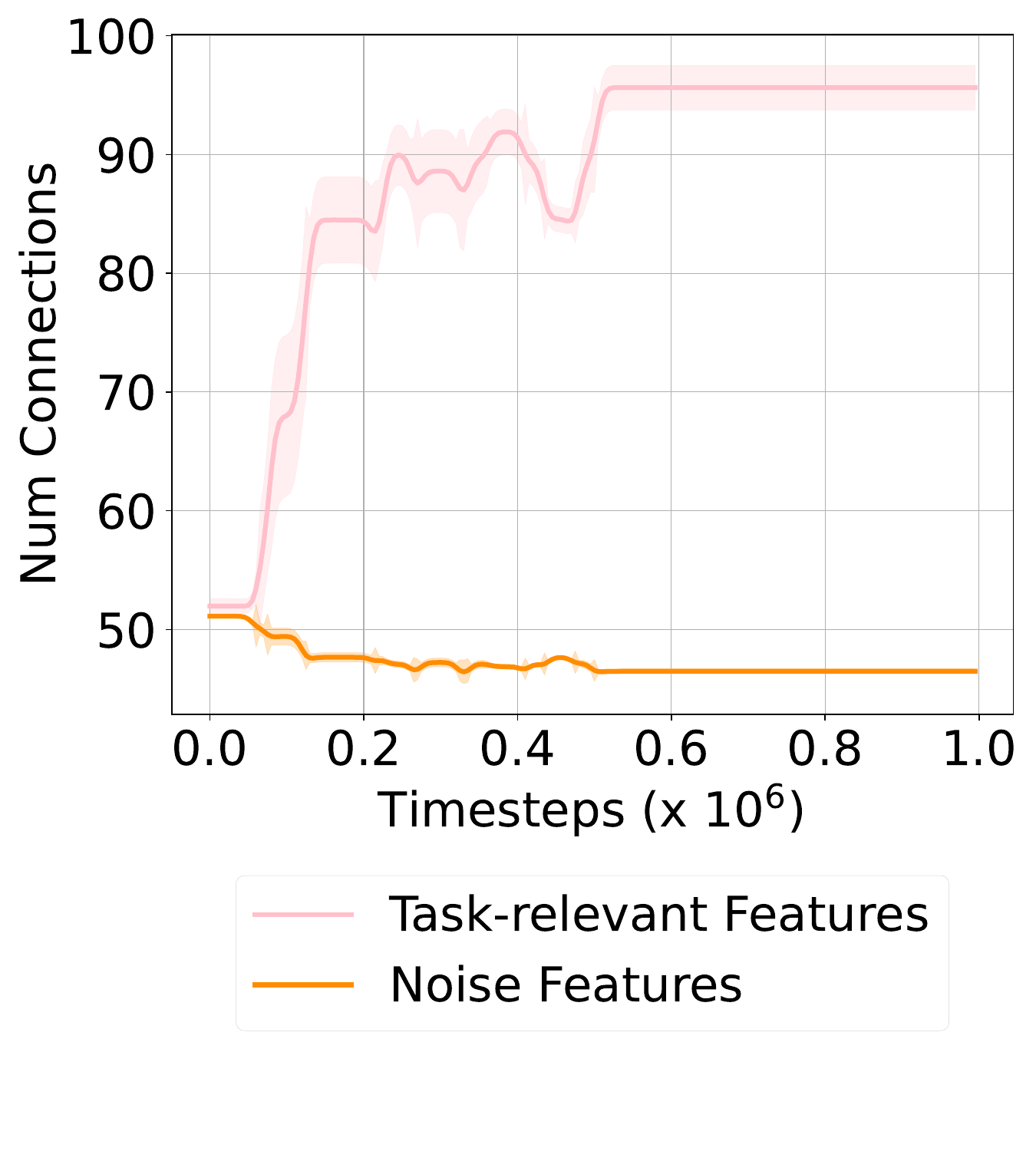}\label{fig:cheetah_run_feats}}
  \hfill   
    \centering
 \subfloat[DST Component Study]{\includegraphics[width=0.49\textwidth,  trim = 0cm 3cm 0cm 0cm, clip]{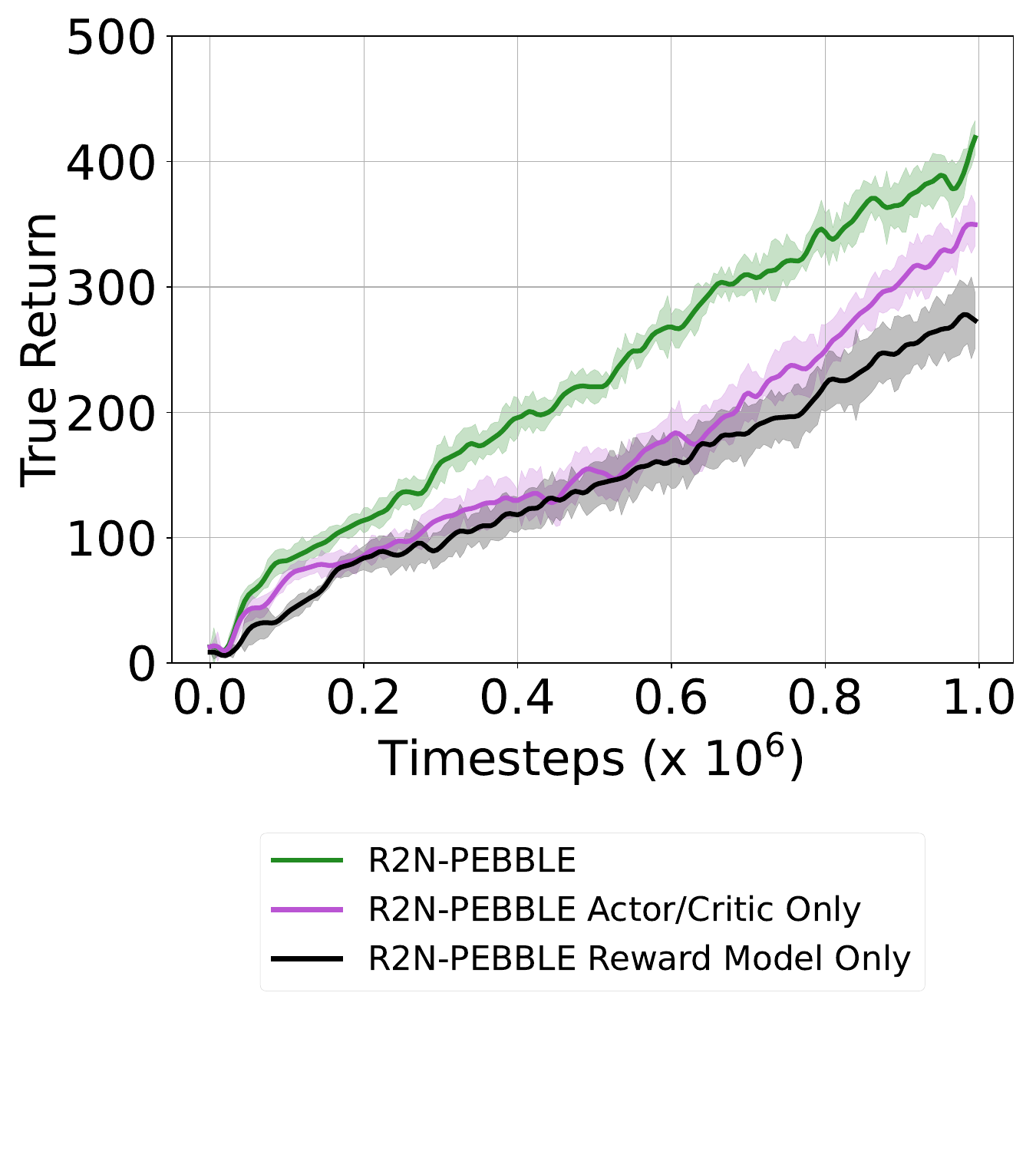}\label{fig:dst_ablation_cheetah_run}}
  \hfill   
    \centering
 \subfloat[Imitating Noise Study]{\includegraphics[width=0.49\textwidth, trim = 0cm 3cm 0cm 0cm, clip]{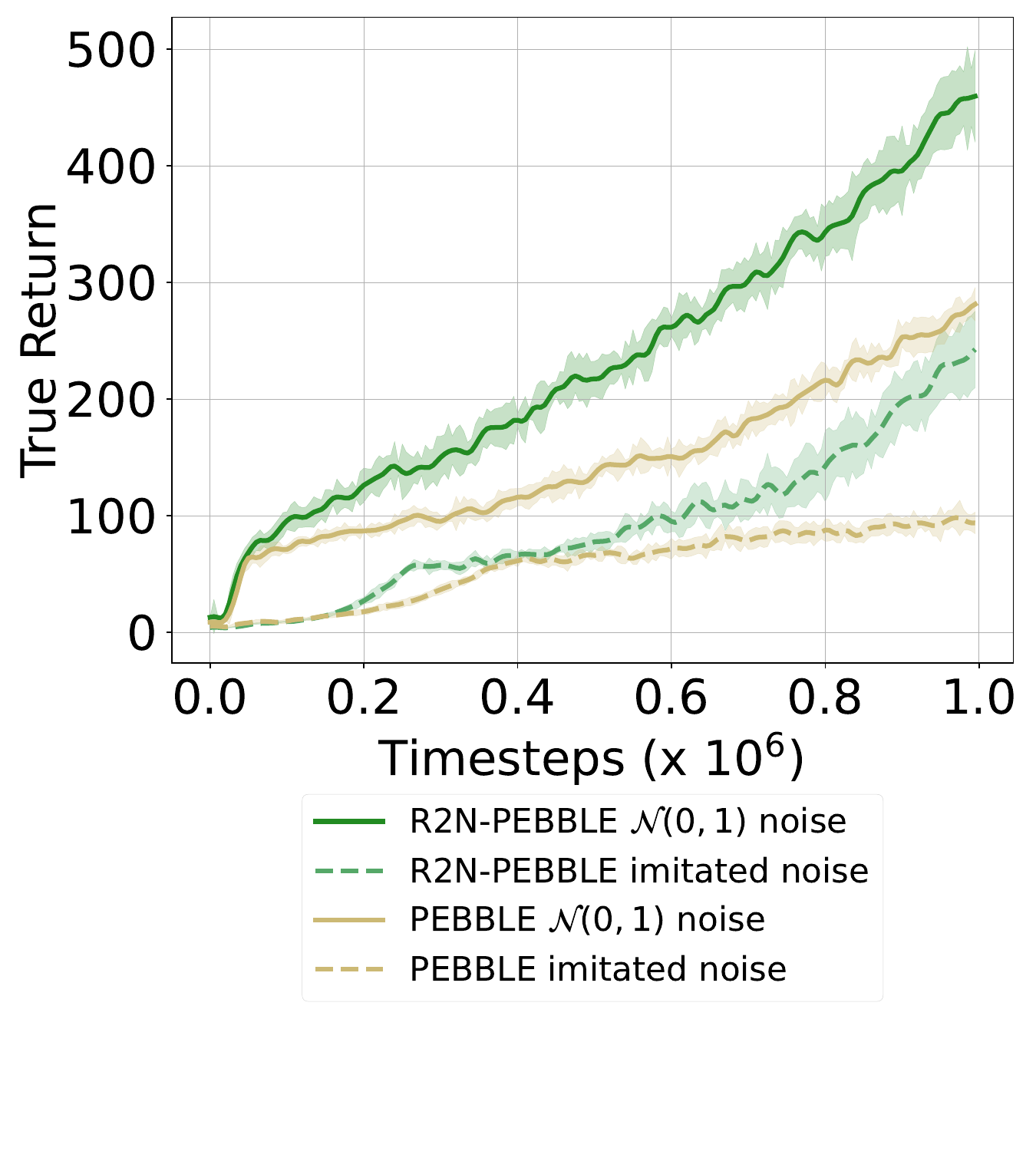}\label{fig:imitating_noise_cheeetah_run}}
  \hfill   
  \caption{Further studies in Cheetah-run: (a) effect of noise fraction, (b) effect of feedback budget, (c) average number of neural network connections to task-relevant versus noise features in a reward model with R2N, (d) DST component ablation, and (e) effect of noise feature distributions. Solid lines and shaded regions represent the mean and standard error across five runs.}\label{fig:ablation_studies}
  \vspace{-1em}
\end{figure*}

\paragraph{Effectiveness of R2N across PbRL Algorithms}
Next, we focus on RQ 2, to understand whether R2N can boost performance across a diverse set of PbRL algorithms.
In Figure \ref{fig:main_results_comparing_pbrl} and Appendix \ref{sec:table_results_rq2}, we find that the addition of R2N (solid lines) significantly improved both the learning efficiency ($p \le 0.021$) and final return ($p \le 0.006$) of the base PbRL algorithm (dotted lines) across all 15 tested baseline-environment combinations.\footnote{We perform three separate Welch's $t$-tests comparing the base PbRL algorithm to its R2N variant.} 
Moreover, in Table \ref{tab:percent_increase}, we find that R2N resulted in a substantial increase in average final return over the base PbRL algorithms, with almost half of the cases achieving a 50\% or greater performance boost.
We argue that R2N can outperform the current PbRL algorithms because it is designed to disregard irrelevant features, an important trait for extremely noisy environments. In Figure \ref{fig:cheetah_run_feats}, we find that R2N adapts its reward model to focus more on task-relevant features. This results in a significantly greater number of connections to the task-relevant features (pink curve) as compared with the noise features (orange curve). We observe a similar pattern for the actor and critic networks (see Appendix \ref{sec:nn_connectivity}). 
Existing PbRL algorithms primarily focus on improving feedback efficiency in conventional RL environments that contain only task-relevant features --- these results show that such improvements alone may be insufficient for more noisy DMControl environments. 
However, we also find that R2N can achieve comparable performance in noise-free environments (see Figure \ref{fig:0noise}), indicating its broader applicability.


\section{Sensitivity and Ablation Studies}\label{sec:ablations} 
Research questions 3-5 pertain to the utility of R2N across the following dimensions: (1) the number of noise features, (2) the
number of feedback queries, (3) the effect of DST on individual learning modules in R2N, and (4) the type of noise features. 
We perform this analysis in the Cheetah-run environment. In these studies, we restrict our comparison of R2N-PEBBLE with only PEBBLE, as PEBBLE performed comparably to the other tested PbRL algorithms in Section \ref{sec:main_results}. We include additional results for the Walker-walk environment in Appendix \ref{sec:additional_results}. These results demonstrate similar performance trends.

\paragraph{Noise Study}
First, we aim to understand how effective R2N is at various noise fractions, RQ 3. Therefore, we fix the preference budget to $1000$ and vary the noise fraction $n_{f} \in \{0, 0.2, 0.5, 0.7, 0.9, 0.95\}$. 
In Figure \ref{fig:noise_ablation} and Table \ref{tab:noise_ablation_table}, we find that for higher noise fractions in Cheetah-run, R2N-PEBBLE  
maintains significant improvement over PEBBLE in learning efficiency with $p \leq 0.014$ for $n_{f} \ge 50\%$, and in final return with $p \leq 0.007$ for $n_{f} \ge 70\%$.
Unsurprisingly, given the small feedback budget, both methods perform worse as the noise fraction increases. 
\paragraph{Feedback Study}
We continue our analysis of RQ 3 and evaluate both algorithms under multiple preference budgets. 
We set the noise fraction $n_{f}=0.90$ and vary the preference budget $b \in \{100, 200, 400, 1000, 2000, 4000, 10000\}$. 
In Figure \ref{fig:fb_ablation} and Table \ref{tab:fb_ablation_table}, we find that in Cheetah-run, R2N-PEBBLE significantly outperforms PEBBLE in terms of learning efficiency (AUC) and final return with $p \leq 0.009$ for all tested feedback budgets.


\paragraph{DST Component Ablation}
In R2N, we apply DST to both the reward model and actor/critic networks. In RQ 4, our goal is to understand the importance of DST for both learning modules. In this ablation, we consider a noise fraction $n_{f}= 0.90$ and a preference budget of 1000. 
In Figure \ref{fig:dst_ablation_cheetah_run} and Table \ref{tab:dst_ablation_table}, 
we find that in Cheetah-run, full R2N (green curve) significantly outperforms the R2N variants that apply DST only to the RL module (purple curve; $p \leq 0.022$ for AUC and final return) or only to the reward learning module (black curve; $p \leq 0.002$ for AUC and final return).
This demonstrates that it is important for the reward model and RL agent to learn to avoid irrelevant features in R2N-PEBBLE. 

\paragraph{Imitating Real Features Study.}\label{sec:imitation}
Experiments thus far consider extremely noisy environments where noise features were sampled from $\mathcal{N}(0,1)$. For RQ 5, we increase the difficulty by using noise features that mimic task-relevant features. 
We find the distribution of each task-relevant feature as follows. First, we train a SAC agent in a noise-free environment for 1 million timesteps. Second, we perform policy rollouts and store the state transitions. Third, with the state transition data, we can create a histogram for each feature. 
Fourth, we sample from each feature distribution to create a noise feature. 
Due to increased difficulty, we set the noise fraction $n_{f}= 0.90$ but use a higher preference budget of 4000. 
In Figure \ref{fig:imitating_noise_cheeetah_run} and Table \ref{tab:imitating_noise_table}, we find that in this setting R2N-PEBBLE (dotted green curve) maintains significantly greater performance than PEBBLE (dotted yellow curve) with $p \leq 0.004$ for both AUC and final return. However, for both R2N-PEBBLE and PEBBLE, we observe performance degradation compared to the setting in which the noise is sampled from the standard normal distribution (solid curves). 

\section{Conclusion}
\label{sec:conclusion}
For RL agents to become commonplace, agents should be able to learn from people in human-centered environments (e.g., home, school, office).
However, humans live in a world full of information, most of which is not necessary for completing individual tasks. 
Current state-of-the-state PbRL algorithms do not consider the impact of irrelevant environment features and are, consequently, unable to adequately learn in this setting. 
To that end, we present R2N, a novel \emph{robust-to-noise} PbRL algorithm that leverages dynamic sparse training techniques to learn robust reward models in \emph{extremely noisy environments}.
R2N continually adjusts the network topology of both the reward model and RL agent networks to focus on task-relevant features. This enables R2N to successfully learn in environments where up to 95\% of the features are noise. 

This work represents the first PbRL algorithm specifically designed to learn in extremely noisy environments. As such, several promising research directions remain to be explored. 
For example, to provide a proof-of-concept of R2N, we use a simulated teacher to obtain preferences. To further validate our obtained results and confirm their generalizability to human preferences, future work must perform a human-subjects study. 
In addition, we limit our investigation of R2N to the Extremely Noisy Environment setting, where noise features are added to existing RL environments. To consider more real-world settings, R2N should be studied with real robots that may be receiving pixel input.
Lastly, this work assumes that a feature's relevance is constant (i.e., always useful or always noise). However, an
interesting extension would be learning to filter irrelevant features in the continual learning setting, in which features can be relevant to one task and irrelevant to the next. 
\bibliographystyle{ACM-Reference-Format} 
\bibliography{main}


\appendix

\section*{Appendix}\label{sec:appendix}

\section{Implementation Details}\label{sec:Implementation_Details}

In this section, we present further implementation details on our method R2N, as well as the other baselines compared in our study. Our results in Section \ref{sec:main_results} show that R2N can boost the performance of multiple preference-based reinforcement learning algorithms, especially in extremely noisy environments. For an overview of the dimensions of the state space in each ENE studied, see Table~\ref{tab:obs_size_table}. 

The primary R2N-specific hyperparameters are the (1) reward model input layer sparsity level $s$, (2) the reward model topology update period $\Delta T$, and (3) the reward model drop fraction $d_{f}$. To set these values, we performed a grid search. For the reward model input layer sparsity level $s$, we used the default sparsity level as in ANF \citep{anf}.
Next, for the reward model topology update period $\Delta T$, we tested values for $\Delta T \in \{20, 100, 250, 500, 1000\}$. Lastly,  for the reward model drop fraction $d_{f}$, we tested values for $d_{f} \in \{0.05, 0.1, 0.2\}$. 

To train a single R2N model (i.e., one seed), we used 1 GPU, 20-24 hours of run-time and 25-50G of memory.
\begin{table*}[h!]
\centering
\captionsetup{width=0.8\textwidth}
\caption{This table shows the size of the state space for the original environments and the corresponding extremely noisy environment (ENE) variants. }
\label{tab:obs_size_table}
\begin{tabular}{@{}llll@{}}
\toprule
\textbf{Task}    & \textbf{Original State Space} & \textbf{Noise} & \textbf{State Space of ENE} \\ \midrule
Cartpole-swingup & 5                             & 0.9            & 51                          \\
Walker-walk      & 24                            & 0.9            & 241                         \\
Cheetah-run      & 17                            & 0.9            & 171                         \\
Quadruped-walk   & 68                            & 0.7            & 260                         \\
Humanoid-stand   & 67                            & 0.7            & 224           \\
\bottomrule
\end{tabular}%
\end{table*}

\paragraph{Hyperparameters.}
An overview of the hyperparameters of the standard RL algorithms (without preference learning) is provided in Table~\ref{tab:SAC_hyperparams}. The specific settings of R2N and our PbRL baselines are given in Table~\ref{tab:PbRL_hyperparams}. The hyperparameters for the sparse training baselines are outlined in Table \ref{tab:sparsetraining_hyperparams}. 

\begin{table*}[h!]
\centering
\captionsetup{width=0.9\textwidth}
\caption{Hyperparameters for SAC and ANF-SAC, the standard RL algorithms that learn from the environment's true reward signal, not via preference learning.
Note, however, that we present R2N's hyperparameters for the actor and critic networks here as well.}
\label{tab:SAC_hyperparams}
\resizebox{.68\textwidth}{!}{%
\begin{tabular}{ll}
\toprule
\textbf{Hyperparameter}                          & \textbf{Value}                                                                                                                 \\
\midrule
\emph{SAC (Shared by all algorithms)}                                                                                                                                                \\
\hspace{2mm} optimizer                                        & Adam \citep{kingma15_adam}                                                                                                                       \\
\hspace{3mm}discount                                         & 0.99                                                                                                                           \\
\hspace{3mm}actor learning rate                              & $10^{-4}$                                                                                                                             \\
\hspace{3mm}critic learning rate                             & $10^{-4}$                                                                                                                           \\
\hspace{3mm}alpha learning rate                              & $10^{-4}$                                                                                                                          \\
\hspace{3mm}actor betas                                      & 0.9, 0.999                                                                                                                     \\
\hspace{3mm}critic betas                                     & 0.9, 0.999                                                                                                                     \\
\hspace{3mm}alpha betas                                      & 0.9, 0.999                                                                                                                     \\
\hspace{3mm}target smoothing coefficient                     & 0.005                                                                                                                          \\
\hspace{3mm}actor update frequency                           & 1                                                                                                                              \\
\hspace{3mm}critic target update frequency                   & 2                                                                                                                              \\
\hspace{3mm}init temperature                                 & 0.1                                                                                                                            \\
\hspace{3mm}batch size                                       & 1024                                                                                                                           \\
\hspace{3mm}replay buffer size                               & $10^6$                                                                                                                            \\
\hspace{3mm}initial collect steps                            & 1000                                                                                                                           \\
\hspace{3mm}network type                                     & MLP                                                                                                                            \\
\hspace{3mm}number of hidden layers                          & 2                                                                                                                              \\
\hspace{3mm}number of neurons per hidden layer               & 1024                                                                                                                           \\
\hspace{3mm}nonlinearity                                     & ReLU                                                                                                                           \\
\hspace{3mm}number of training steps                         & $10^6$                                                                                                                            \\
\midrule
\emph{ANF-SAC and R2N (actor \& critics)}                                 &                                                                                                                                \\
\hspace{3mm}sparsity level input layer                      & 0.8                                                                                                                            \\
\hspace{3mm}drop fraction $d_{f}$                                   & 0.05                                                                                                                           \\
\hspace{3mm}topology-change period $\Delta T$                          & 1000                                                                                                                           \\
\hspace{3mm}new weights init value                           & 0                                                                                                                              \\
\hspace{3mm}DST method                                       & RigL \citep{rigl}                                                                                                                         \\
\hspace{3mm}sparsify target networks                        & false   \\  
\bottomrule
\end{tabular}
}%
\end{table*}

\begin{table*}[h!]
\centering
\captionsetup{width=0.85\textwidth}
\caption{Hyperparameters for the PbRL algorithms used, including our method R2N.}
\label{tab:PbRL_hyperparams}
\resizebox{.75\textwidth}{!}{%
\begin{tabular}{ll}
\toprule
\textbf{Hyperparameter}                                                                                       & \textbf{Value}                                                                                                                 \\
\midrule
\emph{PEBBLE (reward model hyperparameters}\\ \emph{shared by all PbRL algorithms)} &                                                                                                                                \\
\hspace{2mm} trajectory segment size                                                                                       & 50                                                                                                                             \\
\hspace{3mm}number of unsupervised exploration steps                                                                           & 9000                                                                                                                           \\
\hspace{3mm}learning rate                                                                                                 & 0.003                                                                                                                          \\
\hspace{3mm}batch size                                                                                                    & 128                                                                                                                            \\
\hspace{3mm}ensemble size                                                                                                 & 3                                                                                                                              \\
\hspace{3mm}network type                                                                                                  & MLP                                                                                                                            \\
\hspace{3mm}number of hidden layers                                                                                       & 4                                                                                                                              \\
\hspace{3mm}number of neurons per hidden layer                                                                            & 128                                                                                                                            \\
\hspace{3mm}nonlinearity                                                                                                  & LeakyReLU                                                                                                                     \\
\hspace{3mm}optimizer                                                                                                     & Adam \citep{kingma15_adam}                                                                                                                         \\
\hspace{3mm}replay buffer size & $10^{5}$
\\
\hspace{3mm}feedback frequency                                                                                            & 5000                                                                                                                           \\
\hspace{3mm}reward batch size                                                                                             & feedback budget / 100                                                                                                          \\
\hspace{3mm}trajectory sampling scheme                                                                                    & \begin{tabular}[c]{@{}l@{}}DMC -- Uniform \end{tabular}                                    \\
\hspace{3mm}training epochs                                                                                               & 50                                                                                                                             \\ \midrule
\emph{SURF}                                                                                                         &                                                                                                                                \\
\hspace{3mm}confidence threshold $\tau$                                                                                                    & 0.99                                                                                                                           \\
\hspace{3mm}loss weight $\lambda$                                                                                                        & 1                                                                                                                                                                                                                       \\
\hspace{3mm}inverse label ratio                                                                                           & 10                                                                                                                             \\
\hspace{3mm}data augmentation window                                                                                      & 5                                                                                                                              \\
\hspace{3mm}crop range                                                                                                    & 5                                                                                                                              \\ \midrule
\emph{RUNE}                                                                                                         &                                                                                                                                \\
\hspace{3mm}beta schedule                                                                                                 & linear decay                                                                                                                   \\
\hspace{3mm}beta init                                                                                                     & 0.05                                                                                                                           \\
\hspace{3mm}beta decay                                                                                                    & 0.00001                                                                                                                        \\ \midrule
\emph{R2N (reward models)}                                                                                                          &                                                                                                                                \\
\hspace{3mm}sparsity level input layer $s_i$                                                                                         & 0.8                                                                                                                            \\
\hspace{3mm}drop fraction $d_{f}$                                                                                     & 0.20                                                                                                                           \\
\hspace{3mm}topology-change period $\Delta T$                                                                                                 & 100  \\
\bottomrule
\end{tabular}%
}
\end{table*}

\begin{table*}[h!]
\centering
\captionsetup{width=0.9\textwidth}
\caption{Hyperparameters for Sparse Training Baslines}
\label{tab:sparsetraining_hyperparams}
\resizebox{.5\textwidth}{!}{%
\begin{tabular}{ll}
\toprule
\textbf{Hyperparameter}                          & \textbf{Value}                                                                                                                 \\
\midrule
             
\emph{SET-PEBBLE}                                                                                 \\
\hspace{3mm} sparsity level input layer $s_i$   & 0.8   \\
\hspace{3mm}drop fraction $d_{f}$ & 0.20        \\                                                            
\emph{Static-PEBBLE} \\
\hspace{3mm} sparsity level input layer $s_i$    & 0.8 \\
\emph{DropConnect} \\  
\hspace{3mm}drop fraction $d_{f}$ & 0.20  \\
\emph{L1 Regularization} \\  
\hspace{3mm}$\lambda$ & 0.01  \\
\bottomrule
\end{tabular}%
}
\end{table*}

\section{Additional Results}\label{sec:additional_results}
\subsection{Neural Network Connections to Relevant versus Noise Features}\label{sec:nn_connectivity}

\begin{figure*}[h!]
  \centering
 \subfloat[Actor Network]{\includegraphics[width=0.33\textwidth]{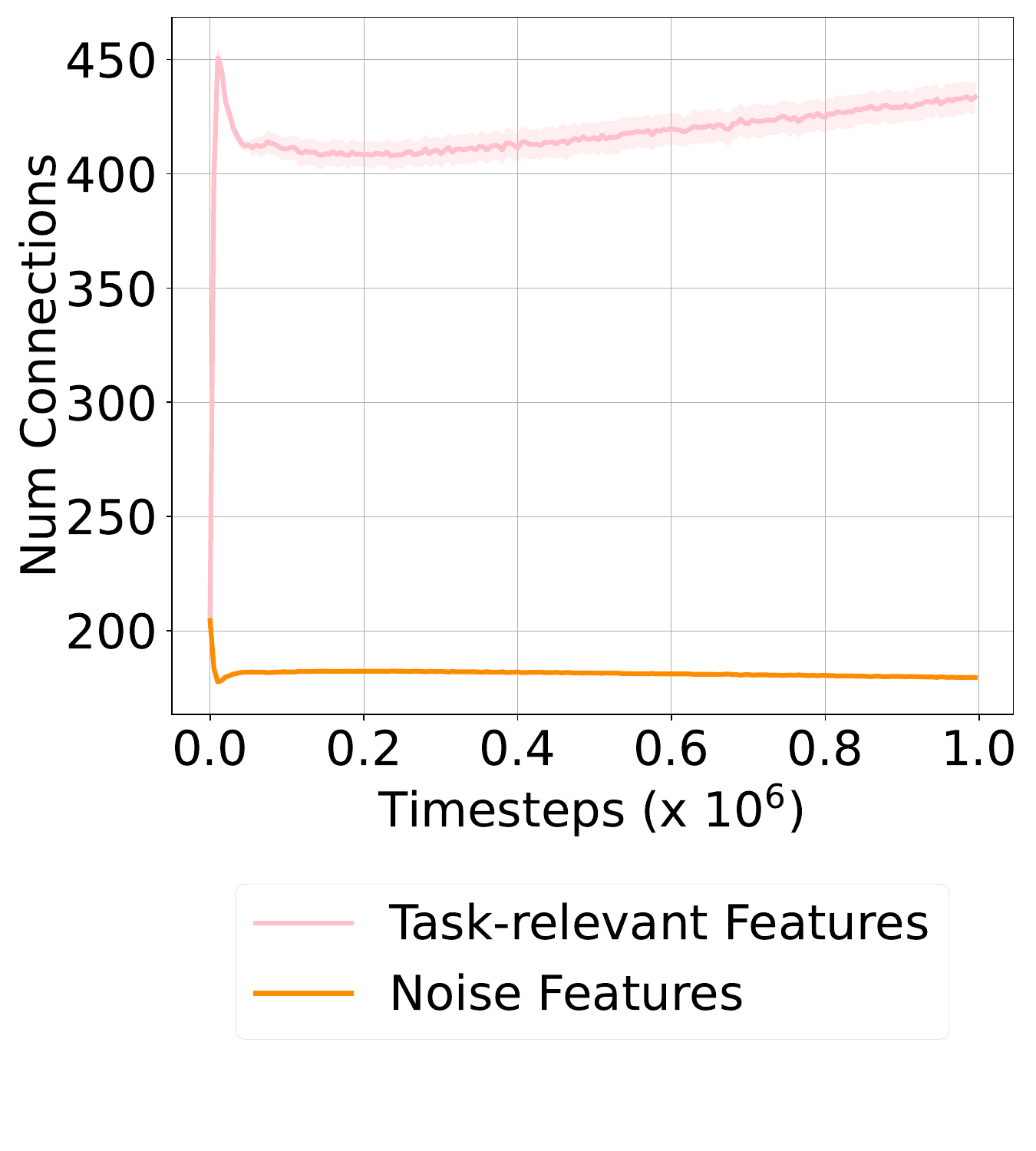}\label{fig:cheetah_run_actor}}
  \hfill
\subfloat[Critic Network 1]{\includegraphics[width=0.33\textwidth]{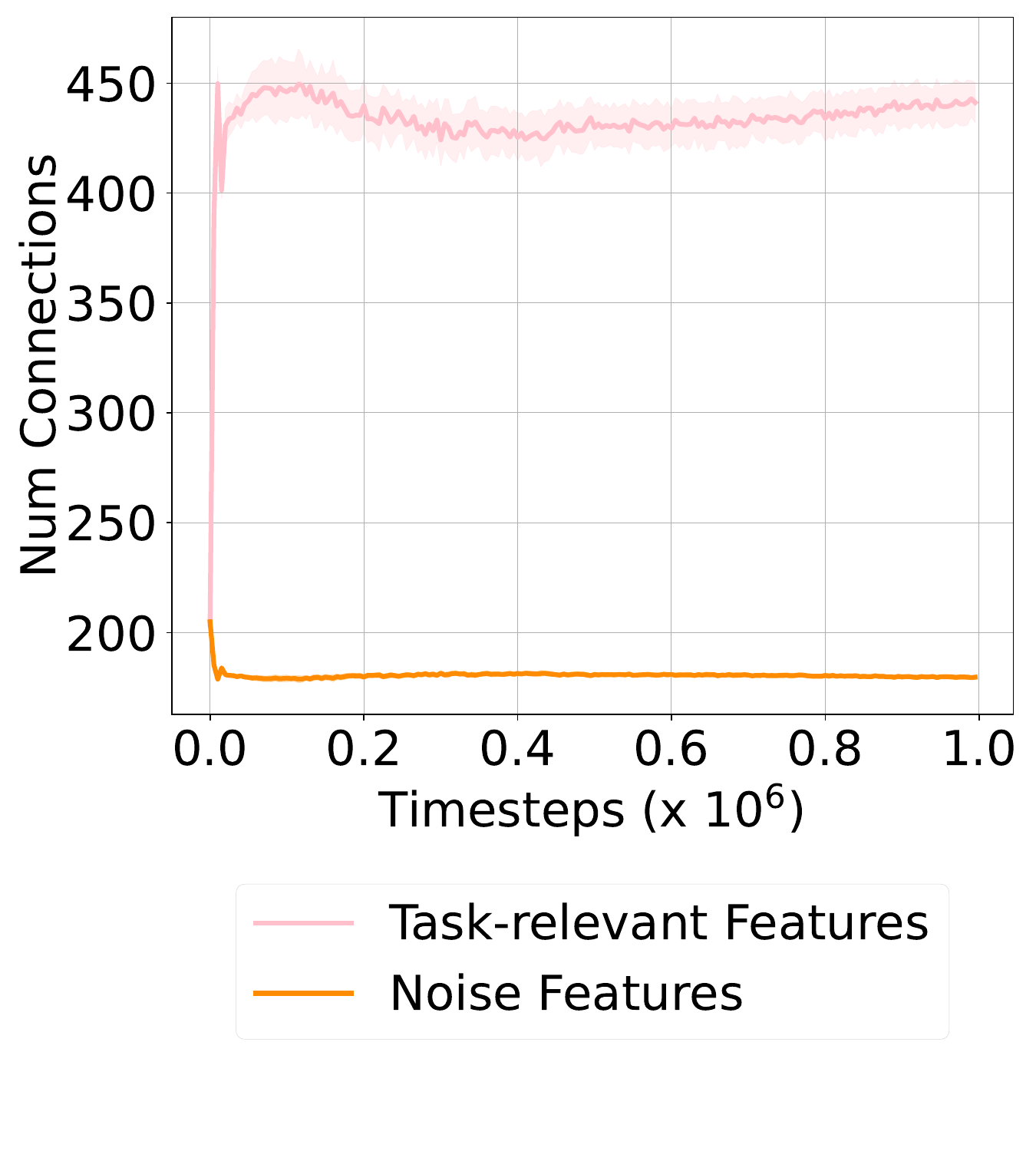}\label{fig:cheetah_run_q0}}
  \hfill
\subfloat[Critic Network 2]
{\includegraphics[width=0.33\textwidth]{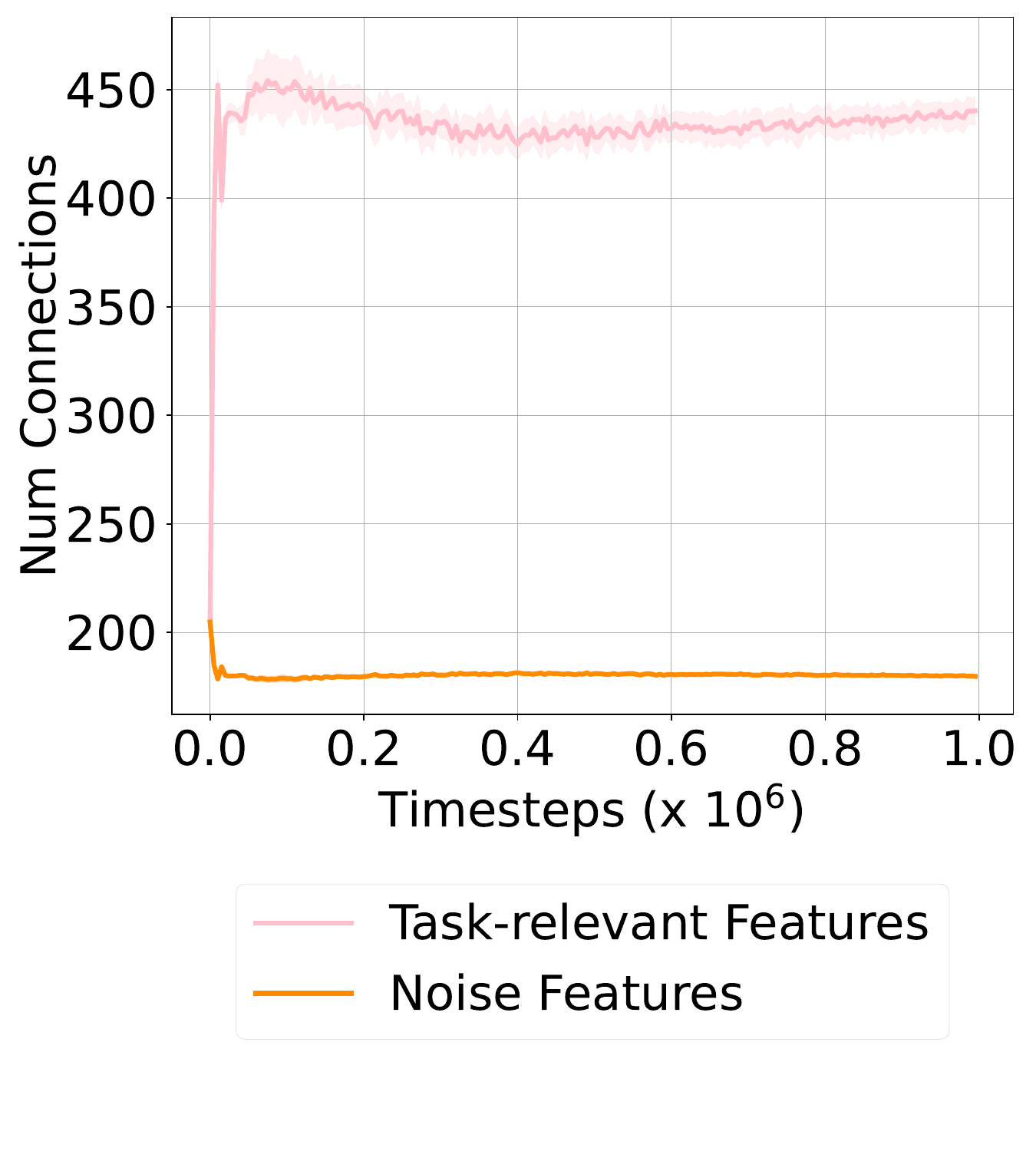}}\label{fig:cheetah_run_q1}
  \hfill
  \subfloat[Reward Model 1]{\includegraphics[width=0.33\textwidth]{figures/appendix/real_v_noise_feats/cheetah_run_90_1000_r0.pdf}\label{fig:cheetah_run_r1}}
  \hfill
\subfloat[Reward Model 2]{\includegraphics[width=0.33\textwidth]{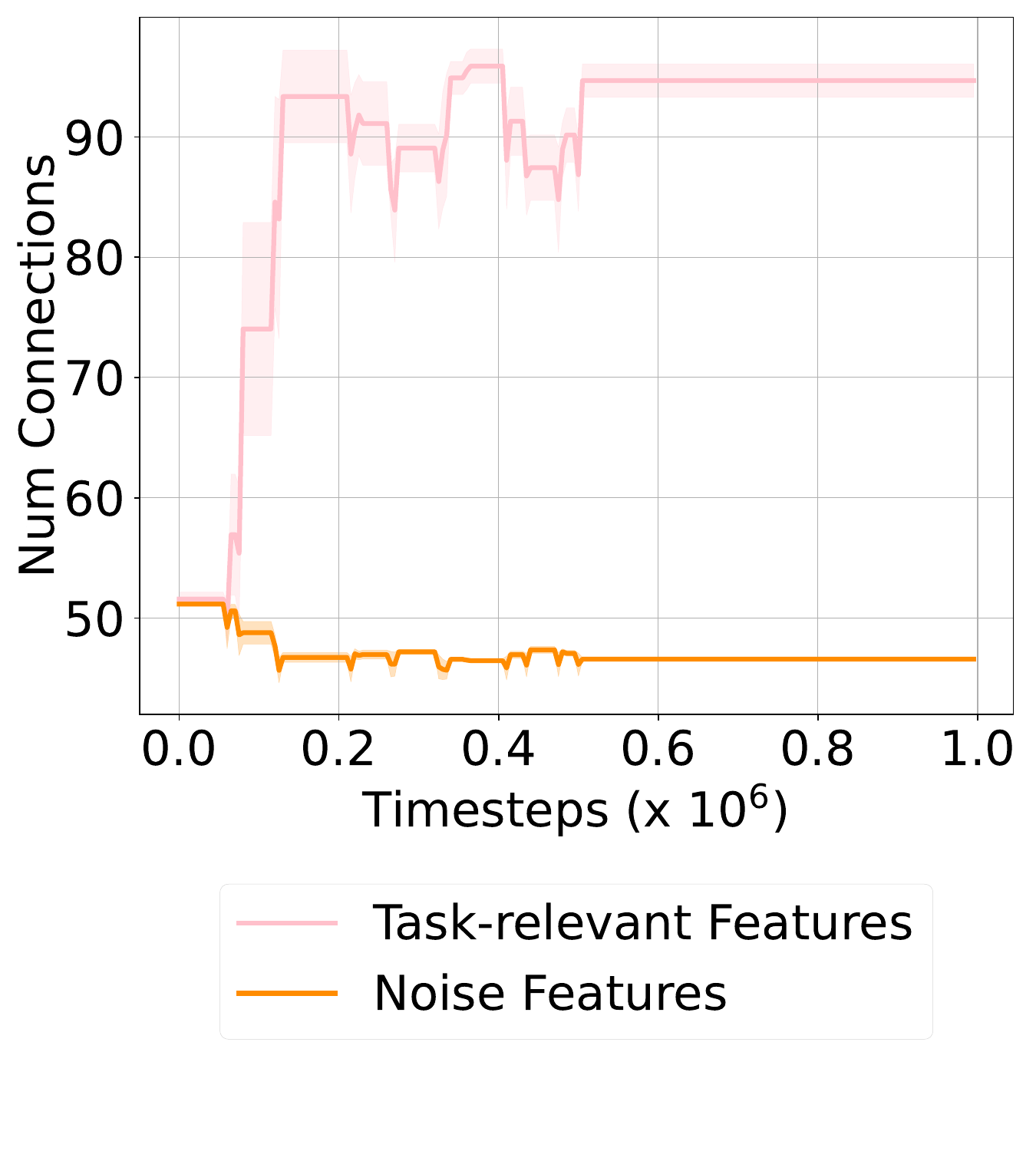}\label{fig:cheetah_run_r2}}
  \hfill
\subfloat[Reward Model 3]
{\includegraphics[width=0.33\textwidth]{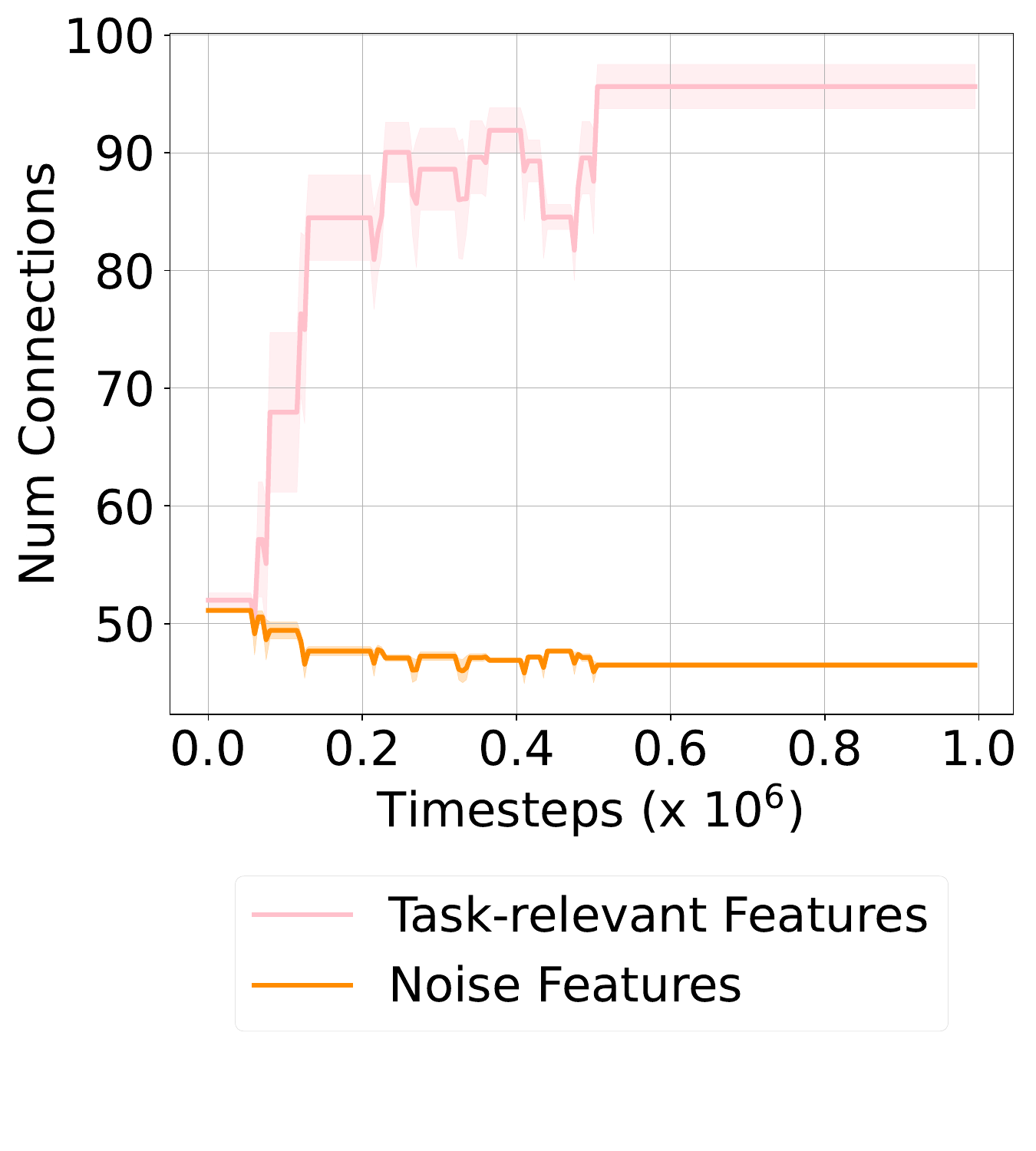}}\label{fig:cheetah_run_r3}
  \hfill
  \caption{Cheetah-run. These plots show the average number of neural network connections to the relevant versus noise features in R2N's RL networks and reward models.}
  \label{fig:real_v_noise_feats_rl_cheetah_run}
\end{figure*}

\begin{figure*}[h!]
  \centering
 \subfloat[Actor Network]{\includegraphics[width=0.33\textwidth]{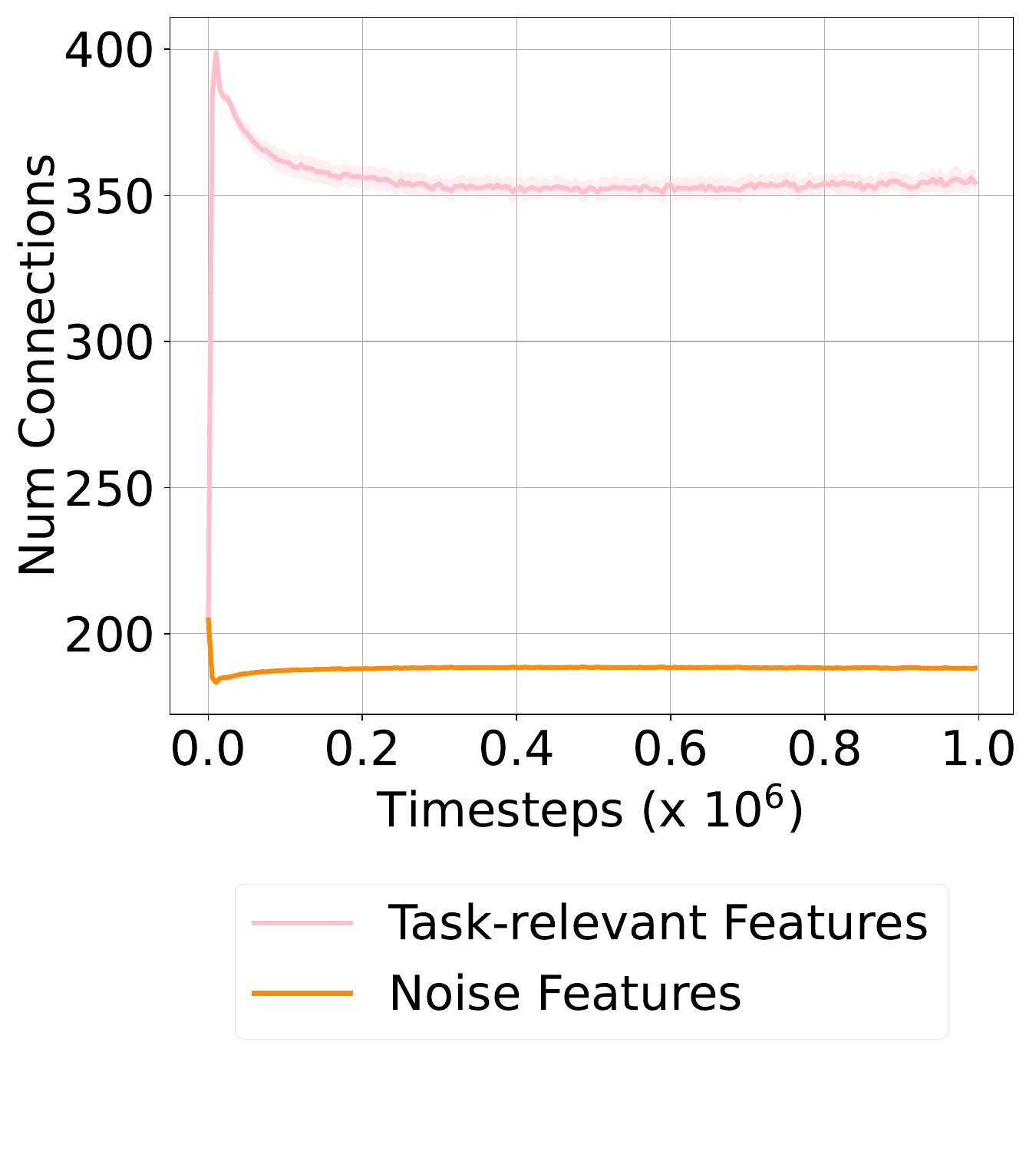}\label{fig:walker_walk_actor}}
  \hfill
\subfloat[Critic Network 1]{\includegraphics[width=0.33\textwidth]{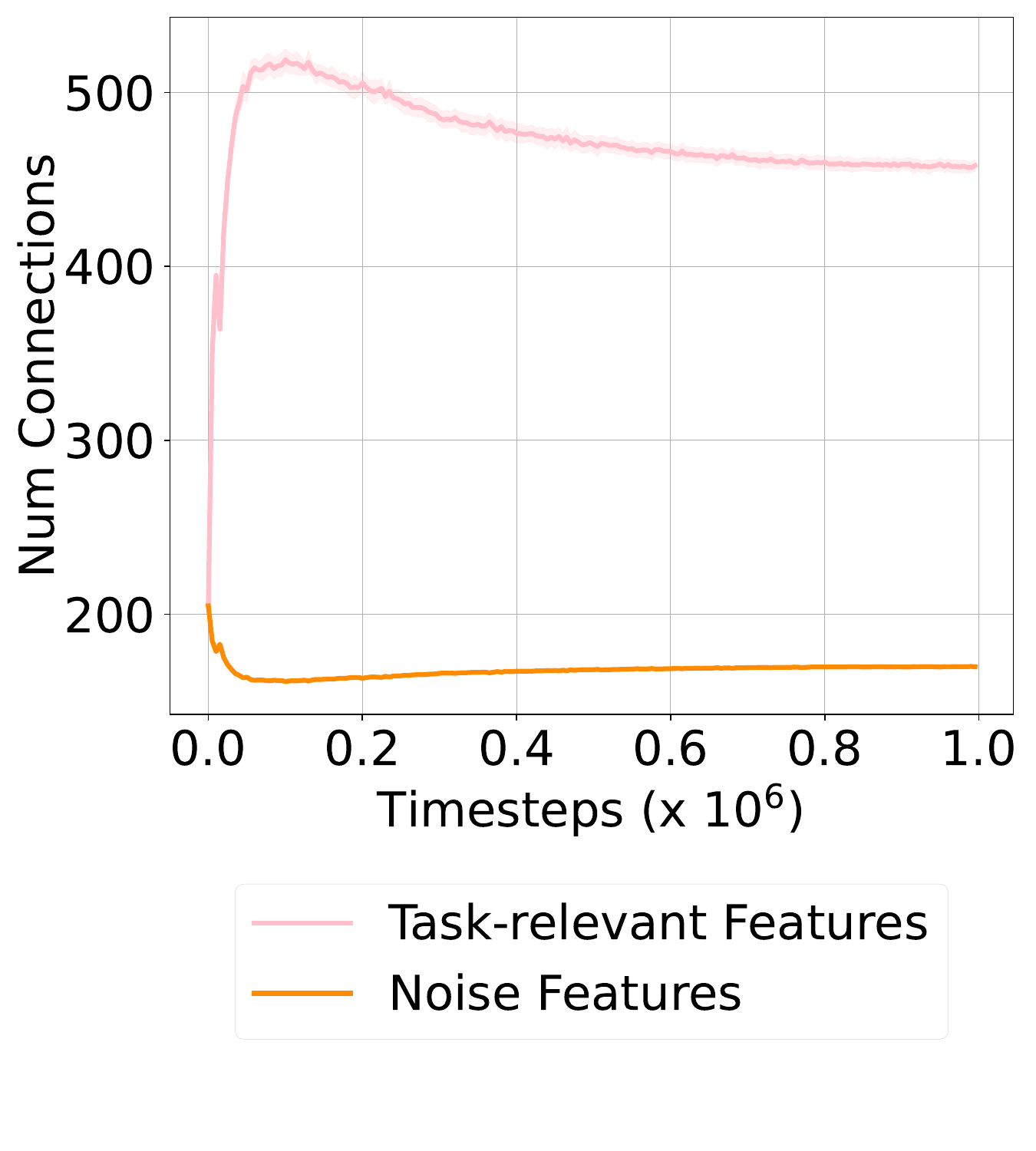}\label{fig:walker_walk_q0}}
  \hfill
\subfloat[Critic Network 2]
{\includegraphics[width=0.33\textwidth]{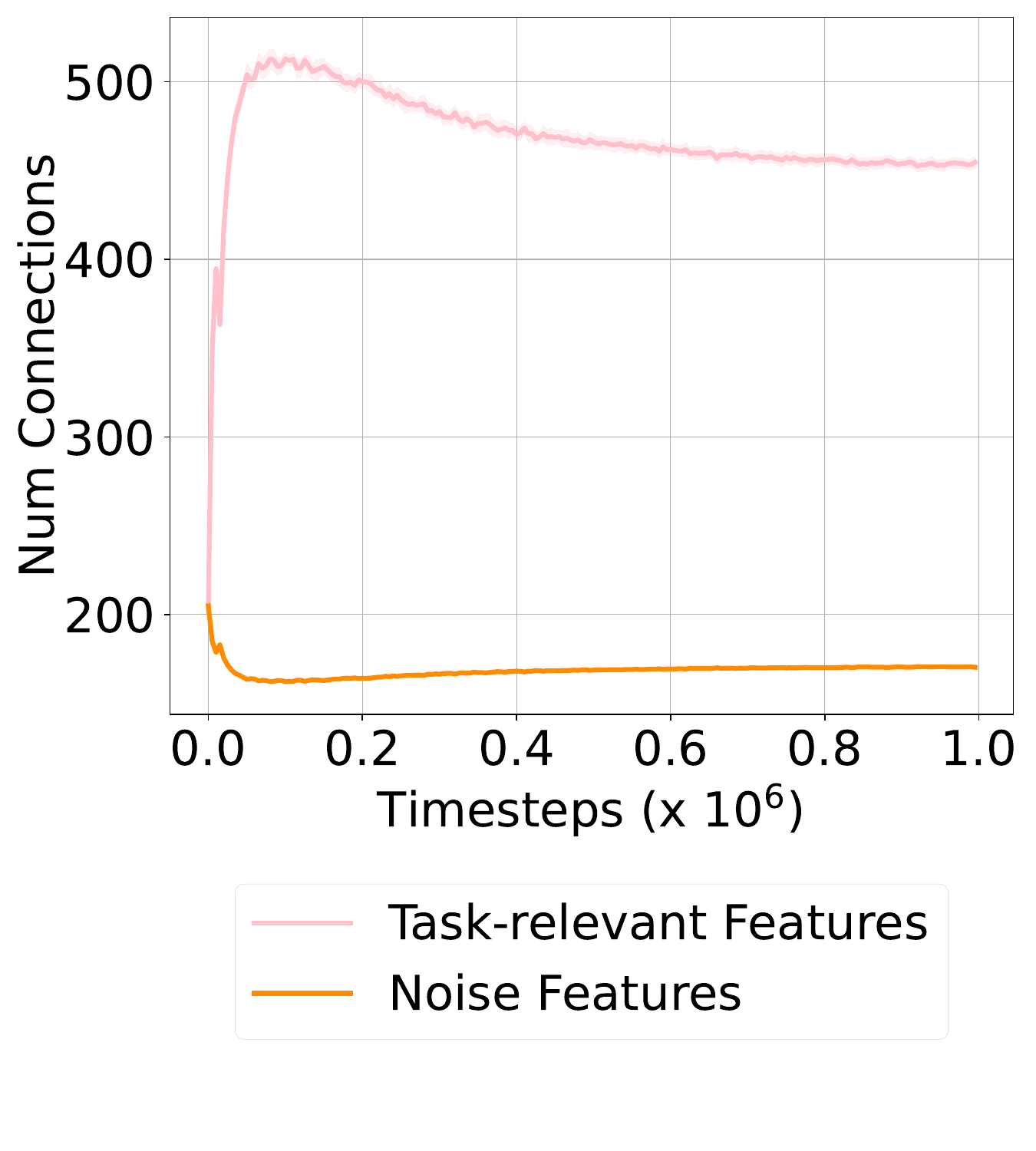}}\label{fig:walker_walk_q1}
  \hfill
  \subfloat[Reward Model 1]{\includegraphics[width=0.33\textwidth]{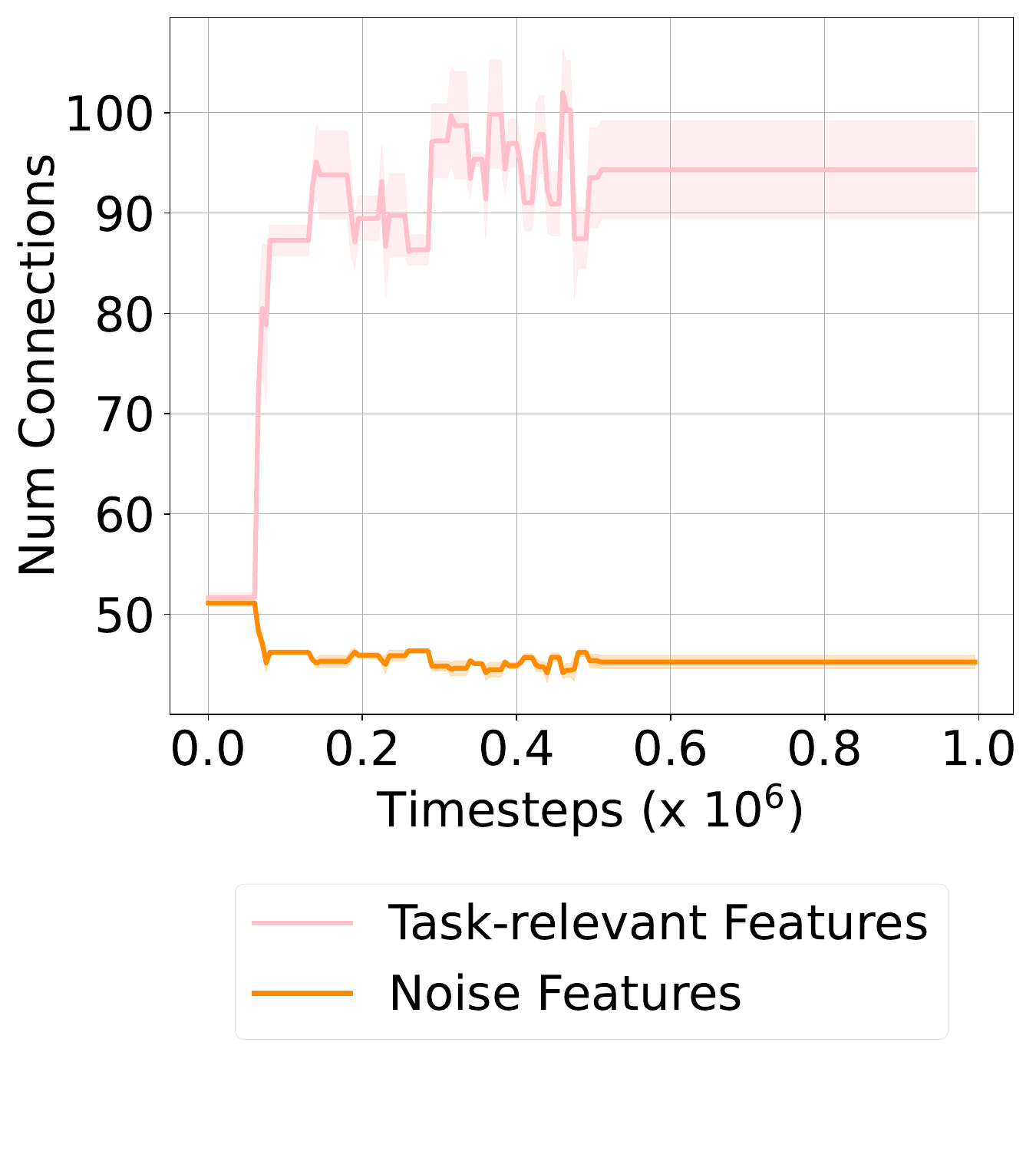}\label{fig:walker_walk_r1}}
  \hfill
\subfloat[Reward Model 2]{\includegraphics[width=0.33\textwidth]{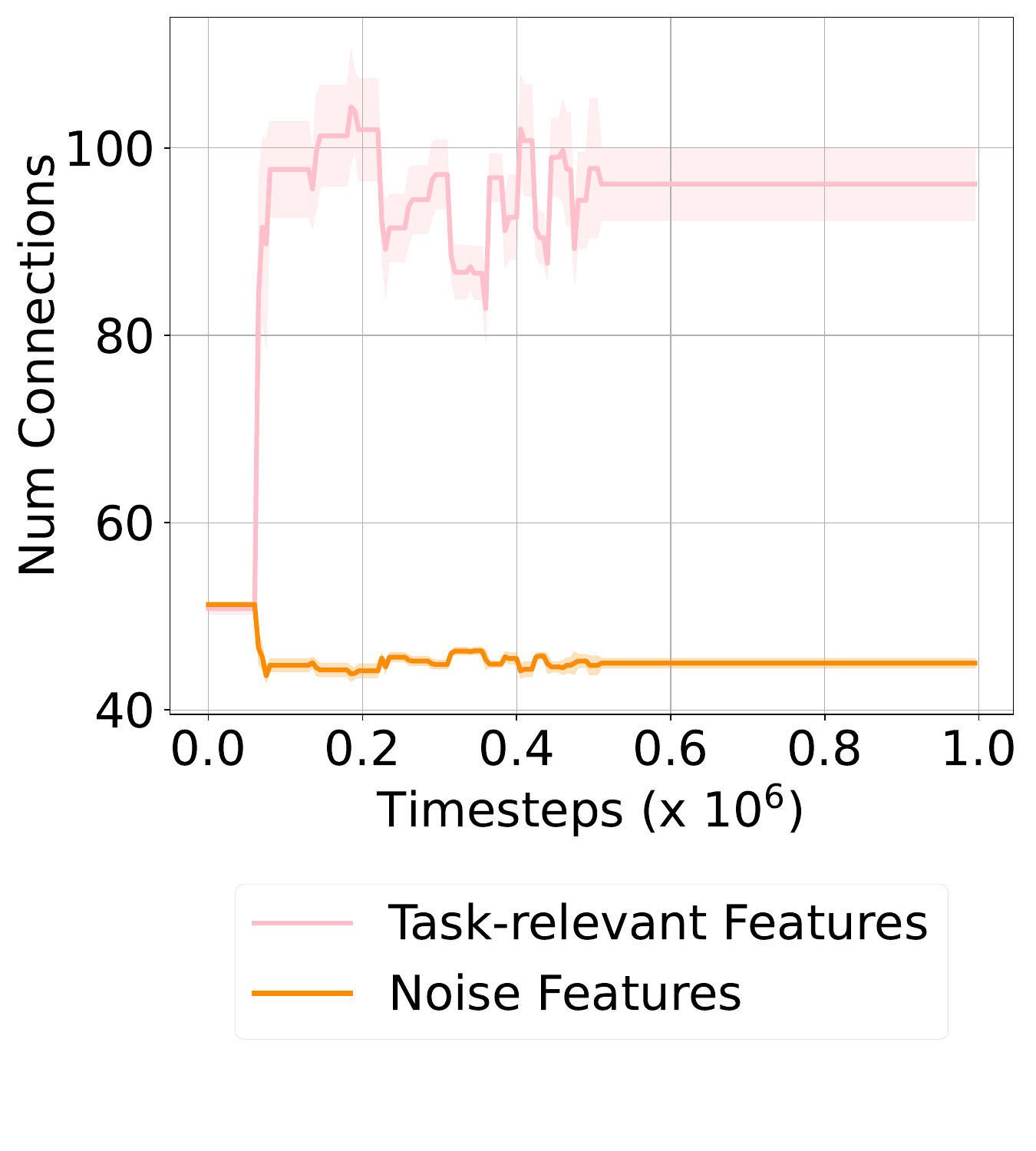}\label{fig:walker_walk_r2}}
  \hfill
\subfloat[Reward Model 3]
{\includegraphics[width=0.33\textwidth]{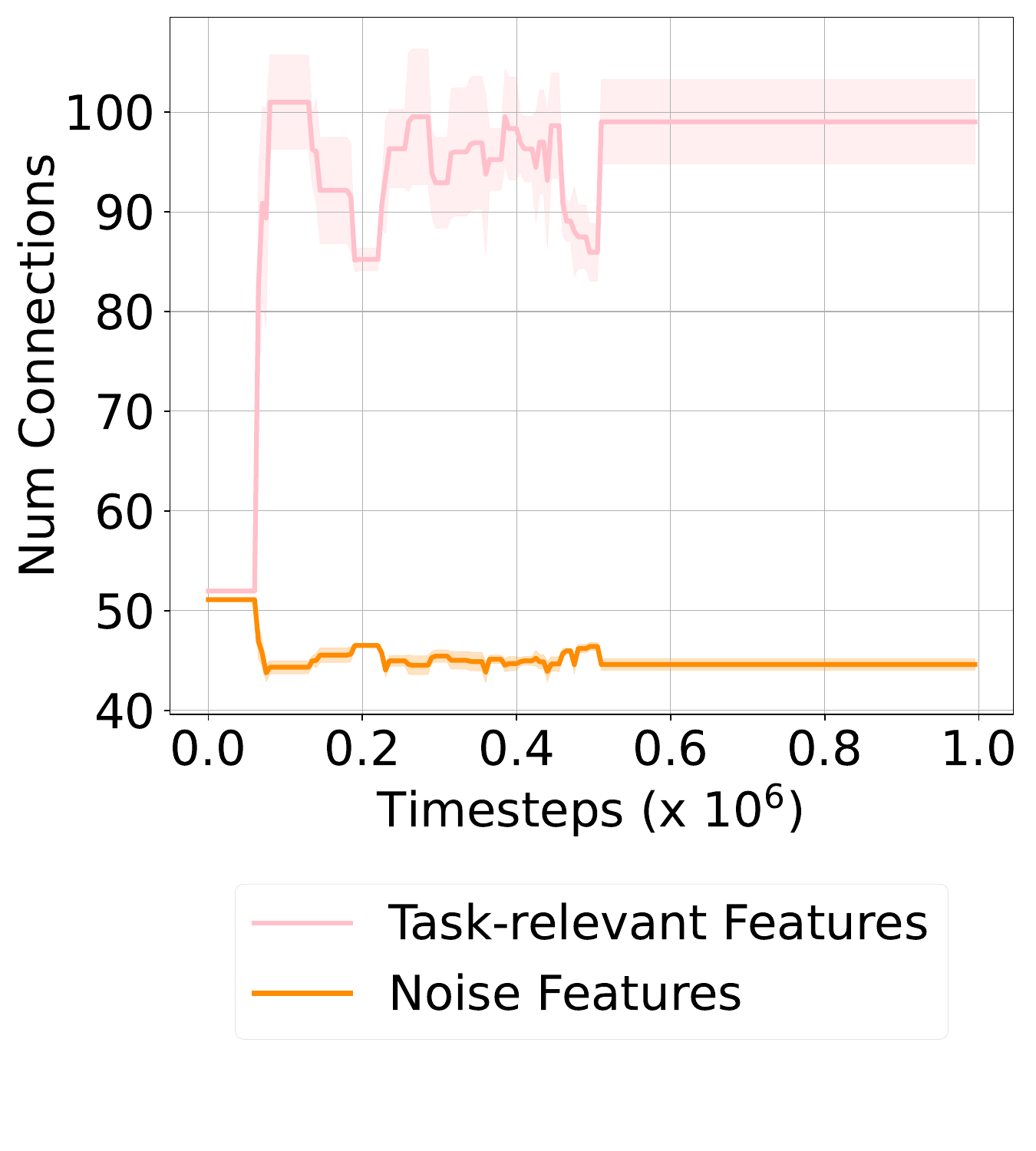}}\label{fig:walker_walk_r3}
  \hfill
  \caption{Walker-walk. These plots show the average number of neural network connections to the relevant versus noise features in R2N's RL networks and reward models.}
  \label{fig:real_v_noise_feats_rl_walker_walk}
\end{figure*}

In this section, we show further analysis of the neural network connections in R2N-PEBBLE for the Cheetah-run experiment (90\% noise features and a feedback budget of 1000) and Walker-walk experiment (90\% noise features and a feedback budget of 4000). More specifically, 
Figures \ref{fig:real_v_noise_feats_rl_cheetah_run} and \ref{fig:real_v_noise_feats_rl_walker_walk} highlight that with R2N, the RL agent and the reward models can significantly increase the number of connections to the task-relevant features (pink curves) compared with the number of noise features (orange curves). However, note that the reward models are trained until timestep $500000$, so the connectivity does not alter after that point.

\subsection{Feedback and Noise Ablations in Cheetah-run}\label{sec:fb_noise_ablations_cheetah_run}

\begin{figure*}[h!]
  \centering
 \subfloat[Feedback = 100]{\includegraphics[width=0.33\textwidth]{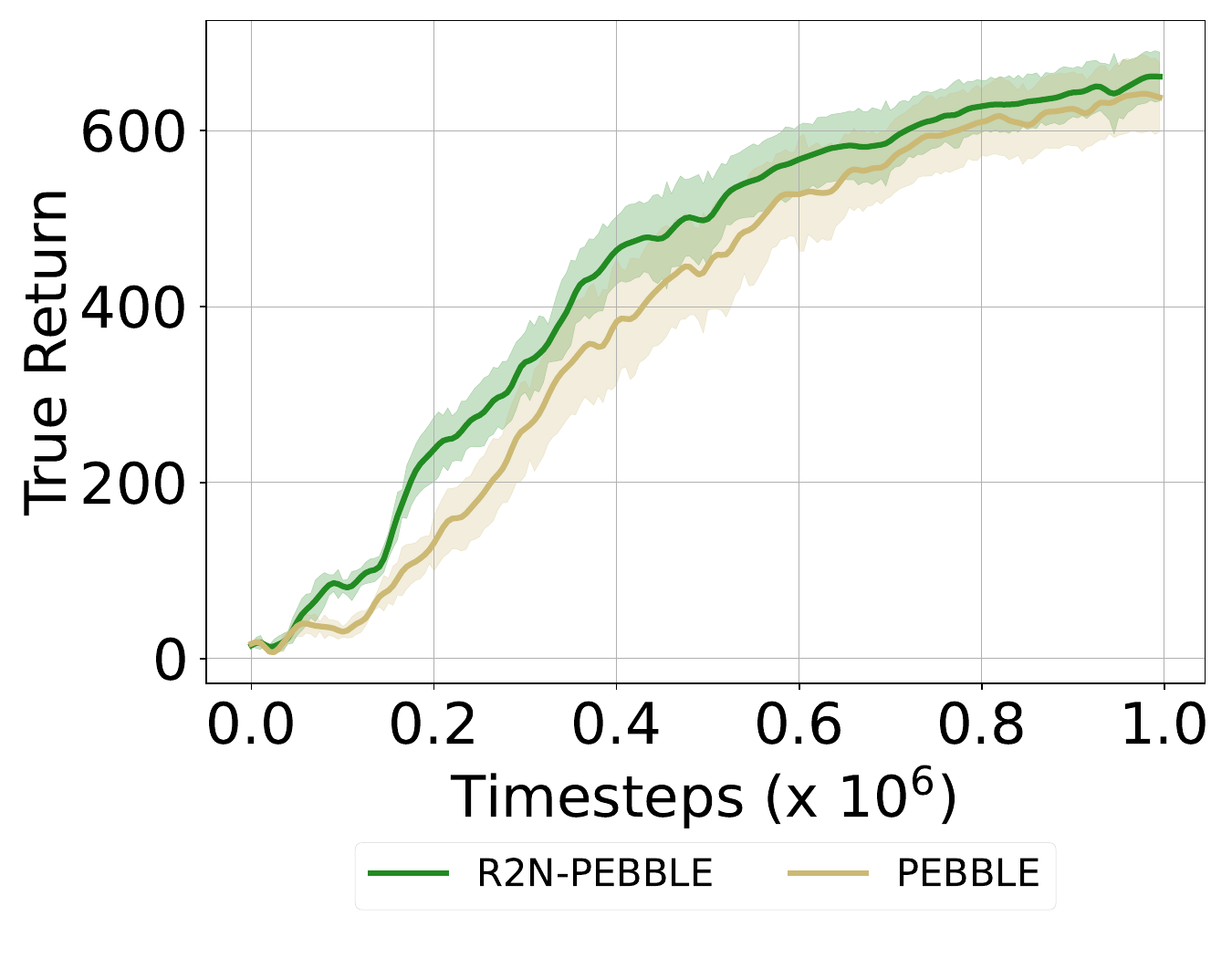}\label{fig:cheetah_run_20_100}}
  \hfill
\subfloat[Feedback = 200]{\includegraphics[width=0.33\textwidth]{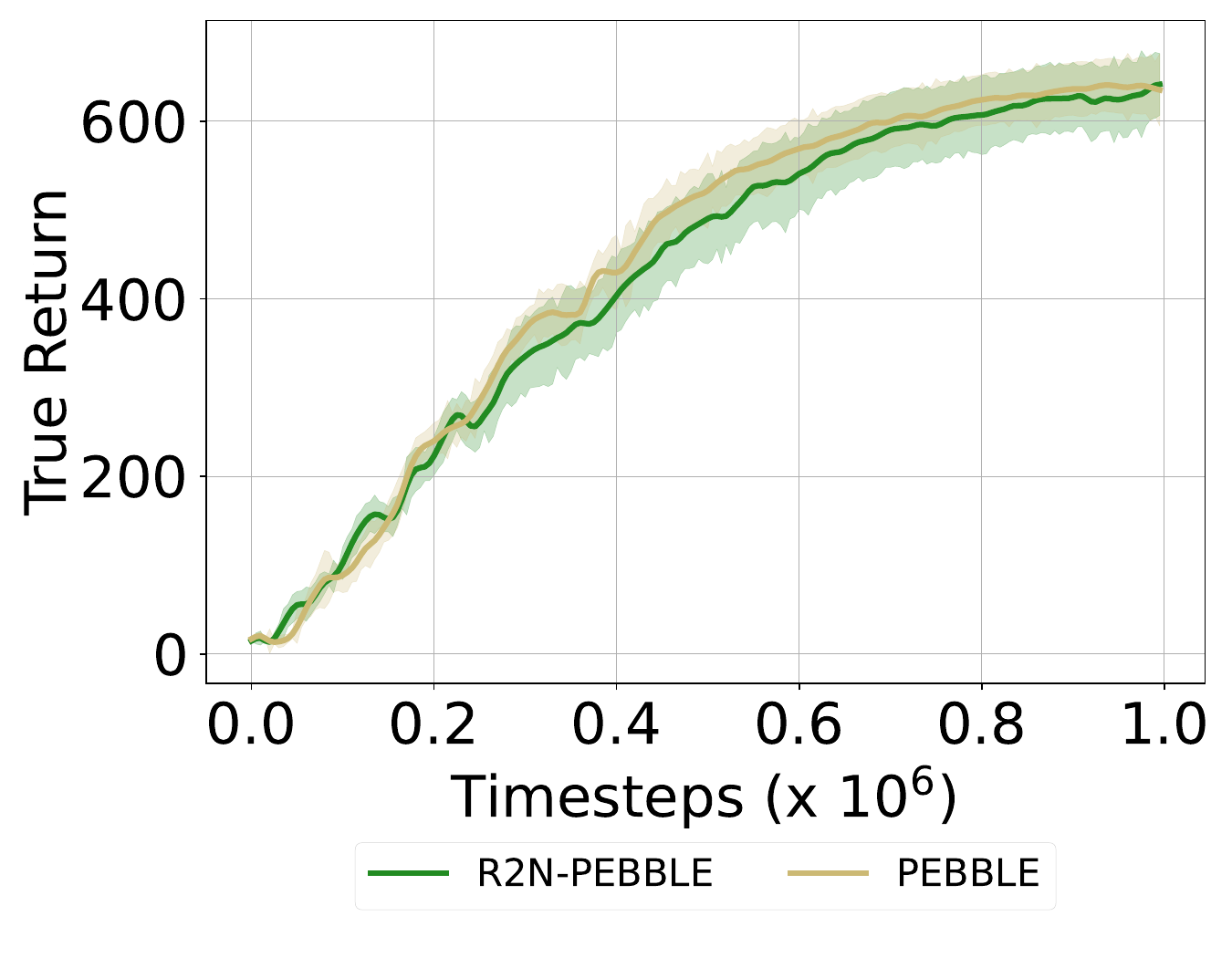}\label{fig:cheetah_run_20_200}}
  \hfill
\subfloat[Feedback = 400]
{\includegraphics[width=0.33\textwidth]{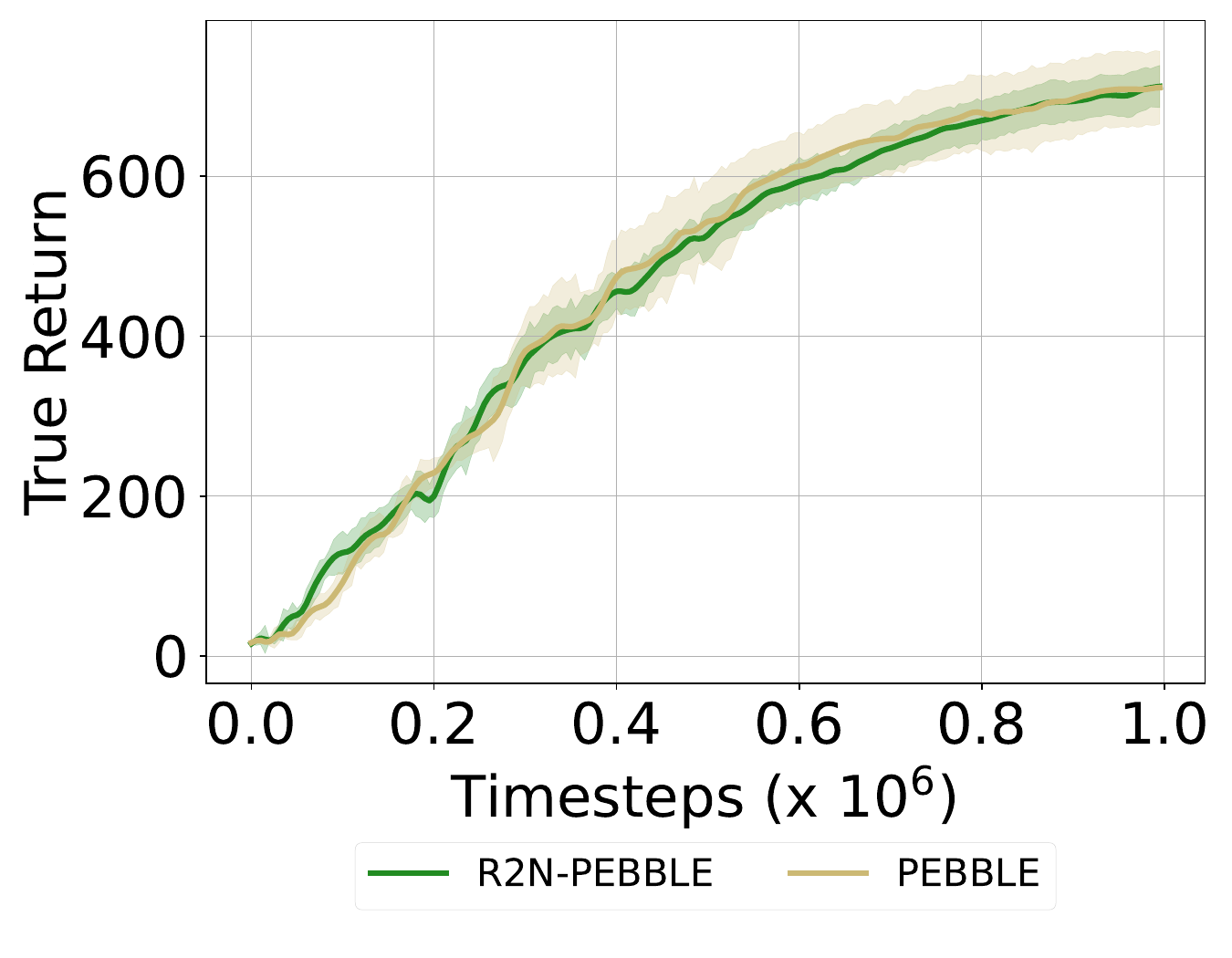}\label{fig:cheetah_run_20_400}}
  \hfill
  \subfloat[Feedback = 1000]{\includegraphics[width=0.33\textwidth]{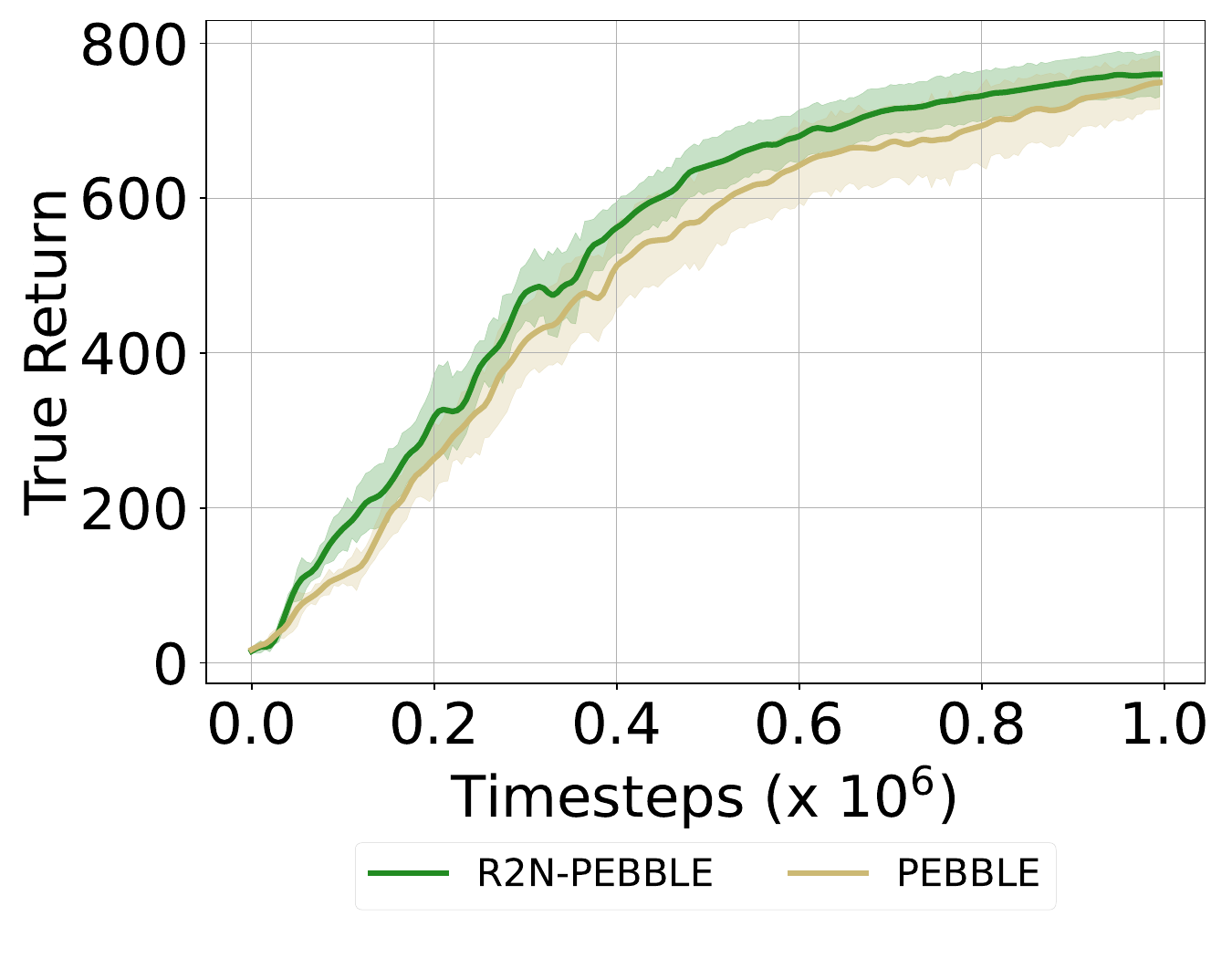}\label{fig:cheetah_run_20_1k}}
    \hfill
  \subfloat[Feedback = 2000]{\includegraphics[width=0.33\textwidth]{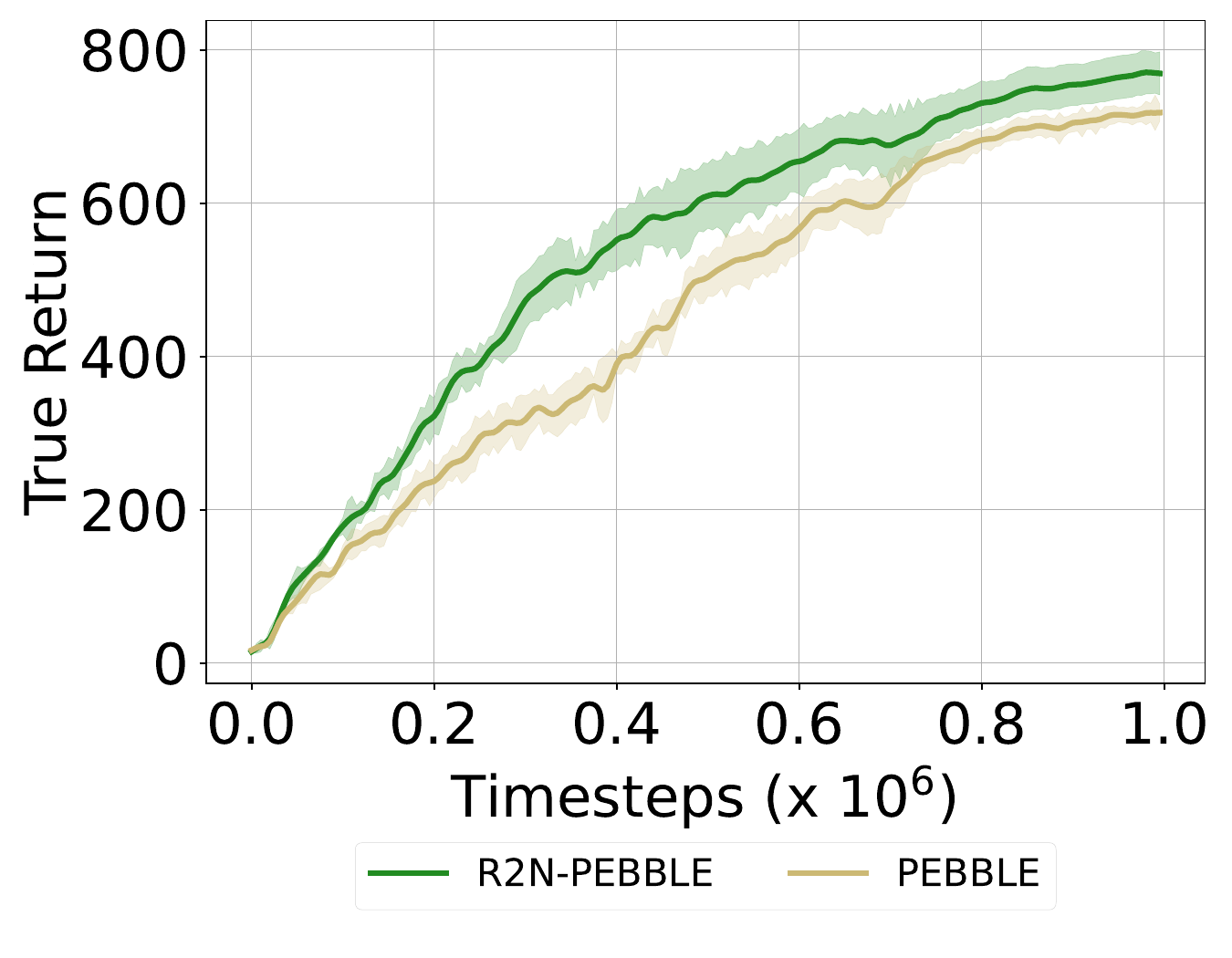}\label{fig:cheetah_run_20_2k}}
    \hfill
  \subfloat[Feedback = 4000]{\includegraphics[width=0.33\textwidth]{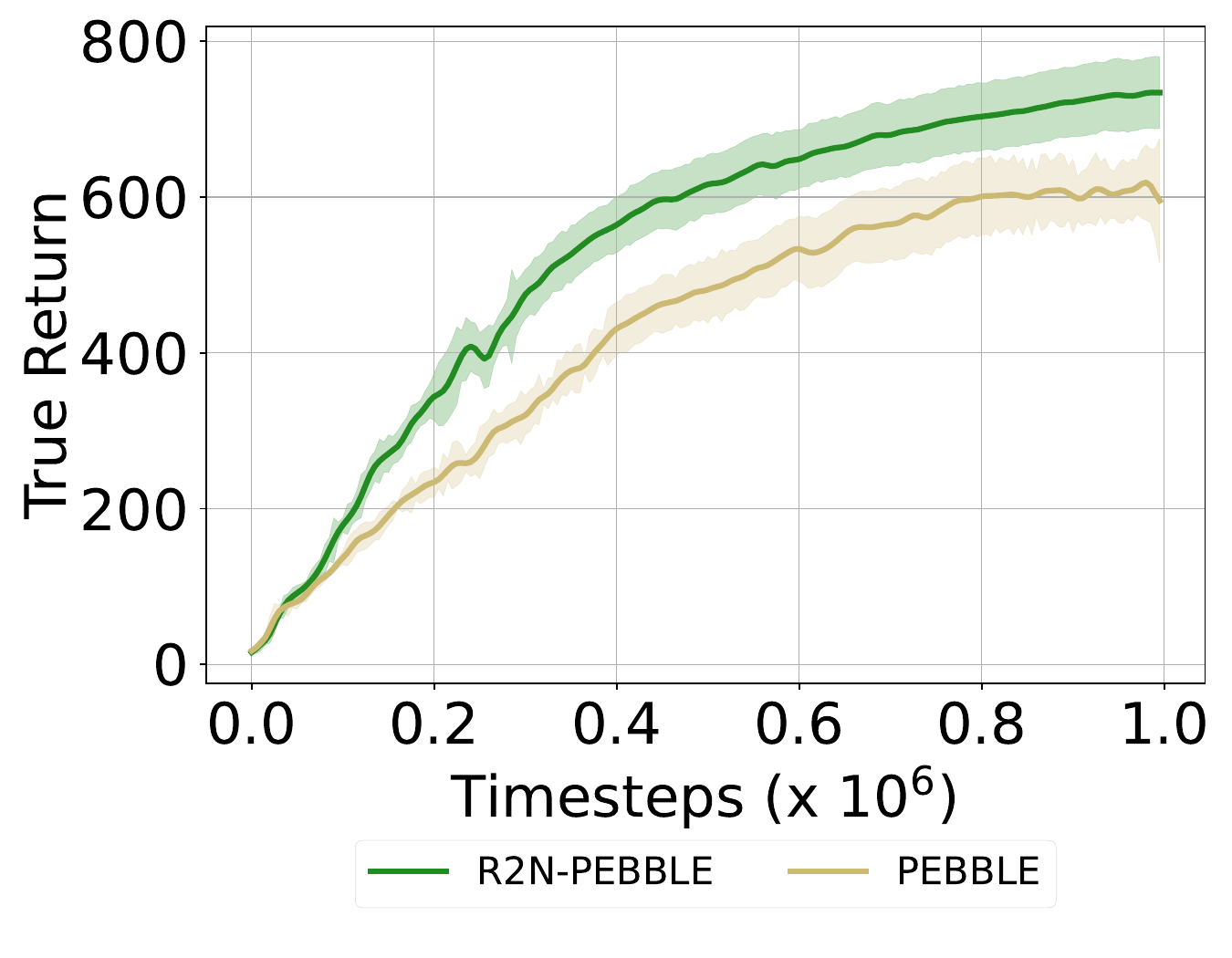}\label{fig:cheetah_run_20_4k}}
    \hfill
  \subfloat[Feedback = 10000]{\includegraphics[width=0.33\textwidth]{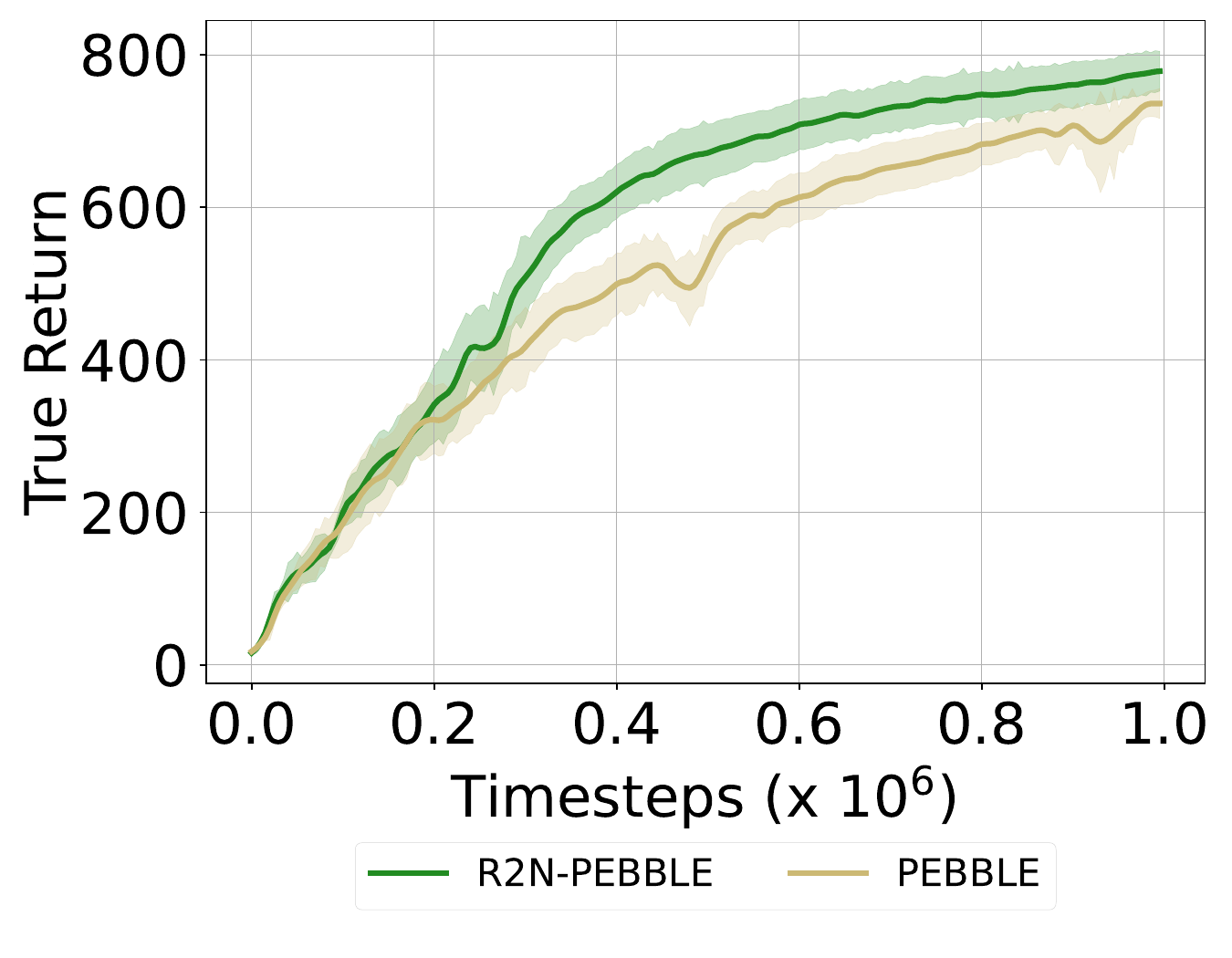}\label{fig:cheetah_run_20_10k}}
   
  \caption{Cheetah-run, Feedback Ablation, Noise = 20\%}\label{fig:cheetah_run_fb_ablation_noise20}
\end{figure*}

\begin{figure*}[h!]
  \centering
 \subfloat[Feedback = 100]{\includegraphics[width=0.33\textwidth]{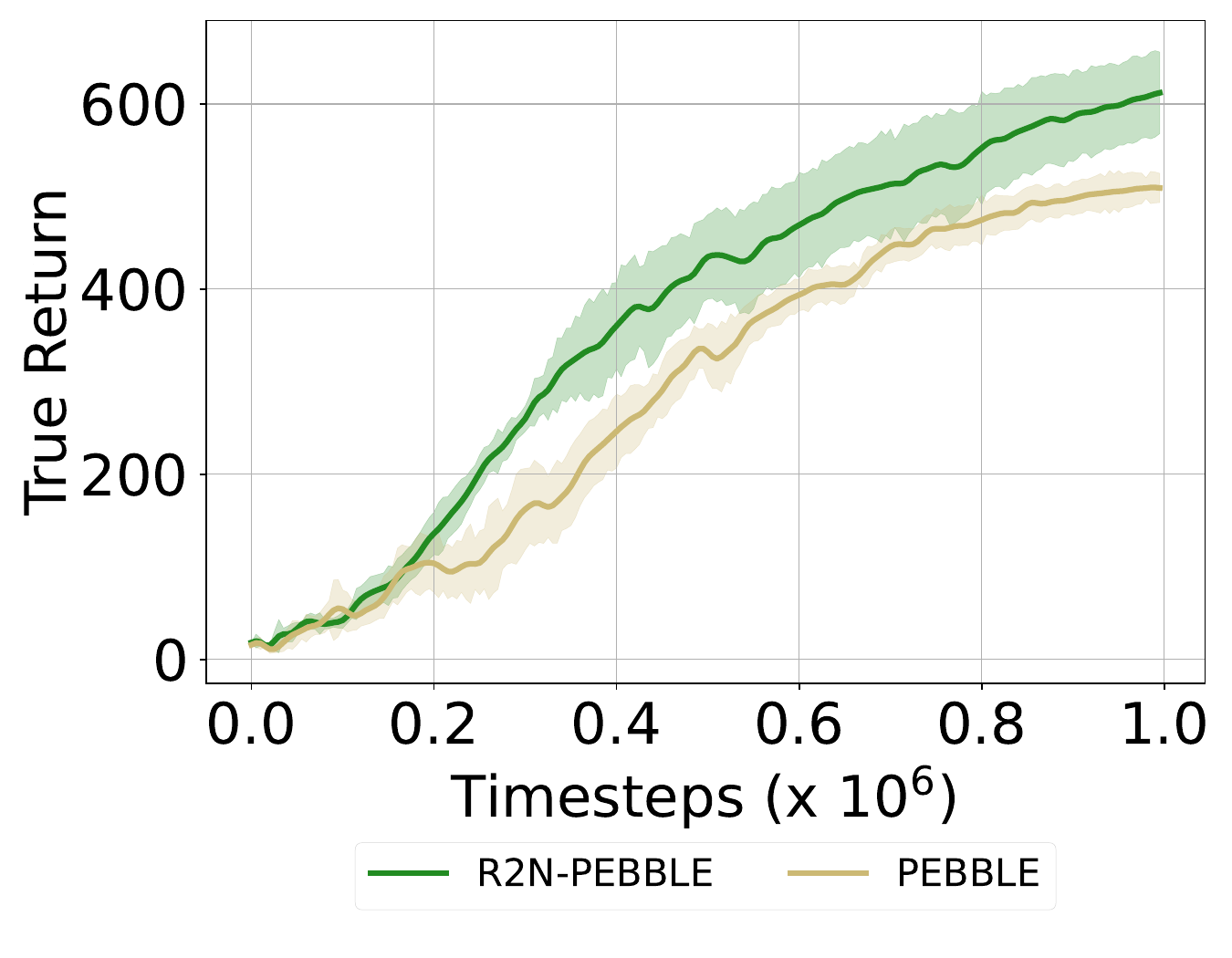}\label{fig:cheetah_run_50_100}}
  \hfill
\subfloat[Feedback = 200]{\includegraphics[width=0.33\textwidth]{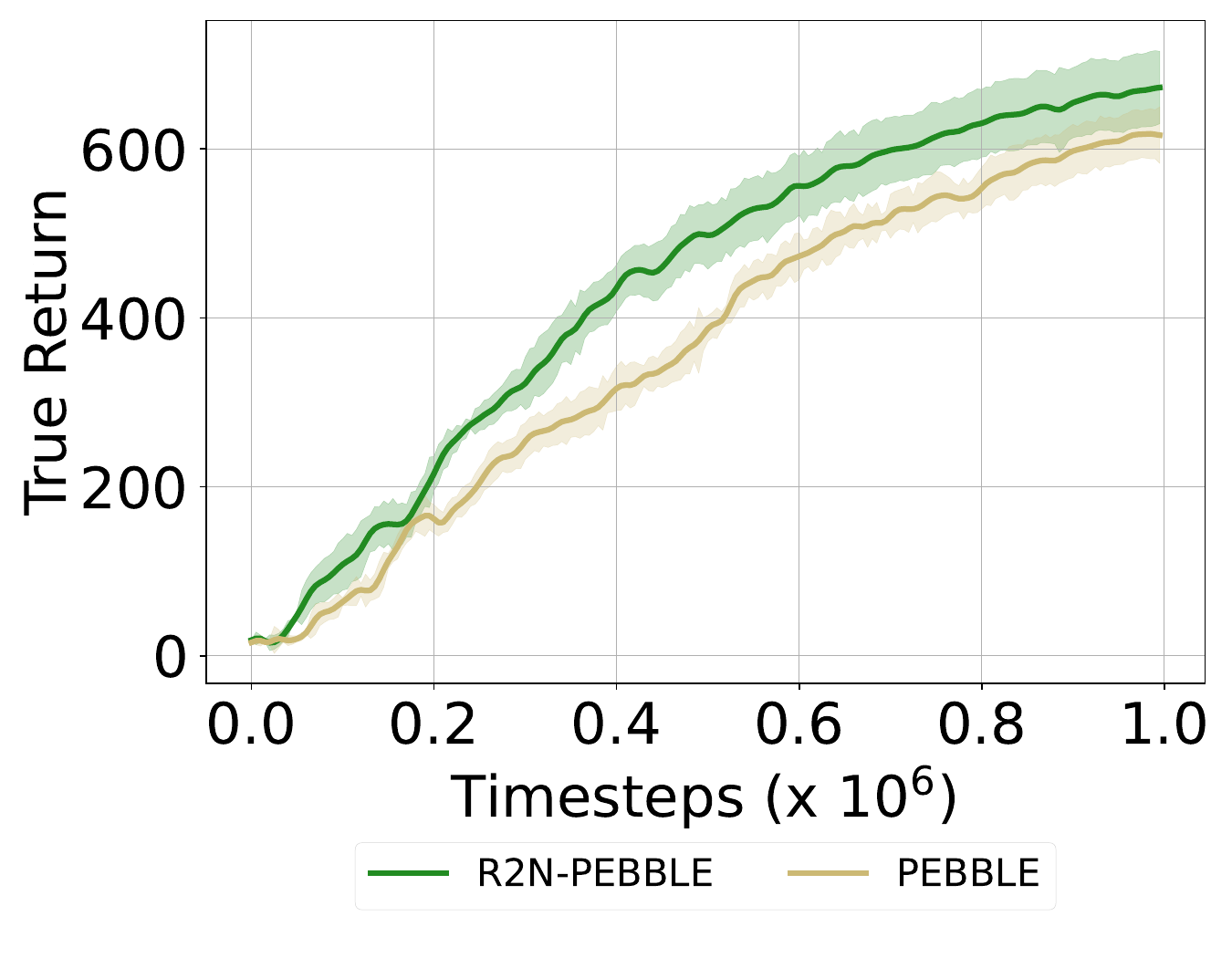}\label{fig:cheetah_run_50_200}}
  \hfill
\subfloat[Feedback = 400]
{\includegraphics[width=0.33\textwidth]{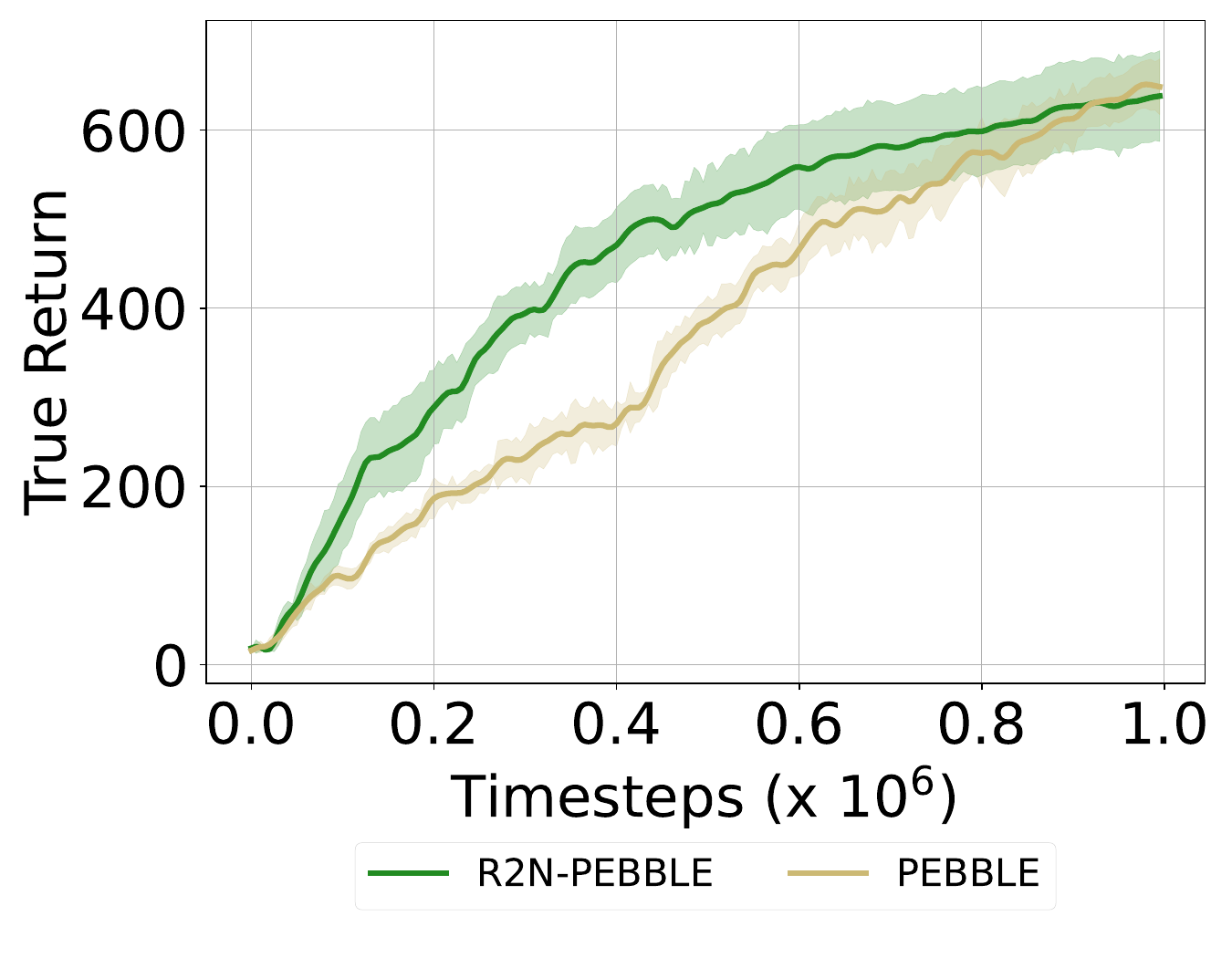}\label{fig:cheetah_run_50_400}}
  \hfill
  \subfloat[Feedback = 1000]{\includegraphics[width=0.33\textwidth]{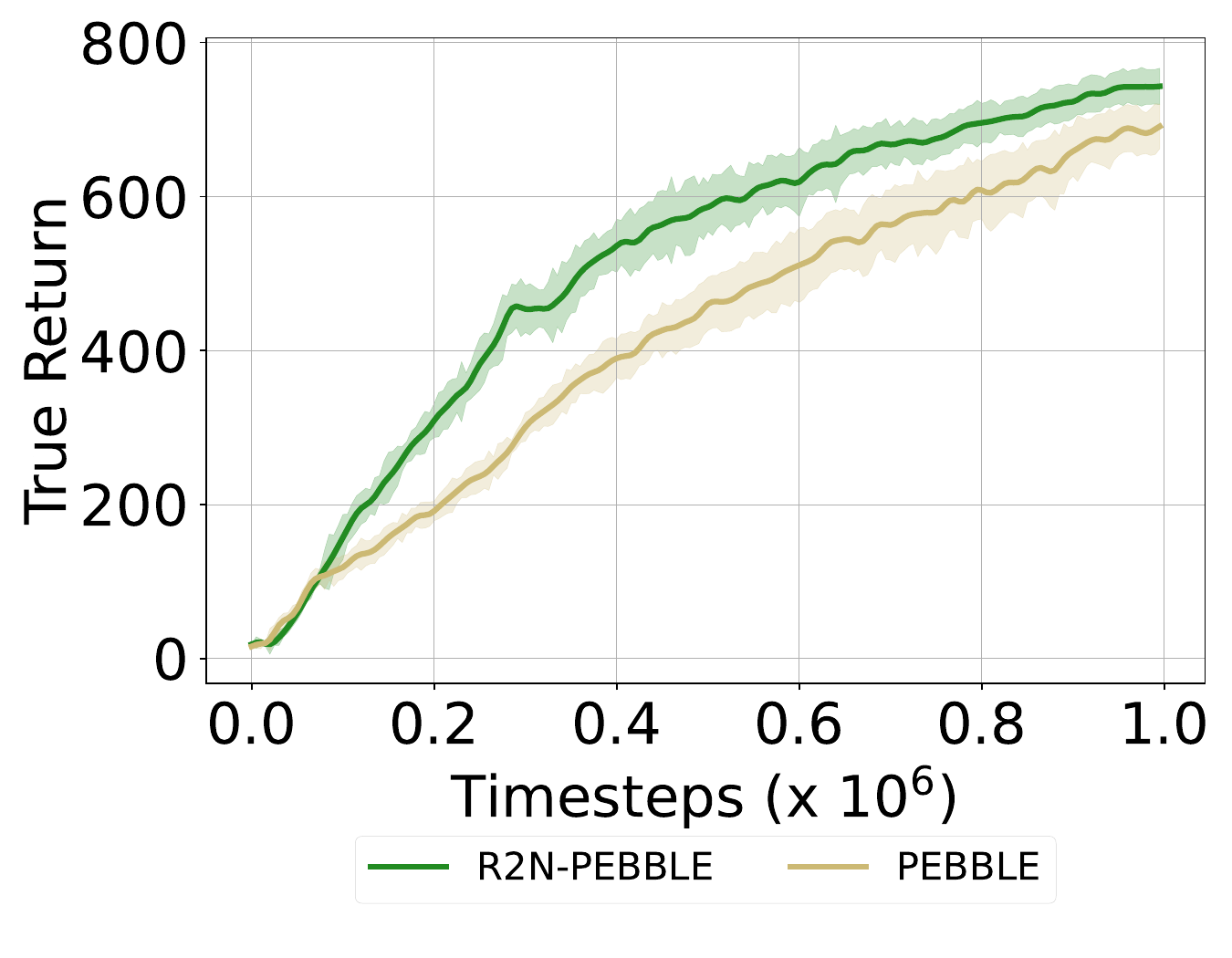}\label{fig:cheetah_run_50_1k}}
    \hfill
  \subfloat[Feedback = 2000]{\includegraphics[width=0.33\textwidth]{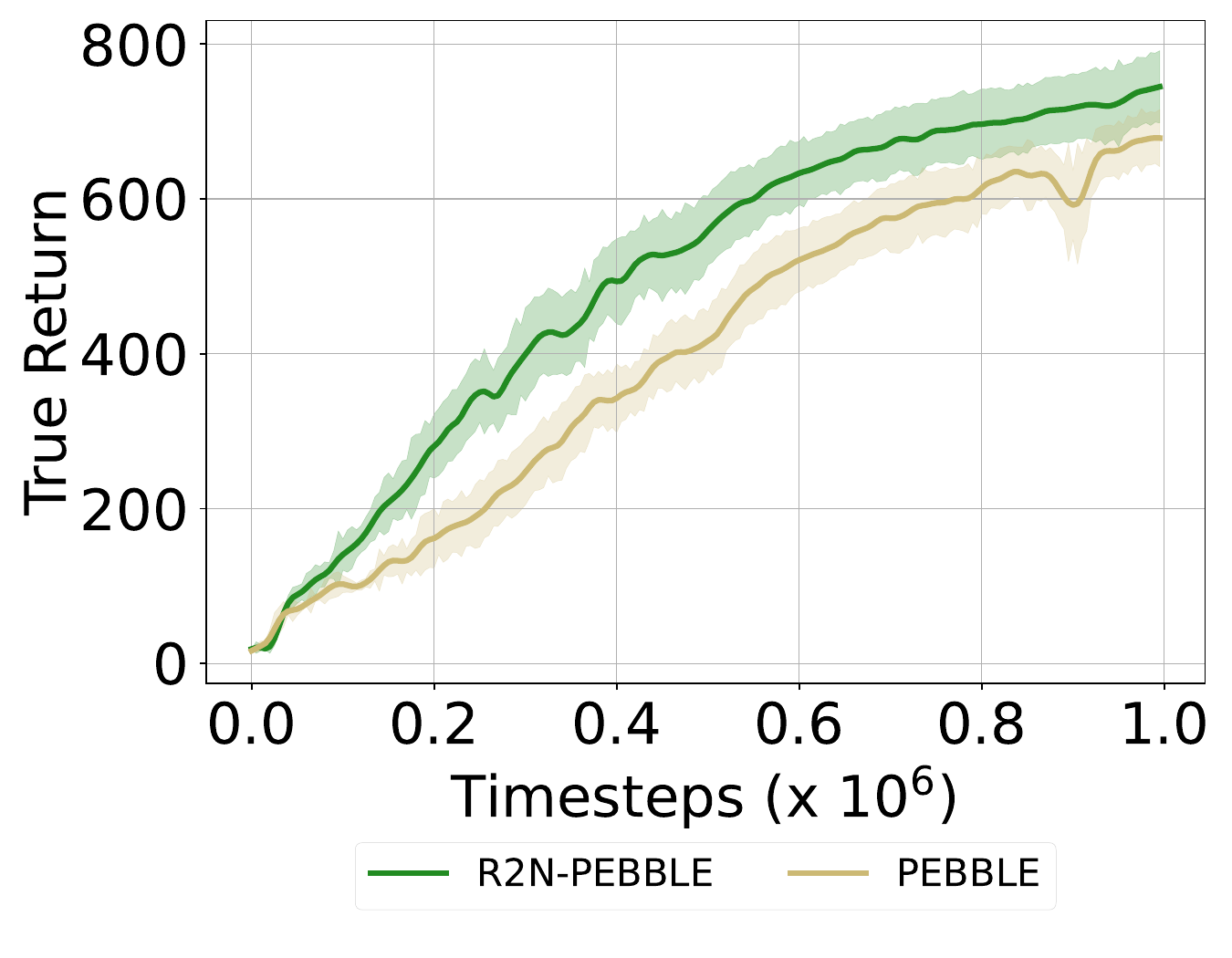}\label{fig:cheetah_run_50_2k}}
    \hfill
  \subfloat[Feedback = 4000]{\includegraphics[width=0.33\textwidth]{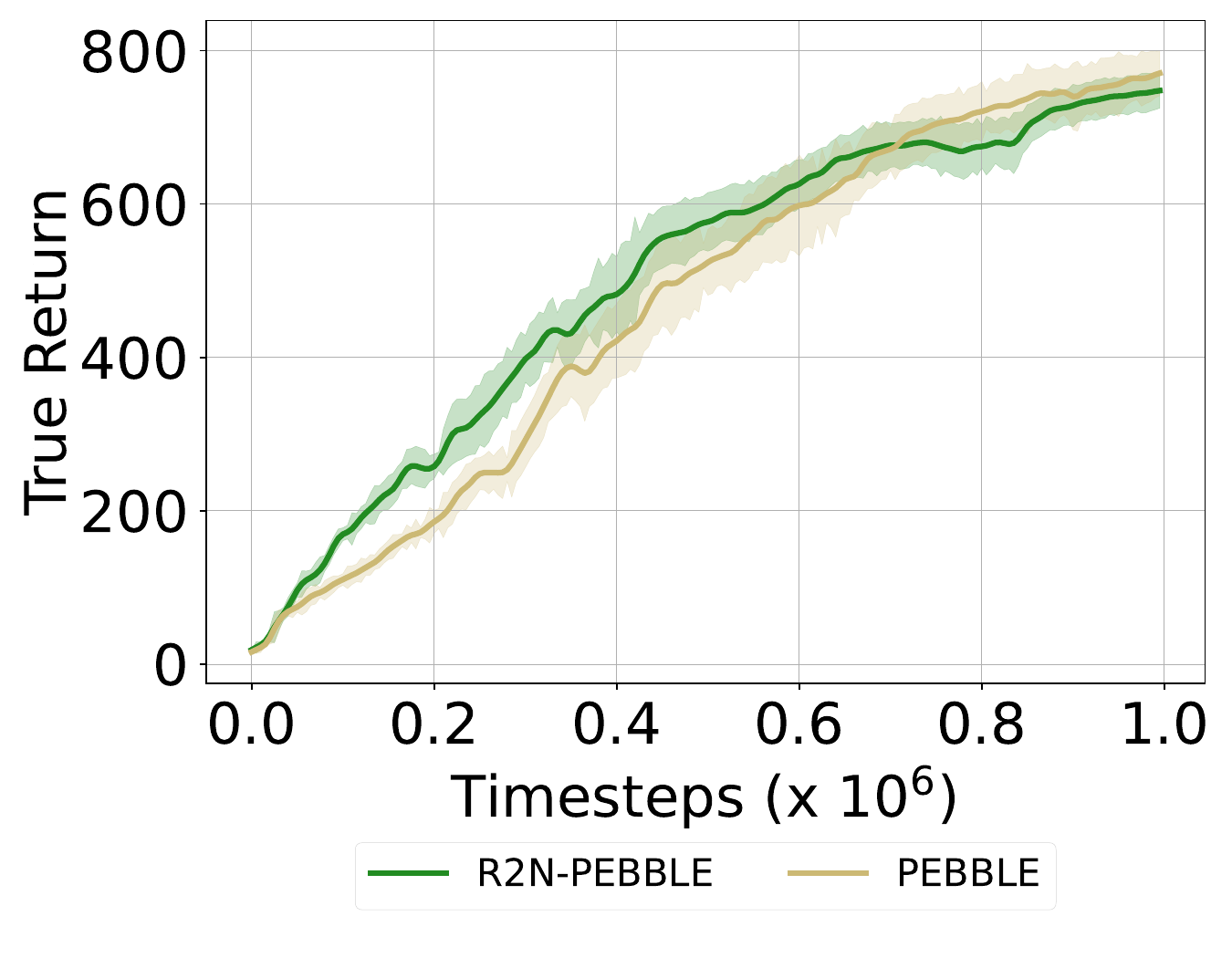}\label{fig:cheetah_run_50_4k}}
    \hfill
  \subfloat[Feedback = 10000]{\includegraphics[width=0.33\textwidth]{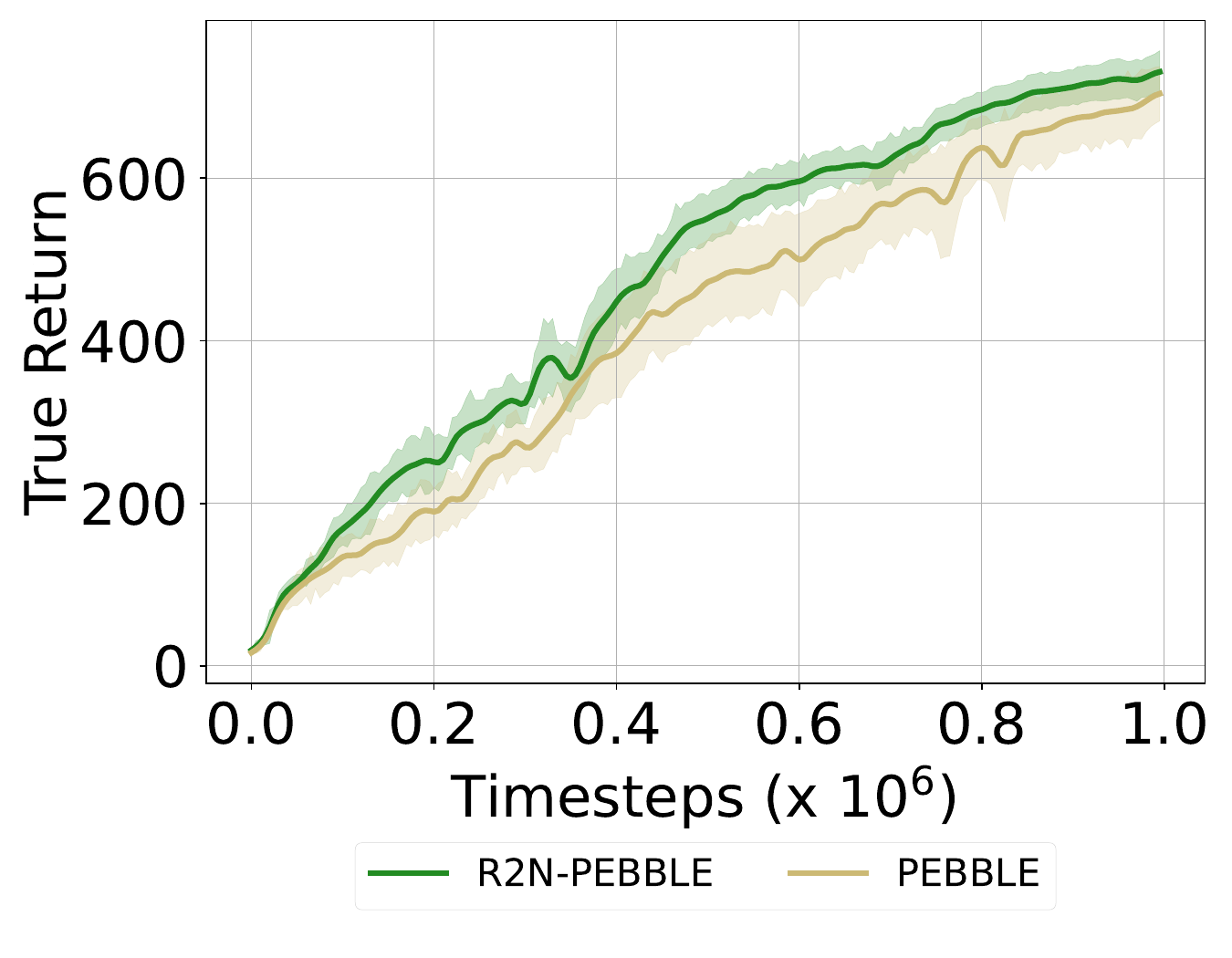}\label{fig:cheetah_run_50_10k}}
   
  \caption{Cheetah-run, Feedback Ablation, Noise = 50\%}\label{fig:cheetah_run_fb_ablation_noise50}
\end{figure*}

\begin{figure*}[!htbp]
  \centering
 \subfloat[Feedback = 100]{\includegraphics[width=0.33\textwidth]{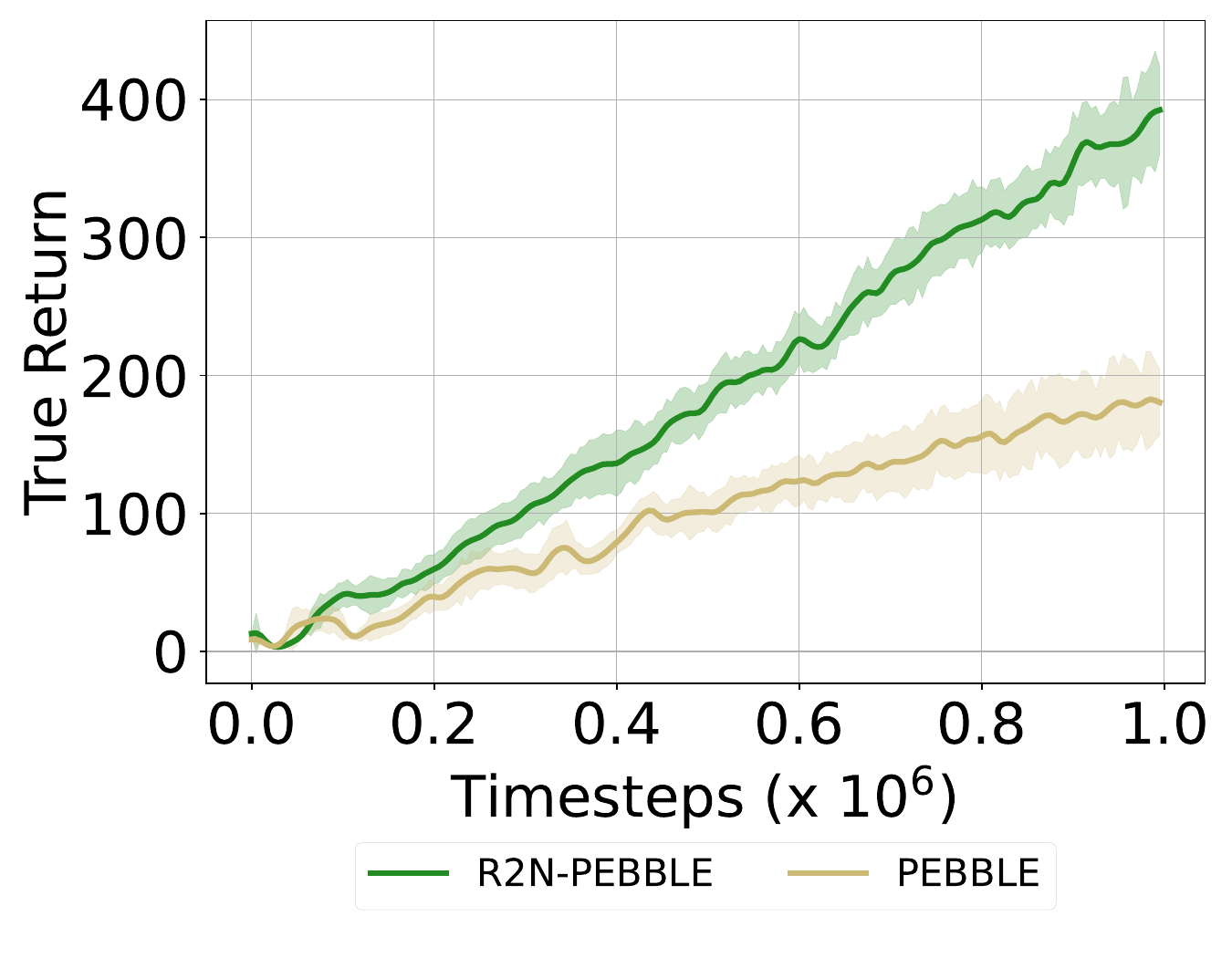}\label{fig:cheetah_run_90_100}}
  \hfill
\subfloat[Feedback = 200]{\includegraphics[width=0.33\textwidth]{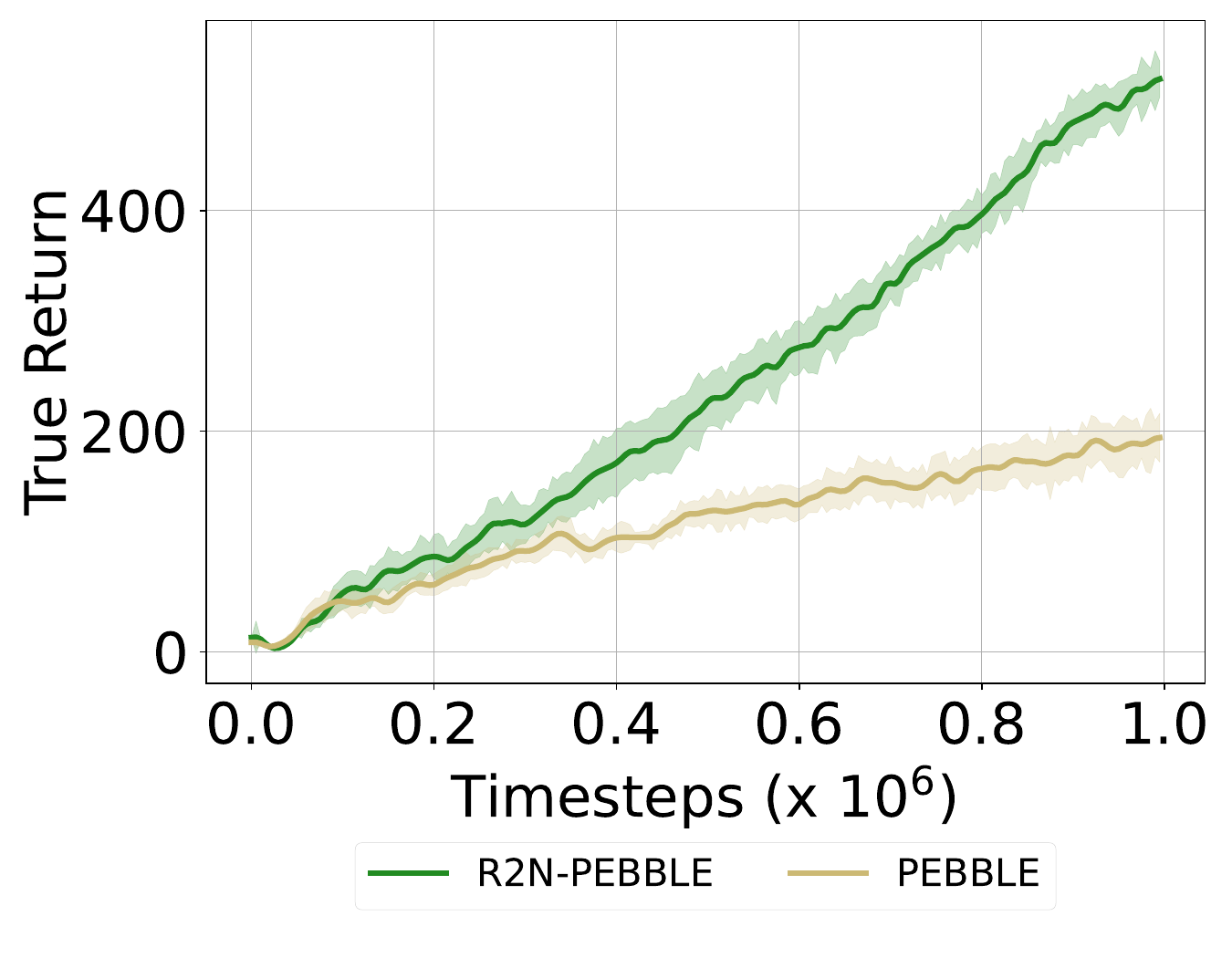}\label{fig:cheetah_run_90_200}}
  \hfill
\subfloat[Feedback = 400]
{\includegraphics[width=0.33\textwidth]{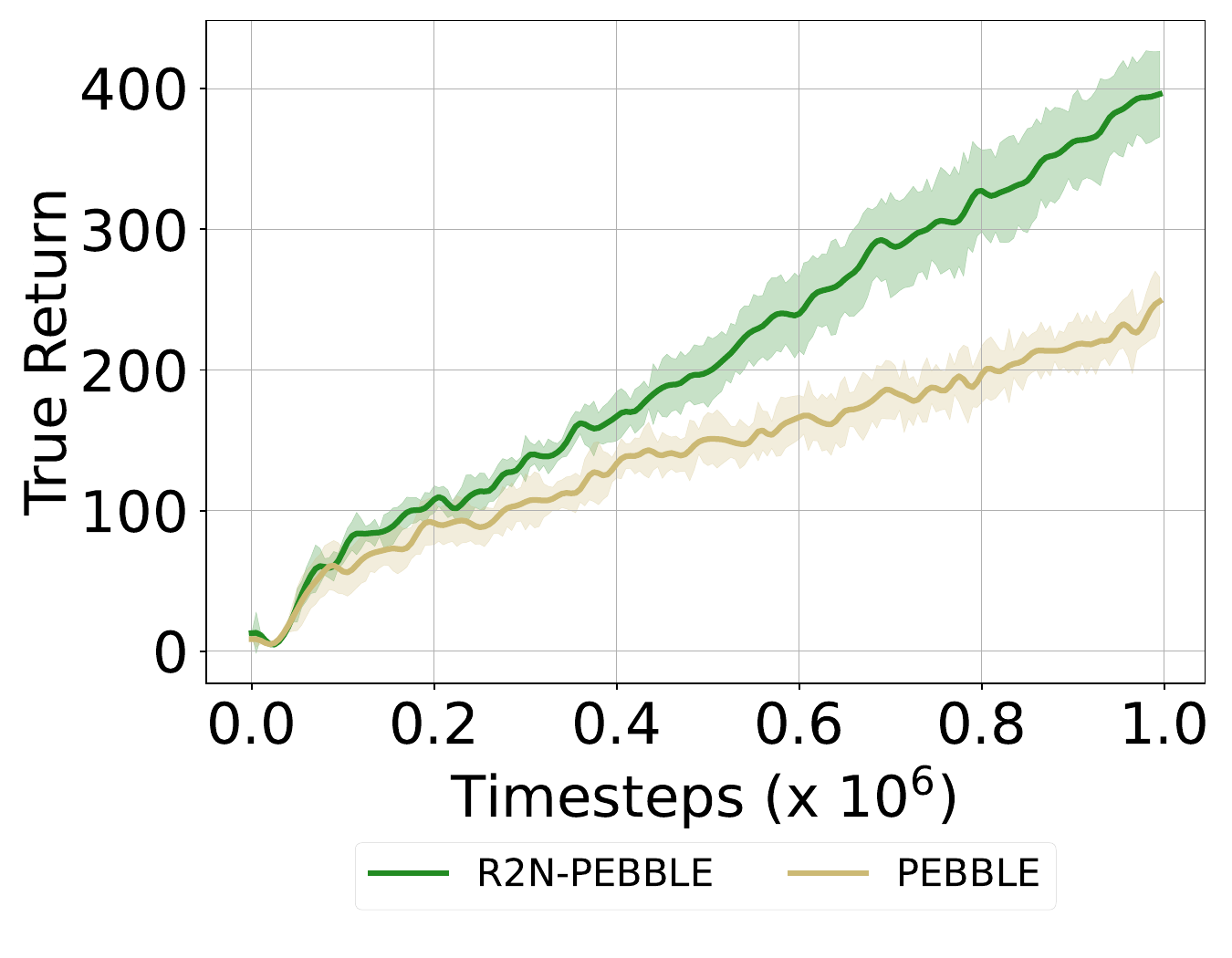}\label{fig:cheetah_run_90_400}}
  \hfill
  \subfloat[Feedback = 1000]{\includegraphics[width=0.33\textwidth]{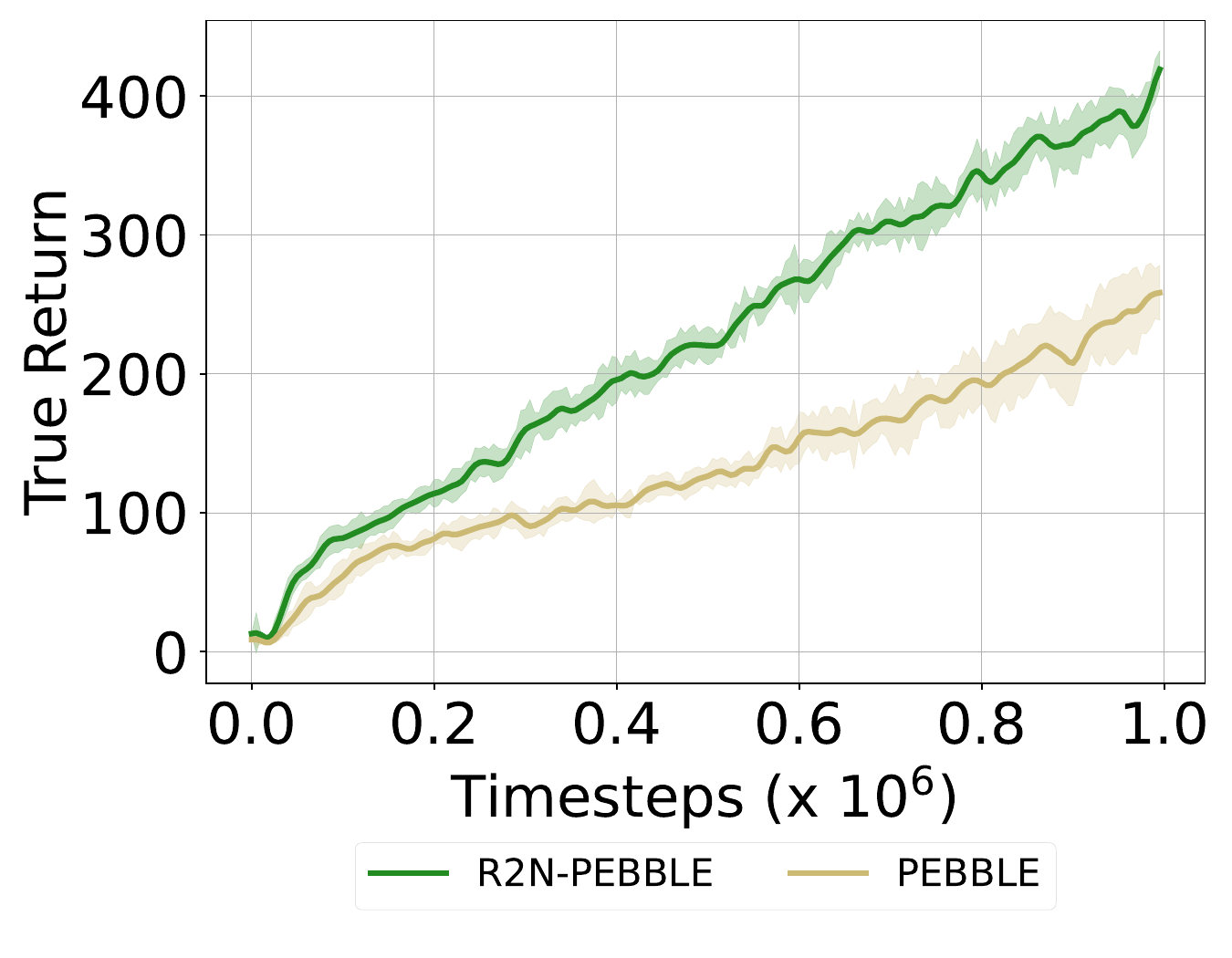}\label{fig:cheetah_run_90_1k}}
    \hfill
  \subfloat[Feedback = 2000]{\includegraphics[width=0.33\textwidth]{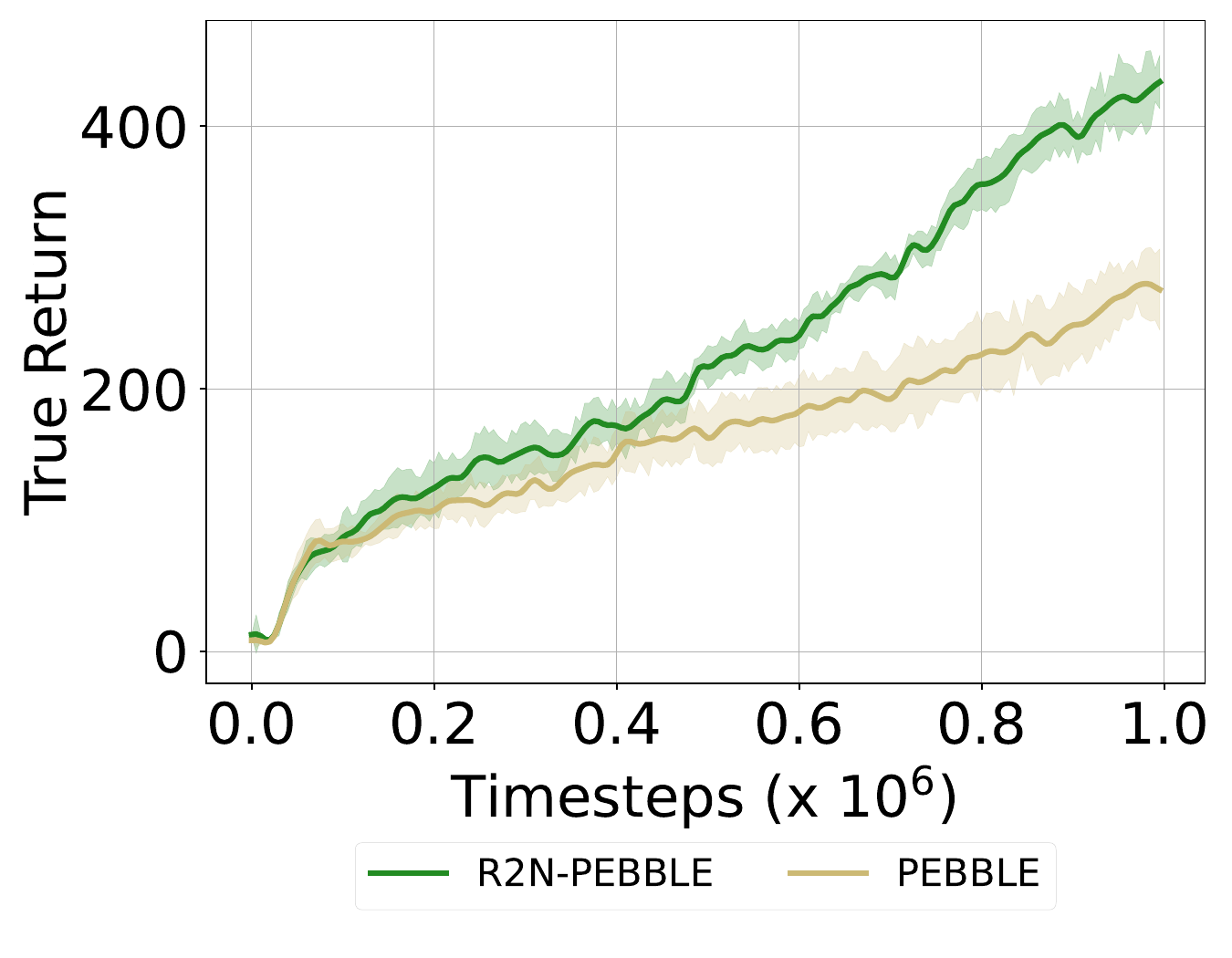}\label{fig:cheetah_run_90_2k}}
    \hfill
  \subfloat[Feedback = 4000]{\includegraphics[width=0.33\textwidth]{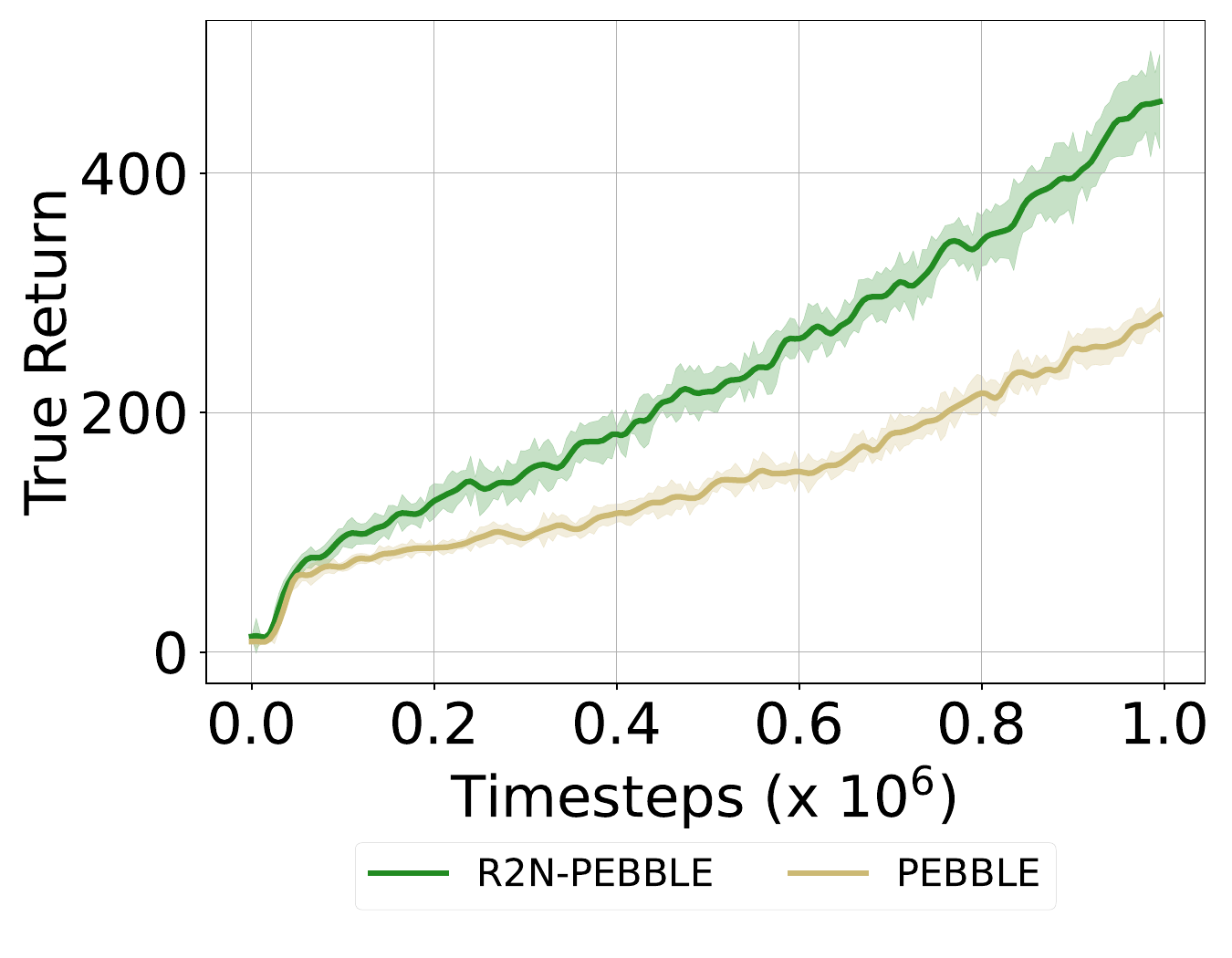}\label{fig:cheetah_run_90_4k}}
    \hfill
  \subfloat[Feedback = 10000]{\includegraphics[width=0.33\textwidth]{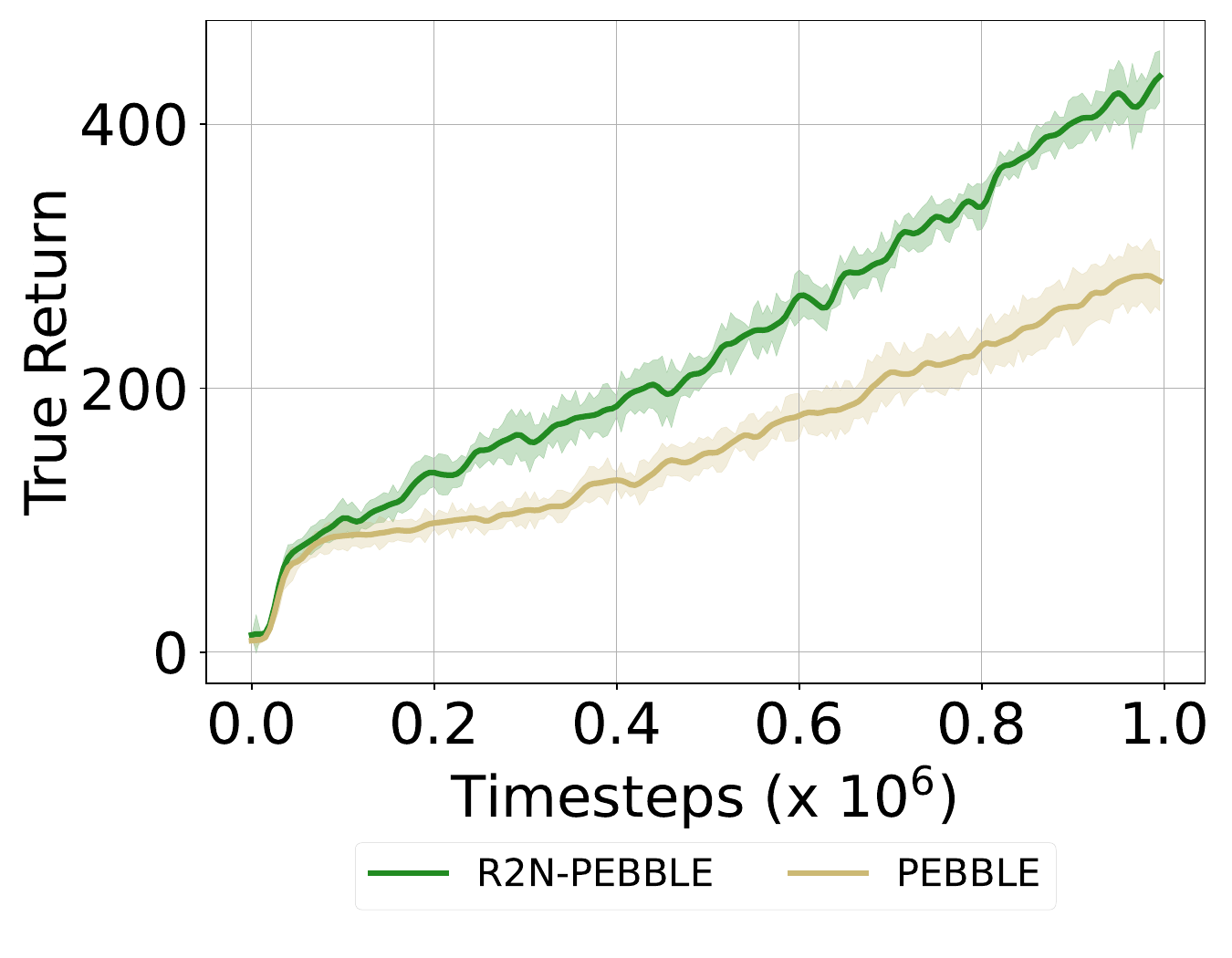}\label{fig:cheetah_run_90_10k}}
   
  \caption{Cheetah-run, Feedback Ablation, Noise = 90\%}\label{fig:cheetah_run_fb_ablation_noise90}
\end{figure*}

\begin{figure*}[t]
  \centering
 \subfloat[Feedback = 100]{\includegraphics[width=0.33\textwidth]{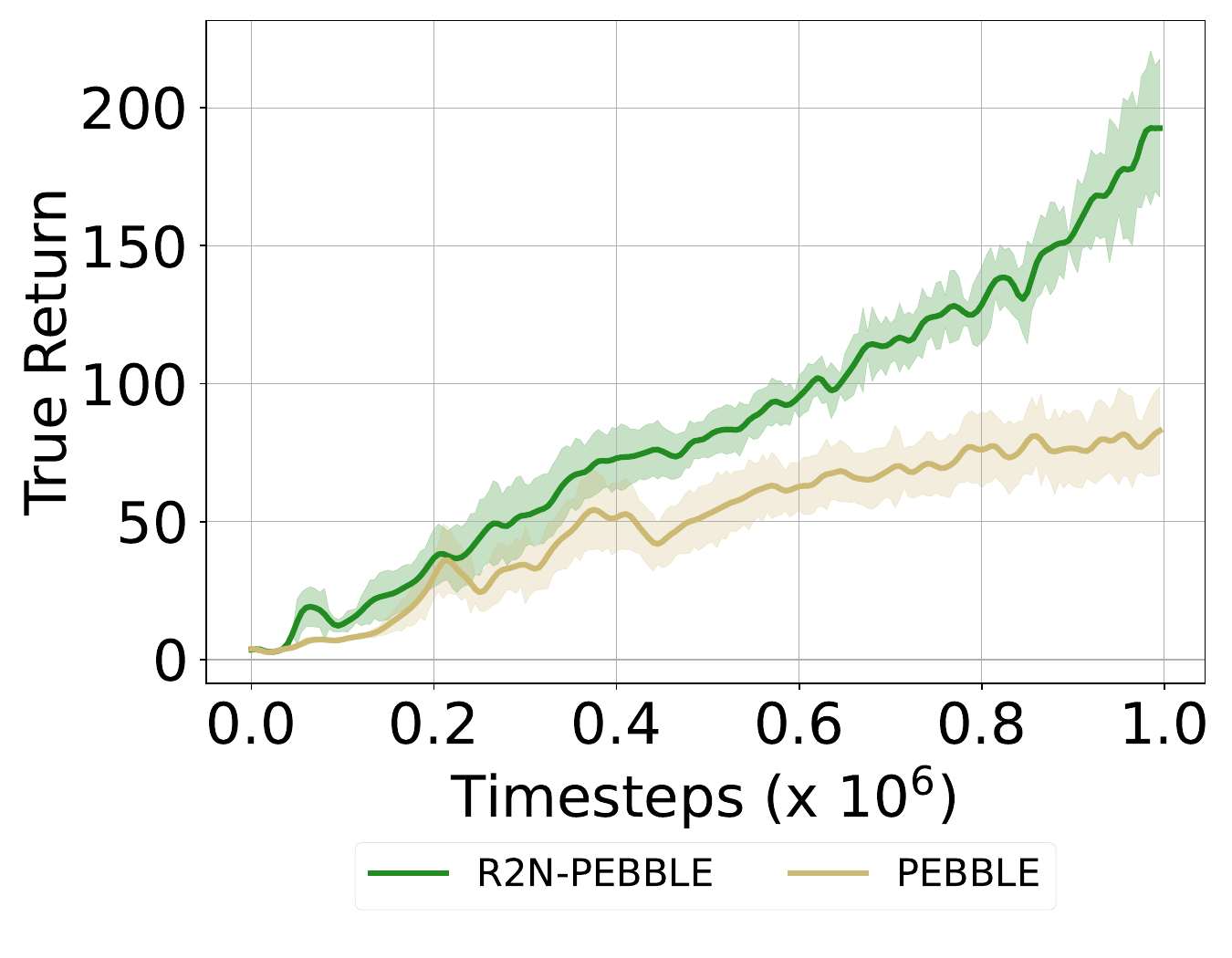}\label{fig:cheetah_run_95_100}}
  \hfill
\subfloat[Feedback = 200]{\includegraphics[width=0.33\textwidth]{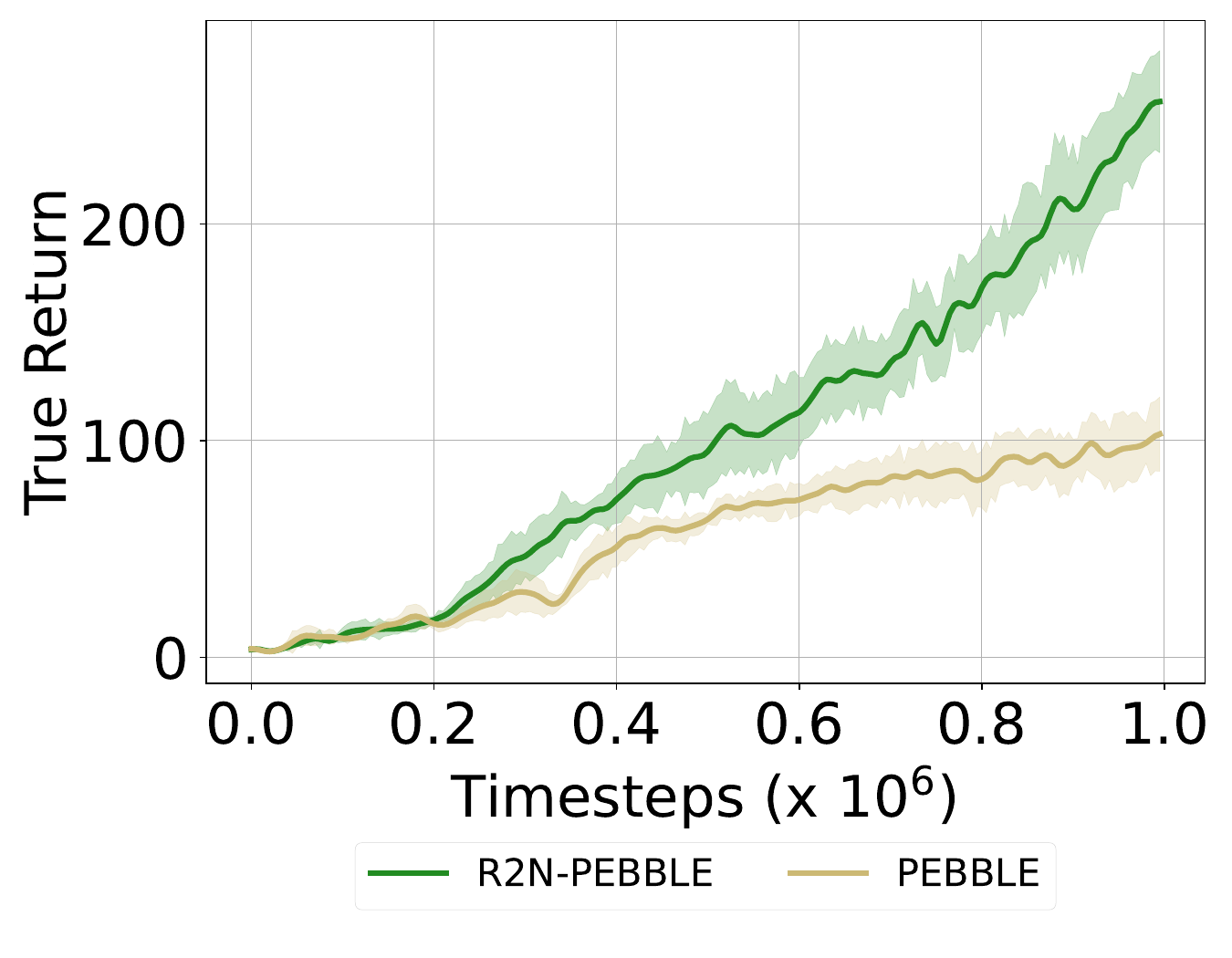}\label{fig:cheetah_run_95_200}}
  \hfill
\subfloat[Feedback = 400]
{\includegraphics[width=0.33\textwidth]{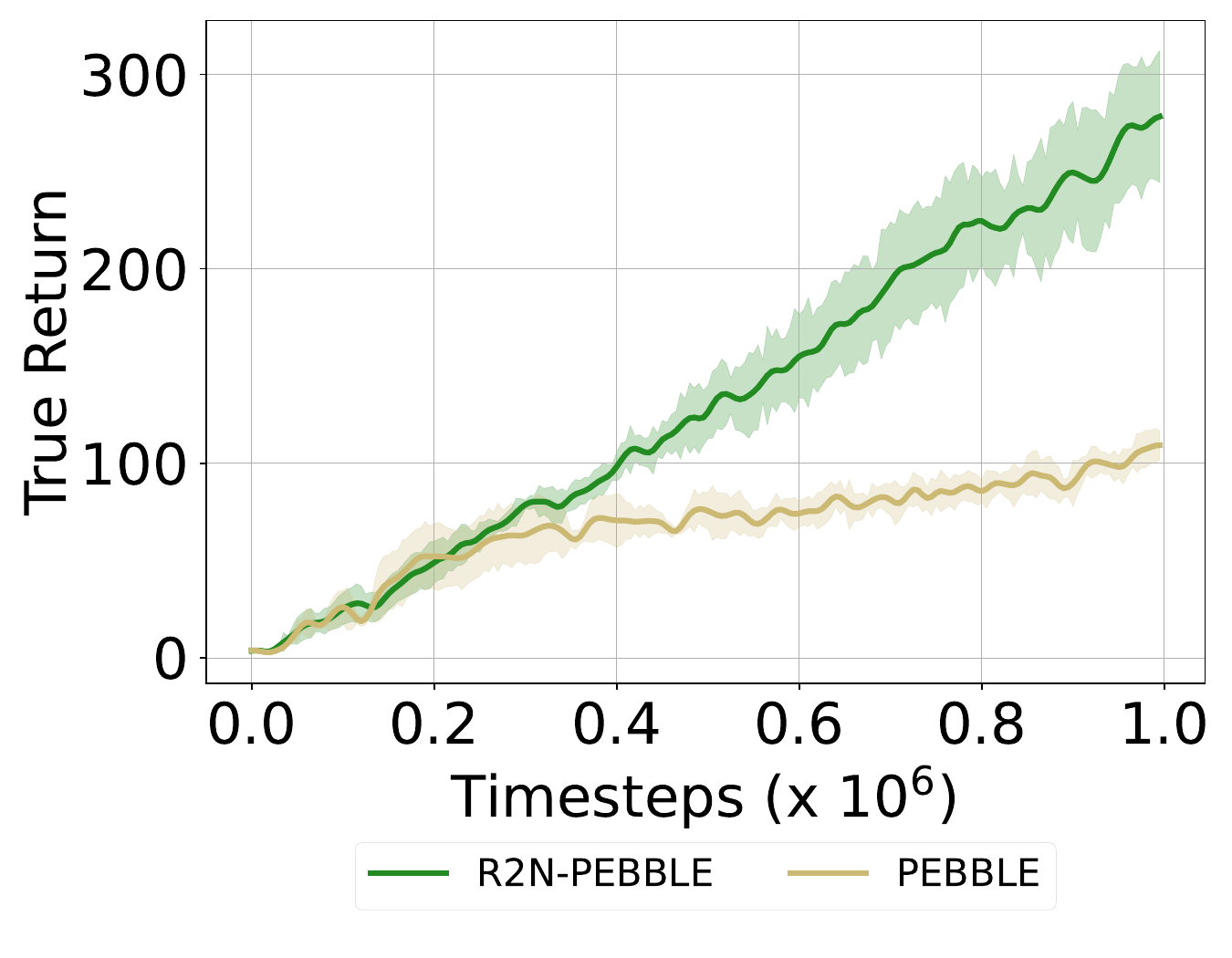}\label{fig:cheetah_run_95_400}}
  \hfill
  \subfloat[Feedback = 1000]{\includegraphics[width=0.33\textwidth]{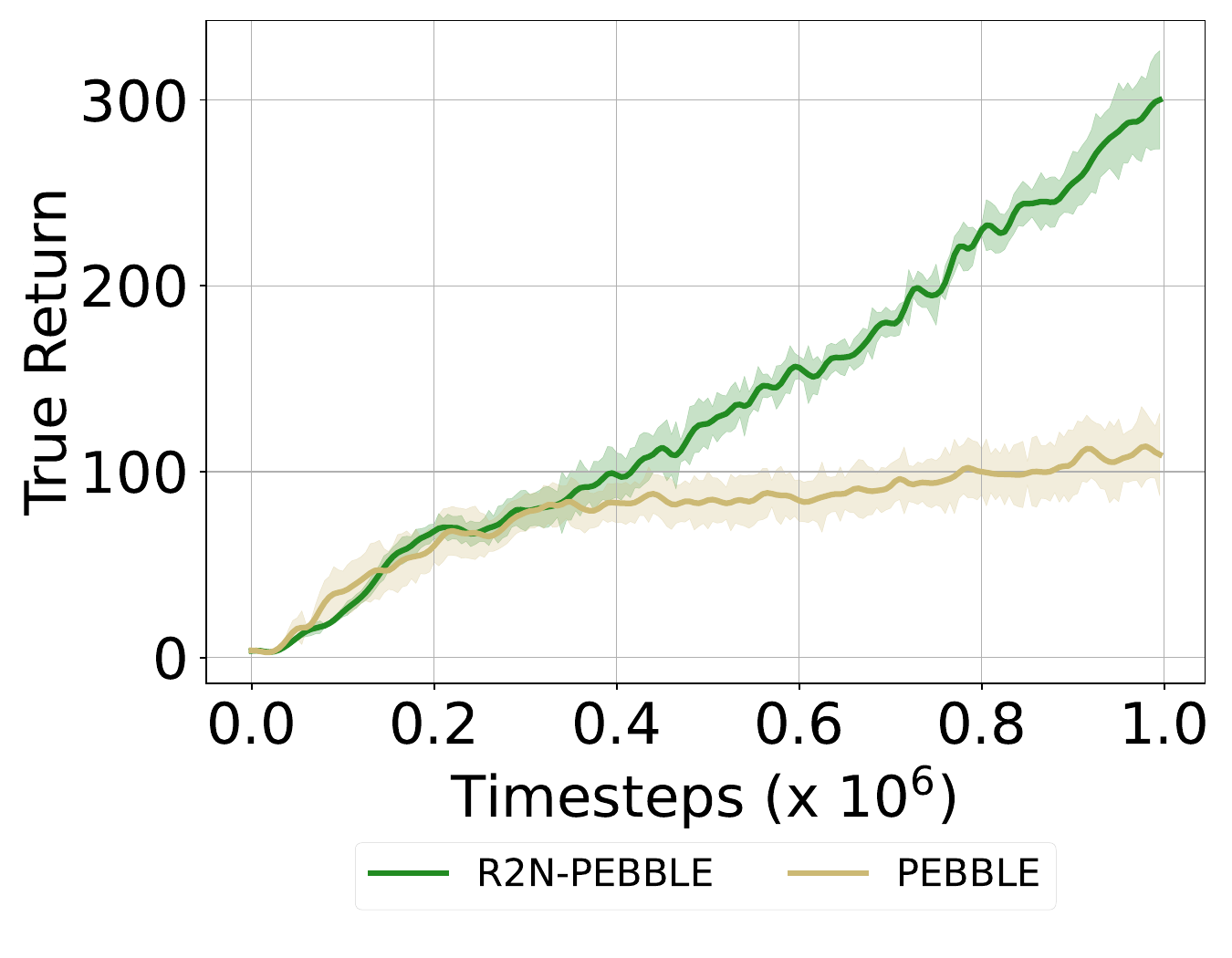}\label{fig:cheetah_run_95_1k}}
    \hfill
  \subfloat[Feedback = 2000]{\includegraphics[width=0.33\textwidth]{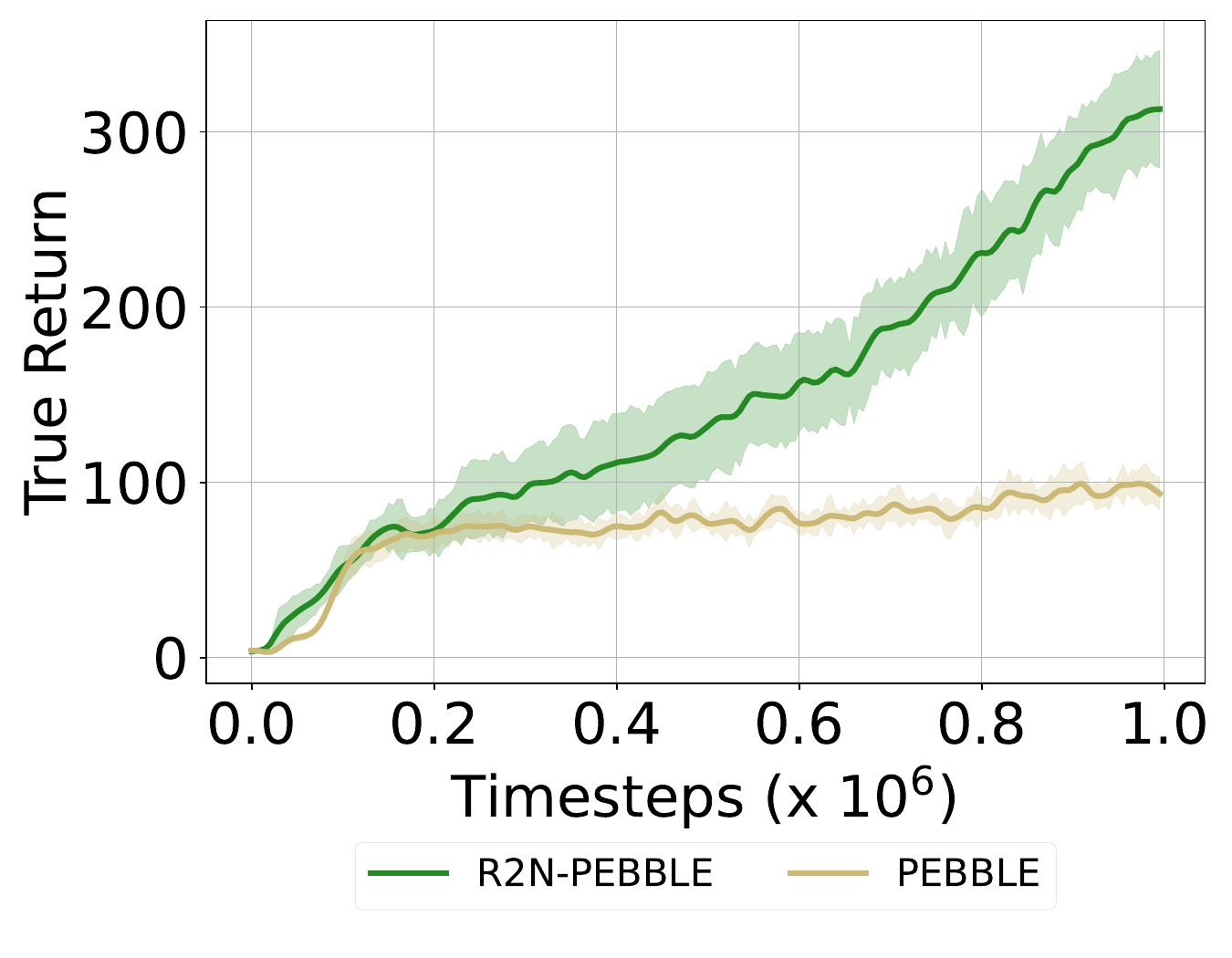}\label{fig:cheetah_run_95_2k}}
    \hfill
  \subfloat[Feedback = 4000]{\includegraphics[width=0.33\textwidth]{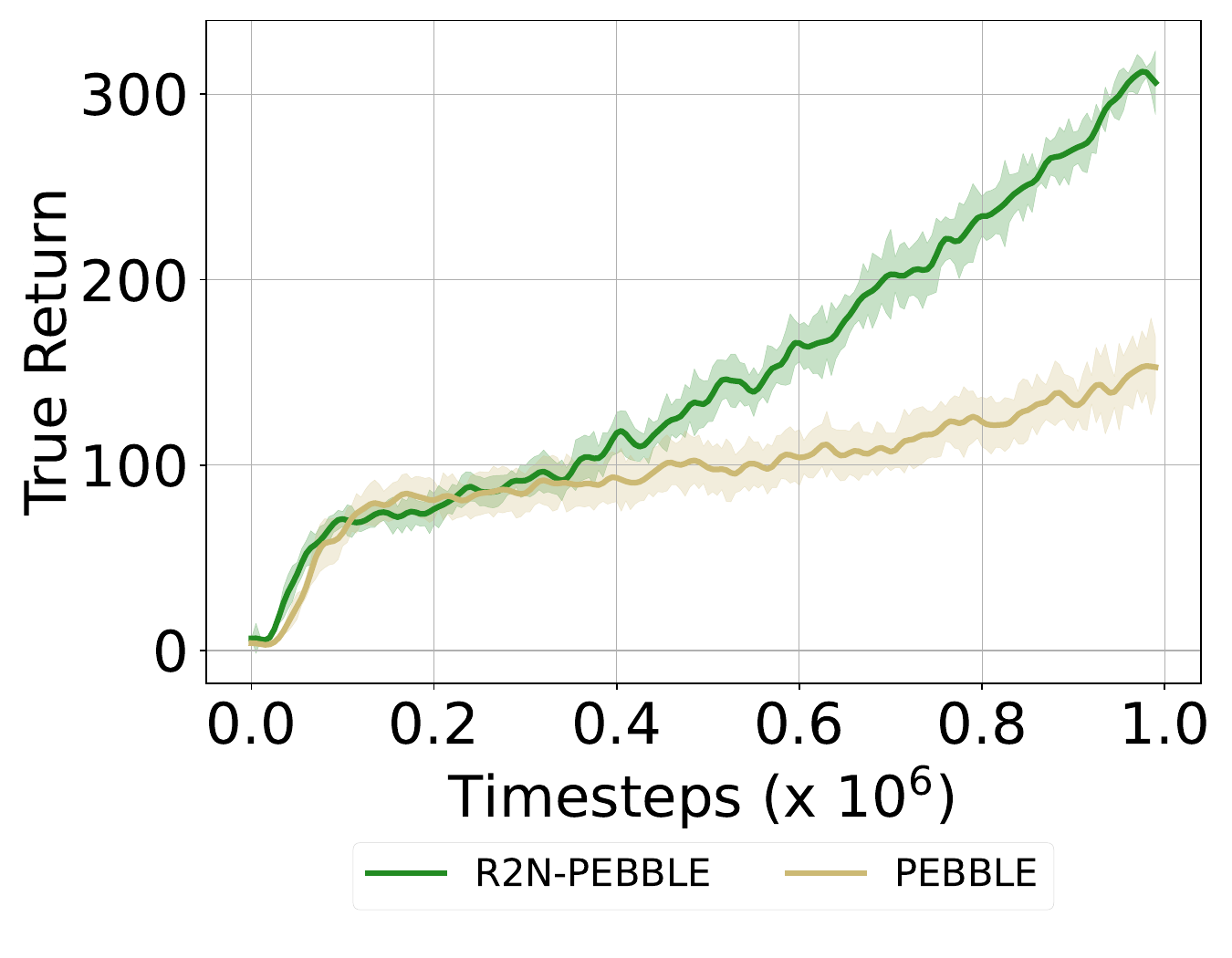}\label{fig:cheetah_run_95_4k}}
    \hfill
  \subfloat[Feedback = 10000]{\includegraphics[width=0.33\textwidth]{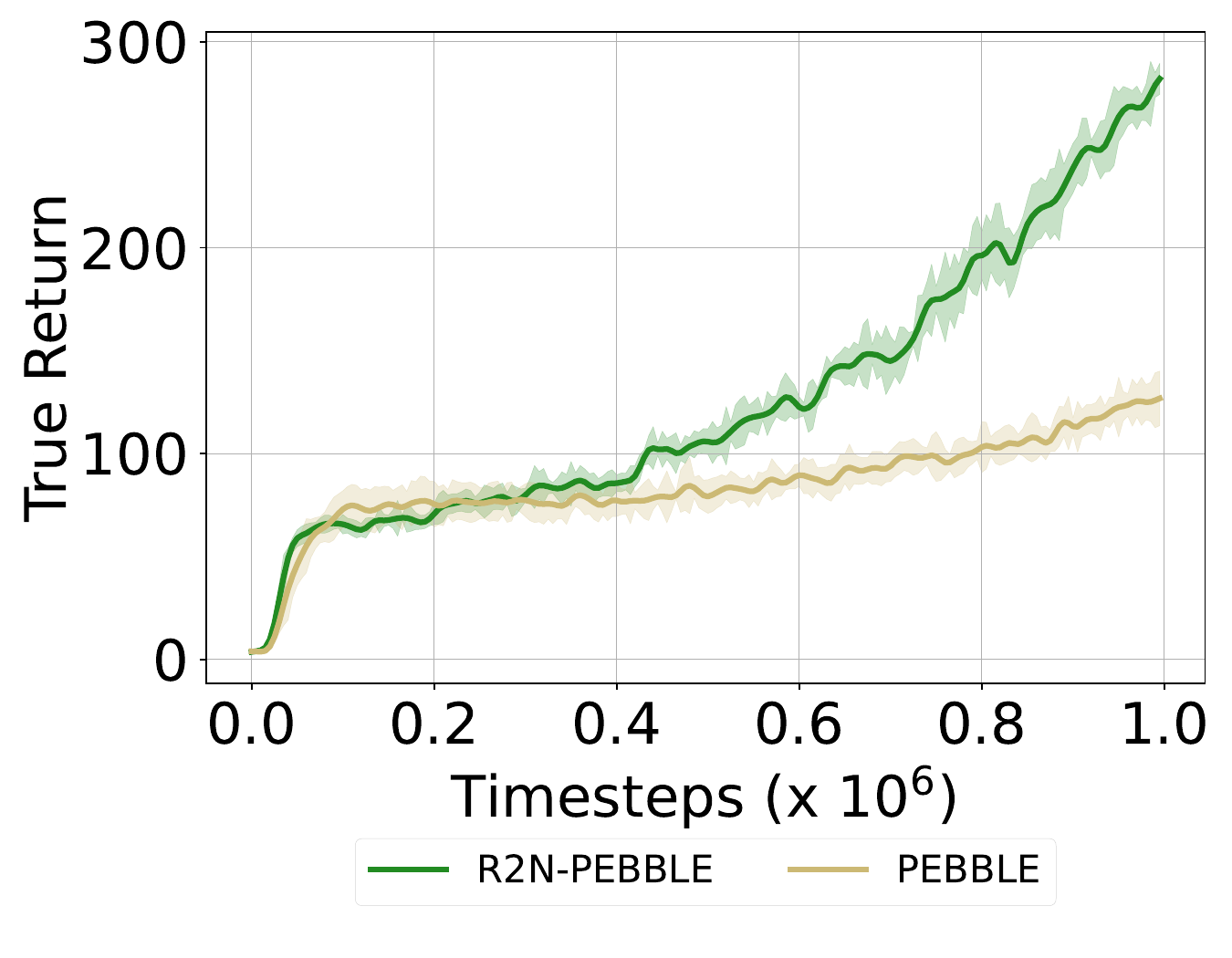}\label{fig:cheetah_run_95_10k}}
   
  \caption{Cheetah-run, Feedback Ablation, Noise = 95\%}\label{fig:cheetah_run_fb_ablation_noise95}
\end{figure*}

In this section, we present additional experimental results ablating over the noise fractions and preference feedback budgets for the Cheetah-run environment.  Overall, in Figures \ref{fig:cheetah_run_fb_ablation_noise20} through \ref{fig:cheetah_run_fb_ablation_noise95}, we find that R2N-PEBBLE (green curves) is generally more robust than PEBBLE (yellow curves) in varying noise levels and feedback amounts. Especially in environments with high noise levels, such as 90-95\%, shown in Figures~\ref{fig:cheetah_run_fb_ablation_noise90} and \ref{fig:cheetah_run_fb_ablation_noise95}, R2N significantly improves the performance of PEBBLE.

\clearpage

\subsection{Feedback and Noise Ablations in Walker-walk}
\label{sec:fb_noise_ablations_walker_walk}

\begin{figure*}[h!]
  \centering
 \subfloat[Feedback = 100]{\includegraphics[width=0.33\textwidth]{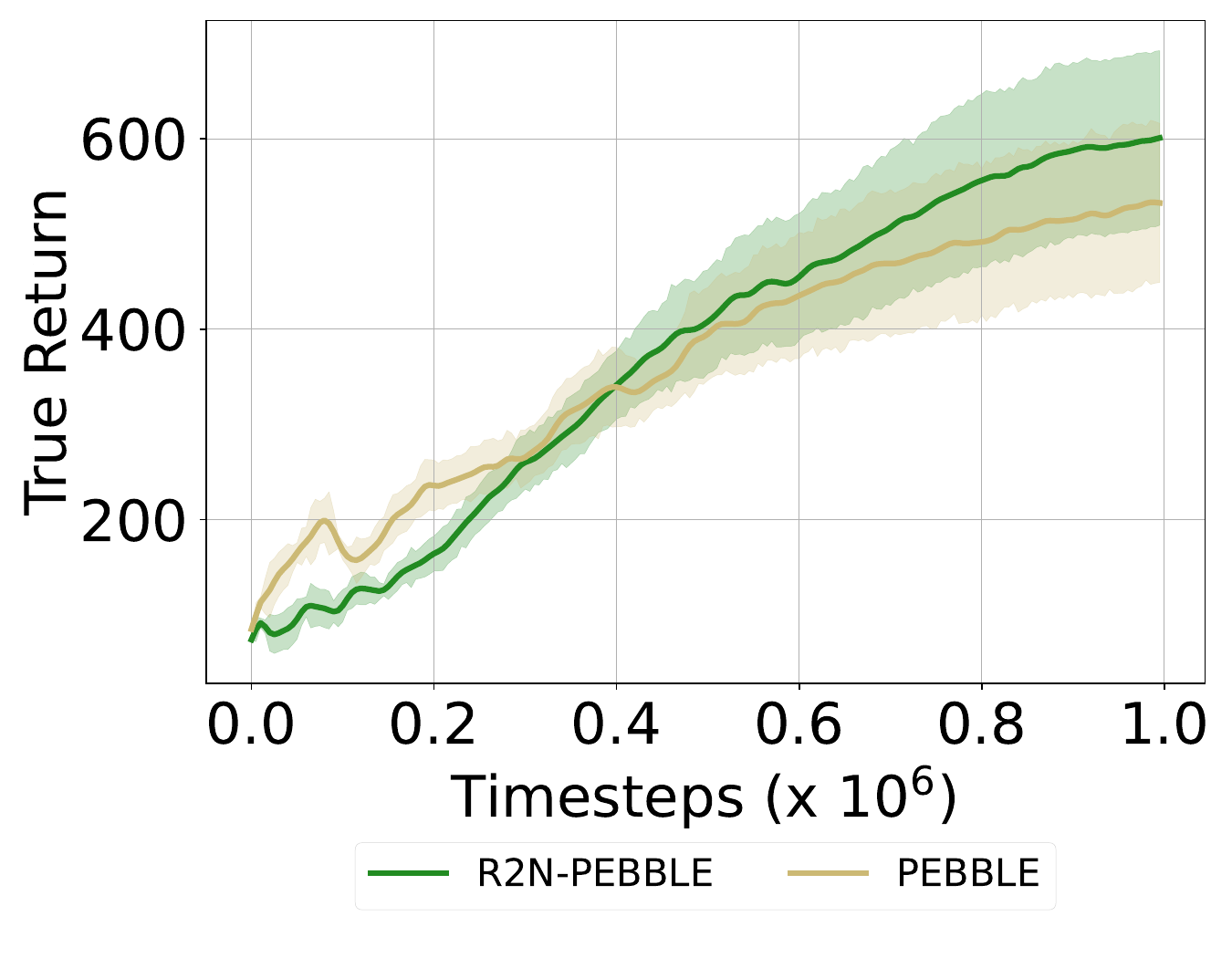}\label{fig:walker_walk_20_100}}
  \hfill
\subfloat[Feedback = 200]{\includegraphics[width=0.33\textwidth]{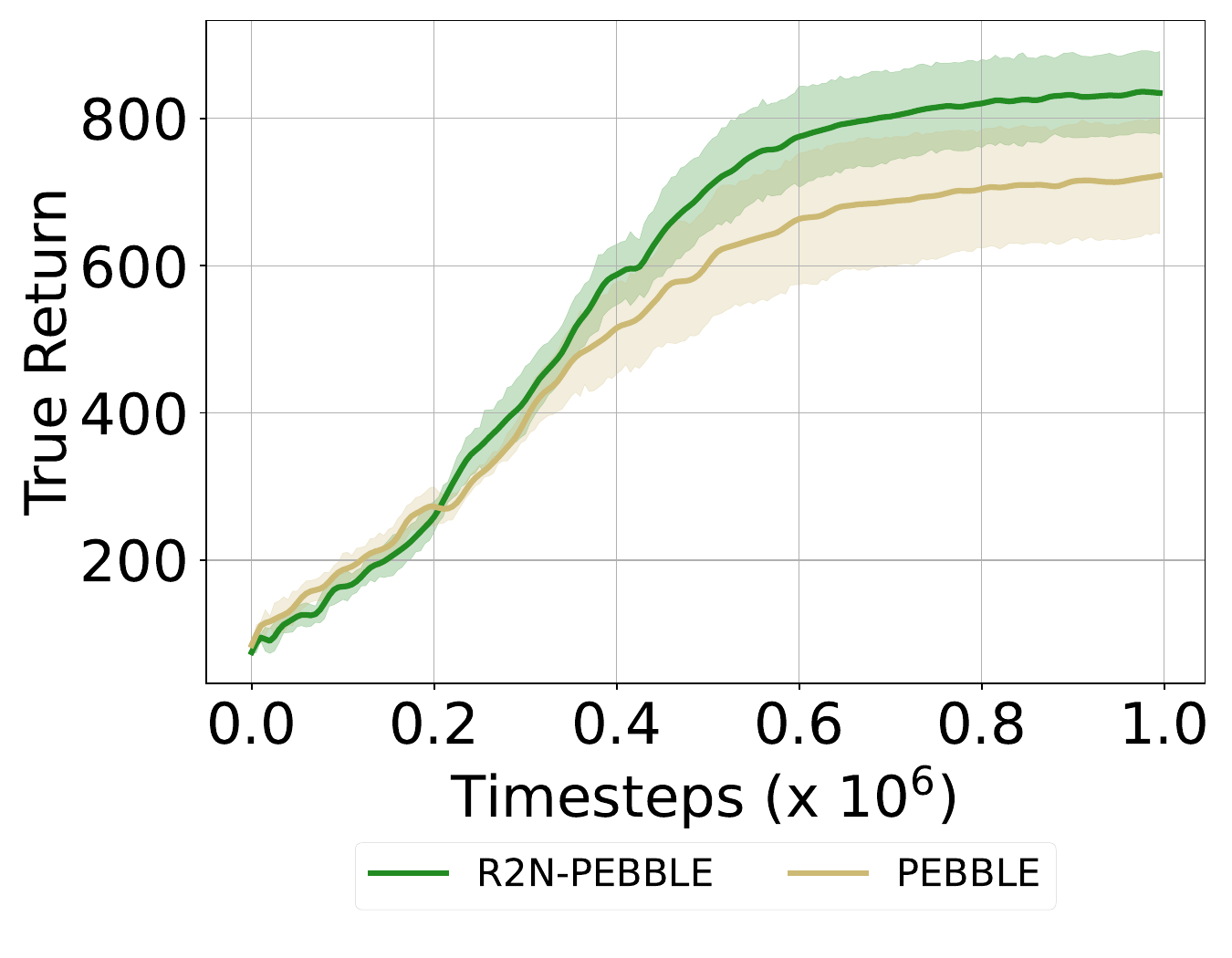}\label{fig:walker_walk_20_200}}
  \hfill
\subfloat[Feedback = 400]
{\includegraphics[width=0.33\textwidth]{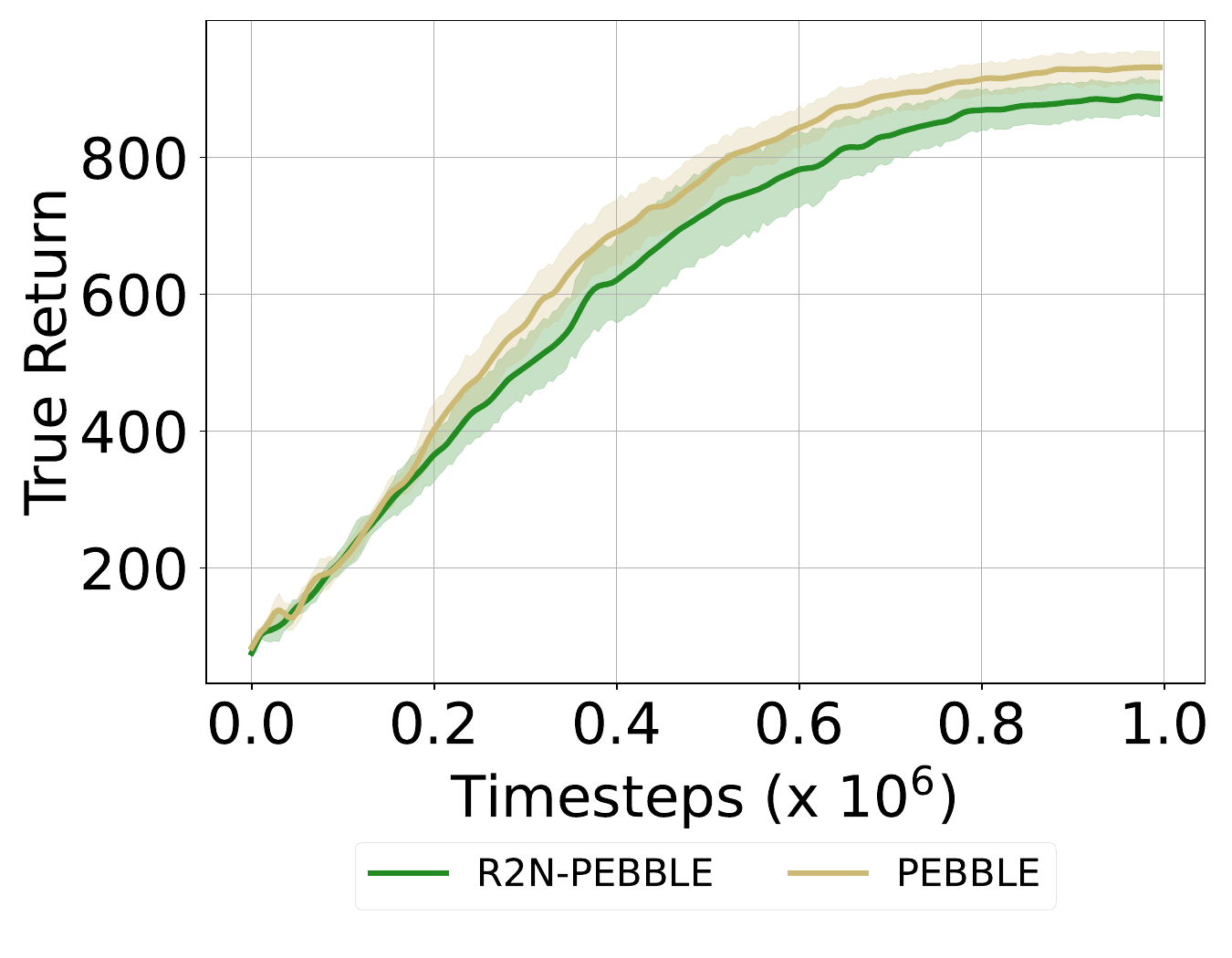}\label{fig:walker_walk_20_400}}
  \hfill
  \subfloat[Feedback = 1000]{\includegraphics[width=0.33\textwidth]{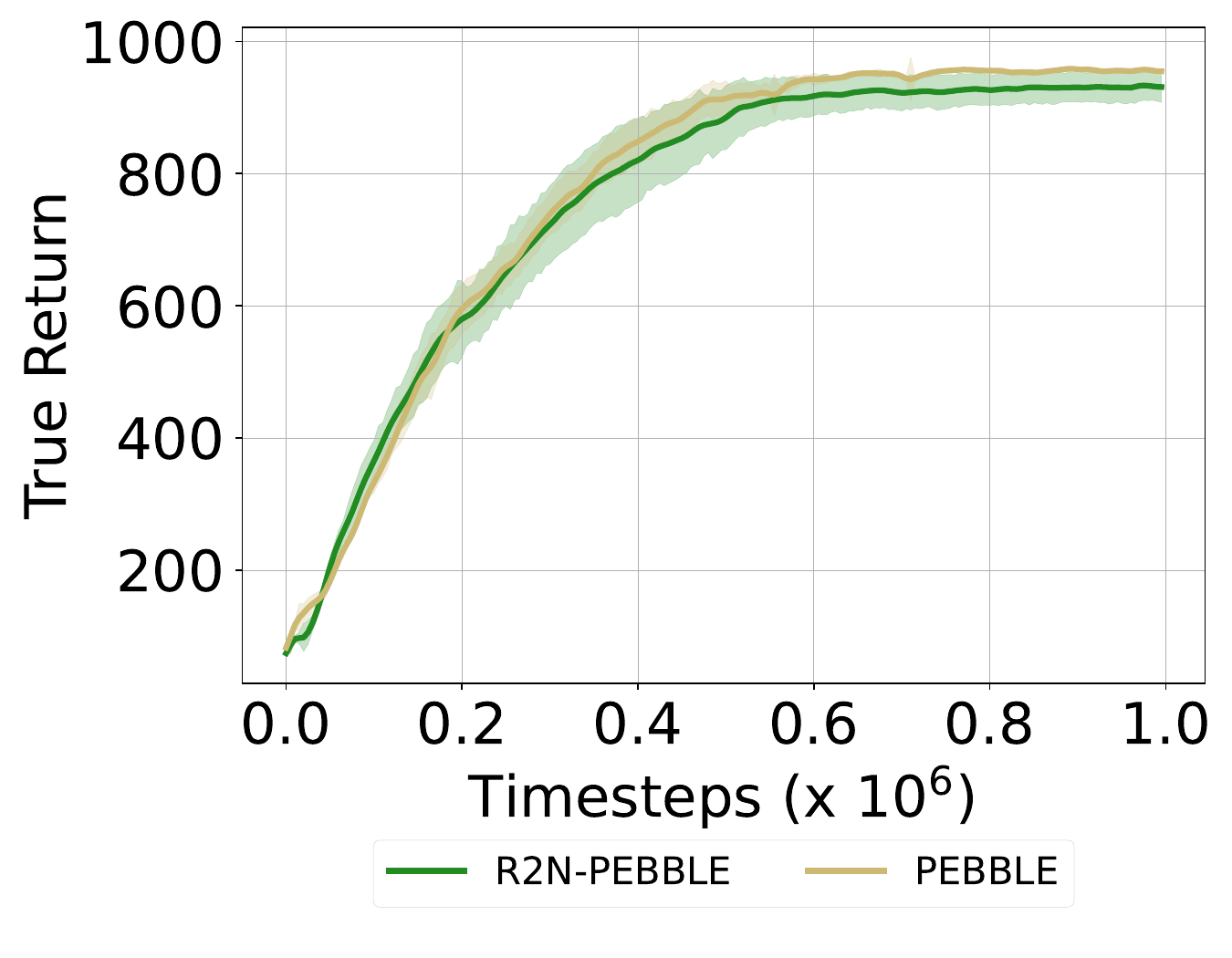}\label{fig:walker_walk_20_1000}}
    \hfill
  \subfloat[Feedback = 2000]{\includegraphics[width=0.33\textwidth]{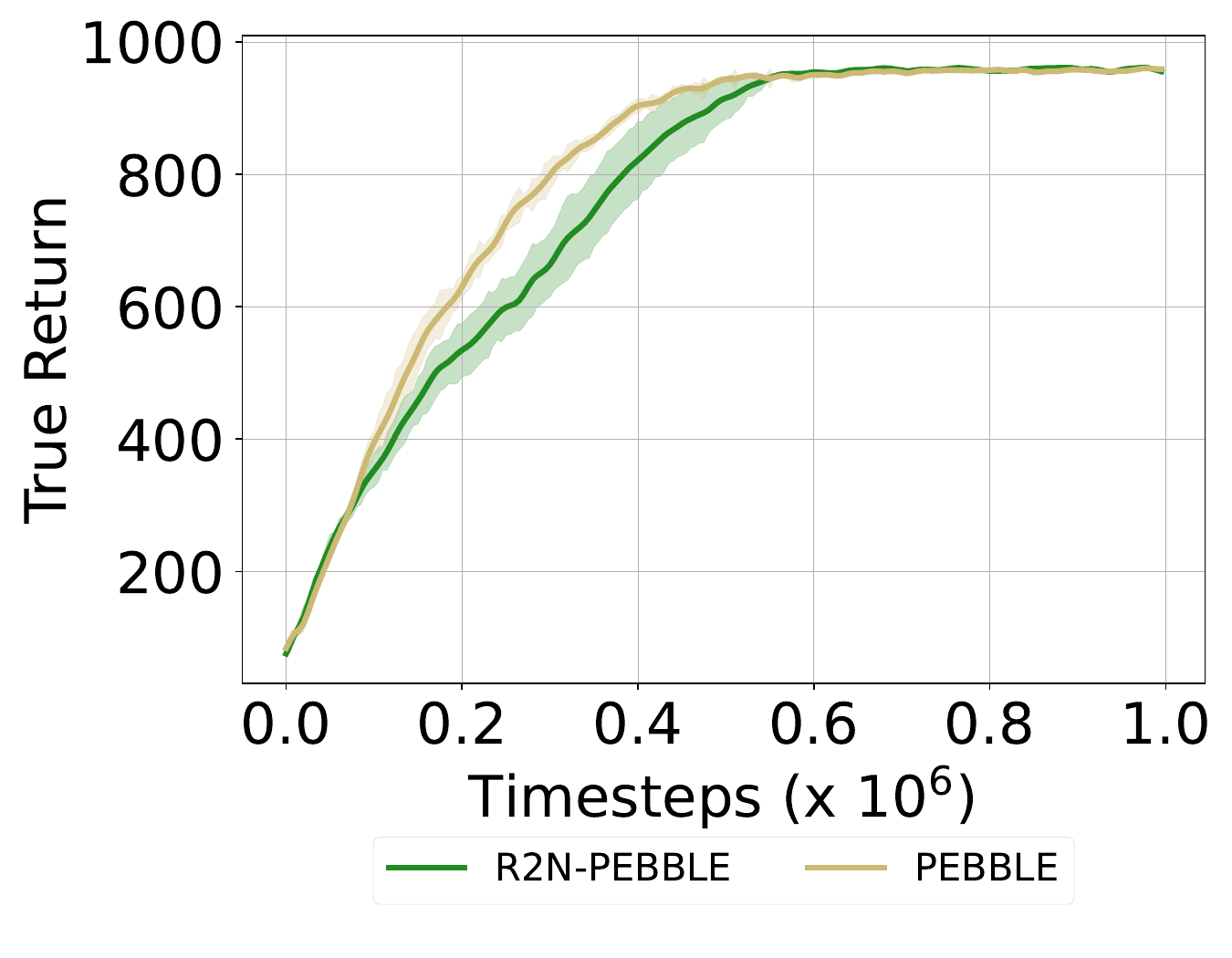}\label{fig:walker_walk_20_2000}}
    \hfill
  \subfloat[Feedback = 4000]{\includegraphics[width=0.33\textwidth]{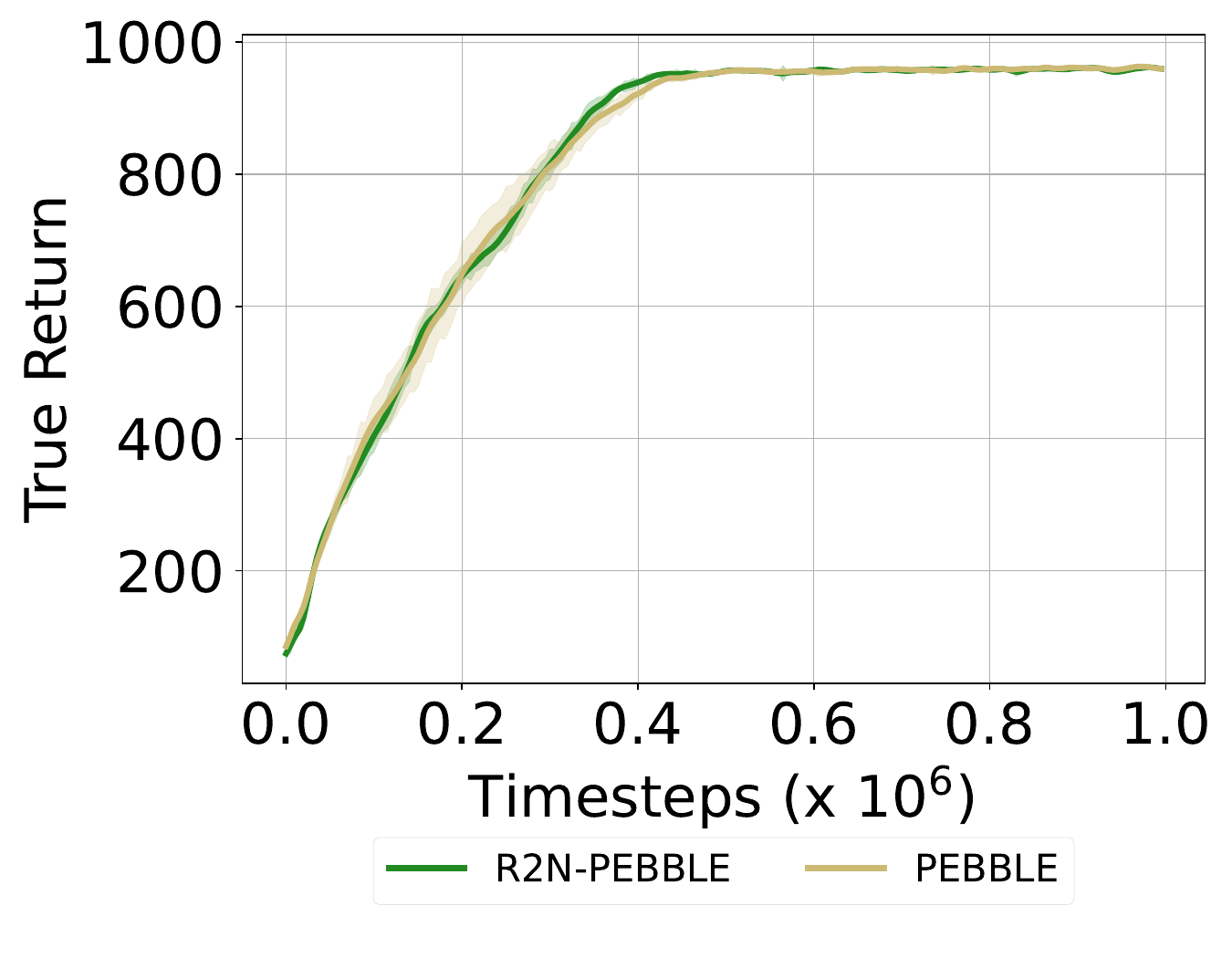}\label{fig:walker_walk_20_4000}}
    \hfill
  \subfloat[Feedback = 10000]{\includegraphics[width=0.33\textwidth]{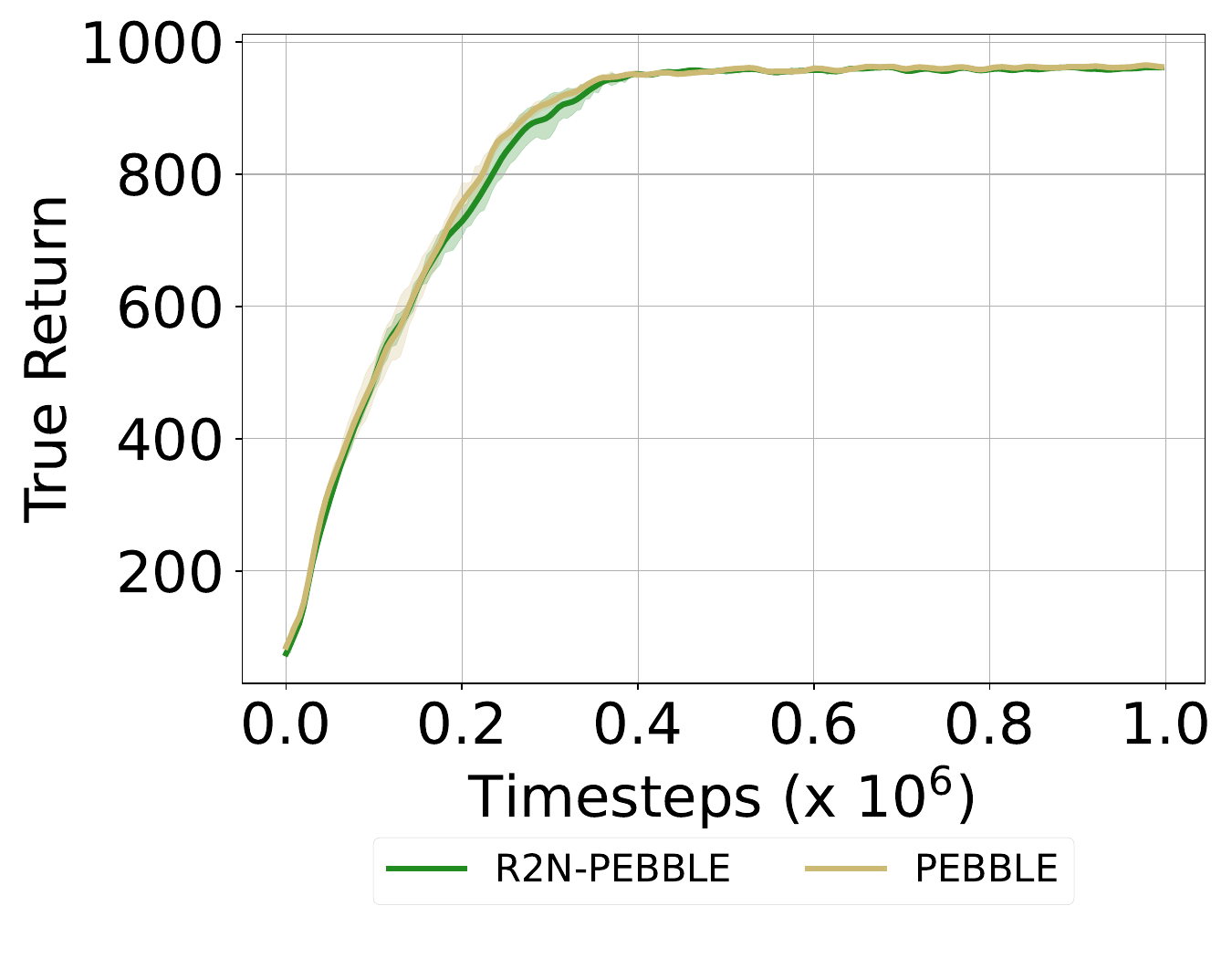}\label{fig:walker_walk_20_10000}}
   
  \caption{Walker-walk, Feedback Ablation, Noise = 20\%}\label{fig:walker_walk_fb_ablation_noise20}
\end{figure*}

\begin{figure*}[h!]
  \centering
 \subfloat[Feedback = 100]{\includegraphics[width=0.33\textwidth]{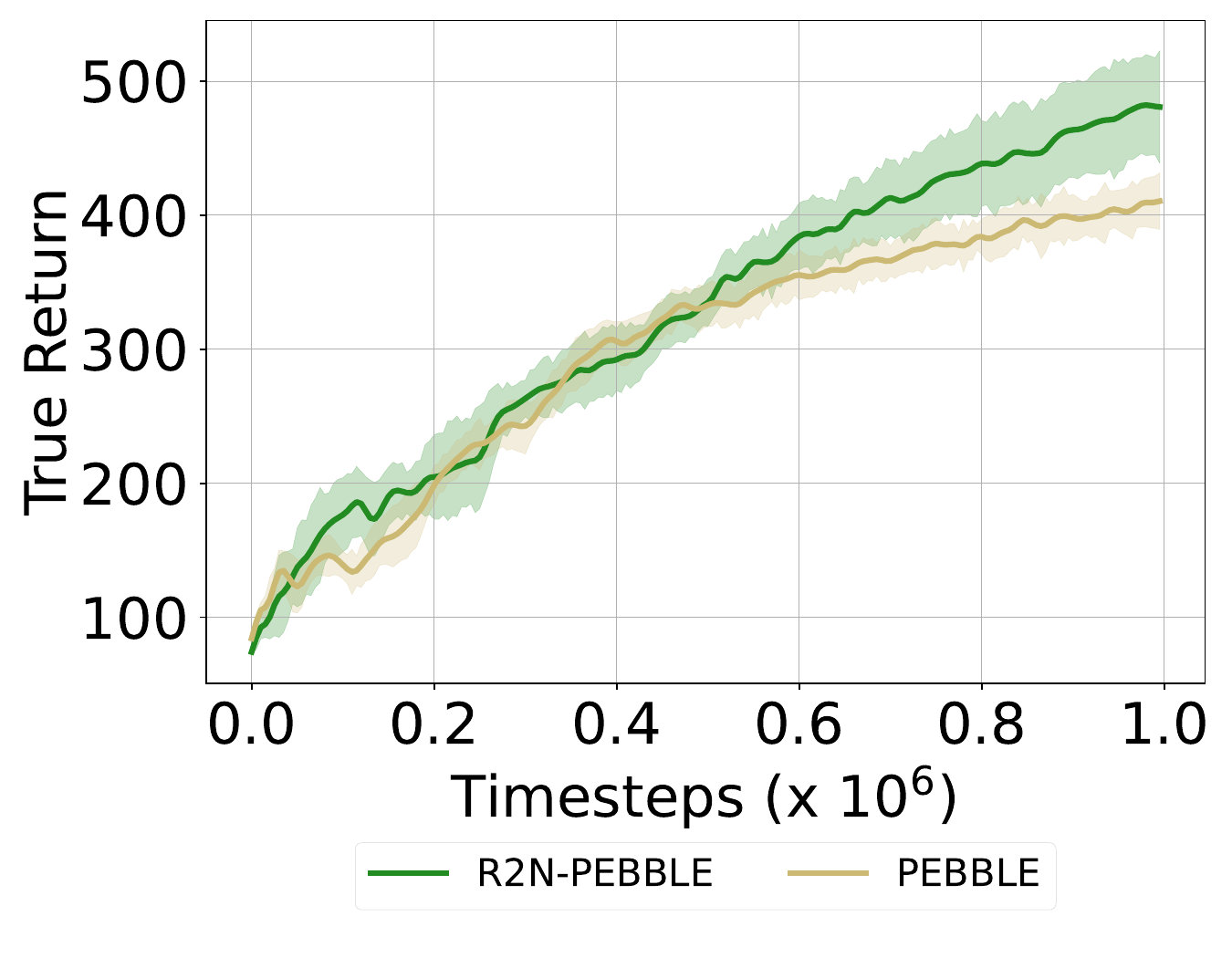}\label{fig:walker_walk_50_100}}
  \hfill
\subfloat[Feedback = 200]{\includegraphics[width=0.33\textwidth]{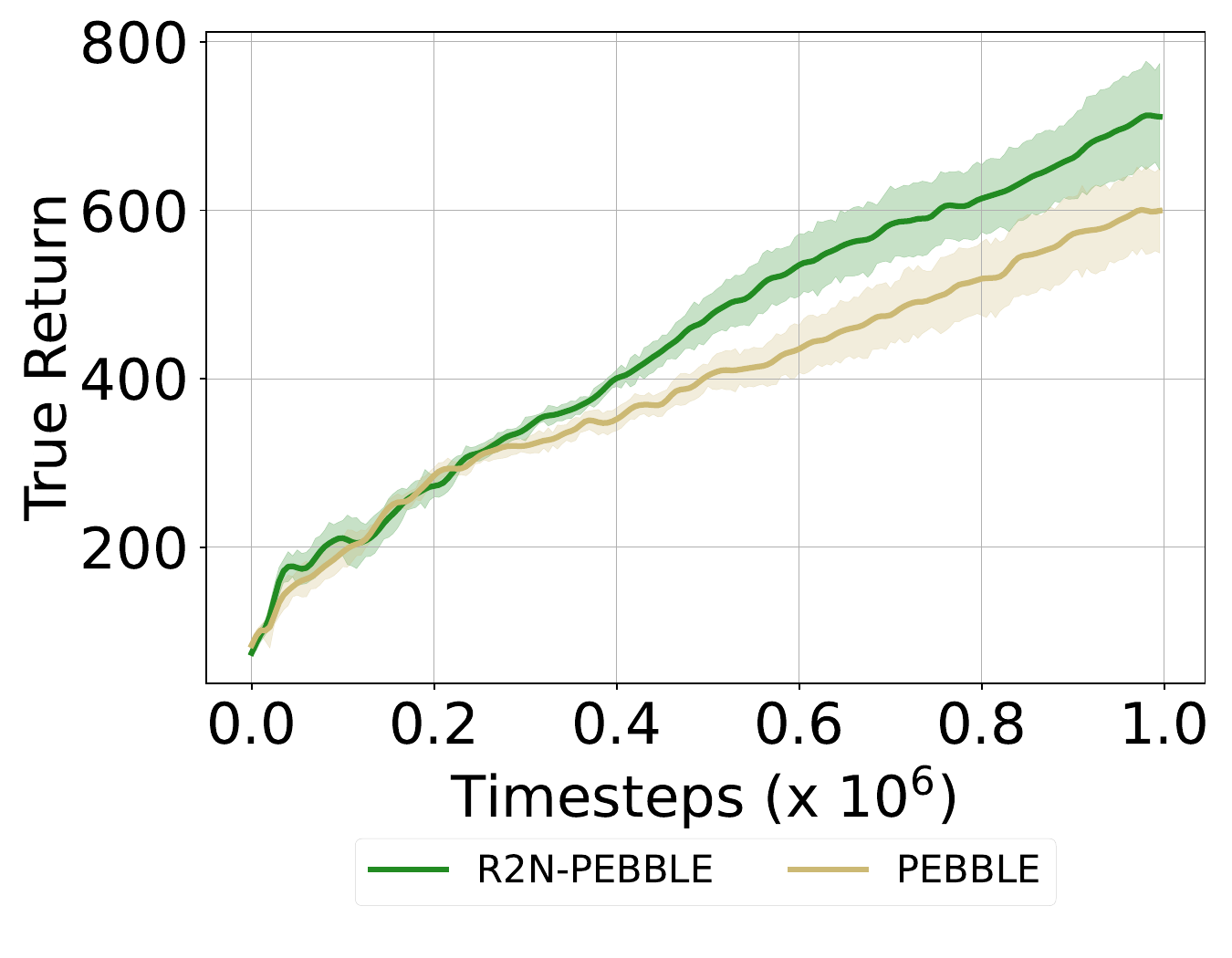}\label{fig:walker_walk_50_200}}
  \hfill
\subfloat[Feedback = 400]
{\includegraphics[width=0.33\textwidth]{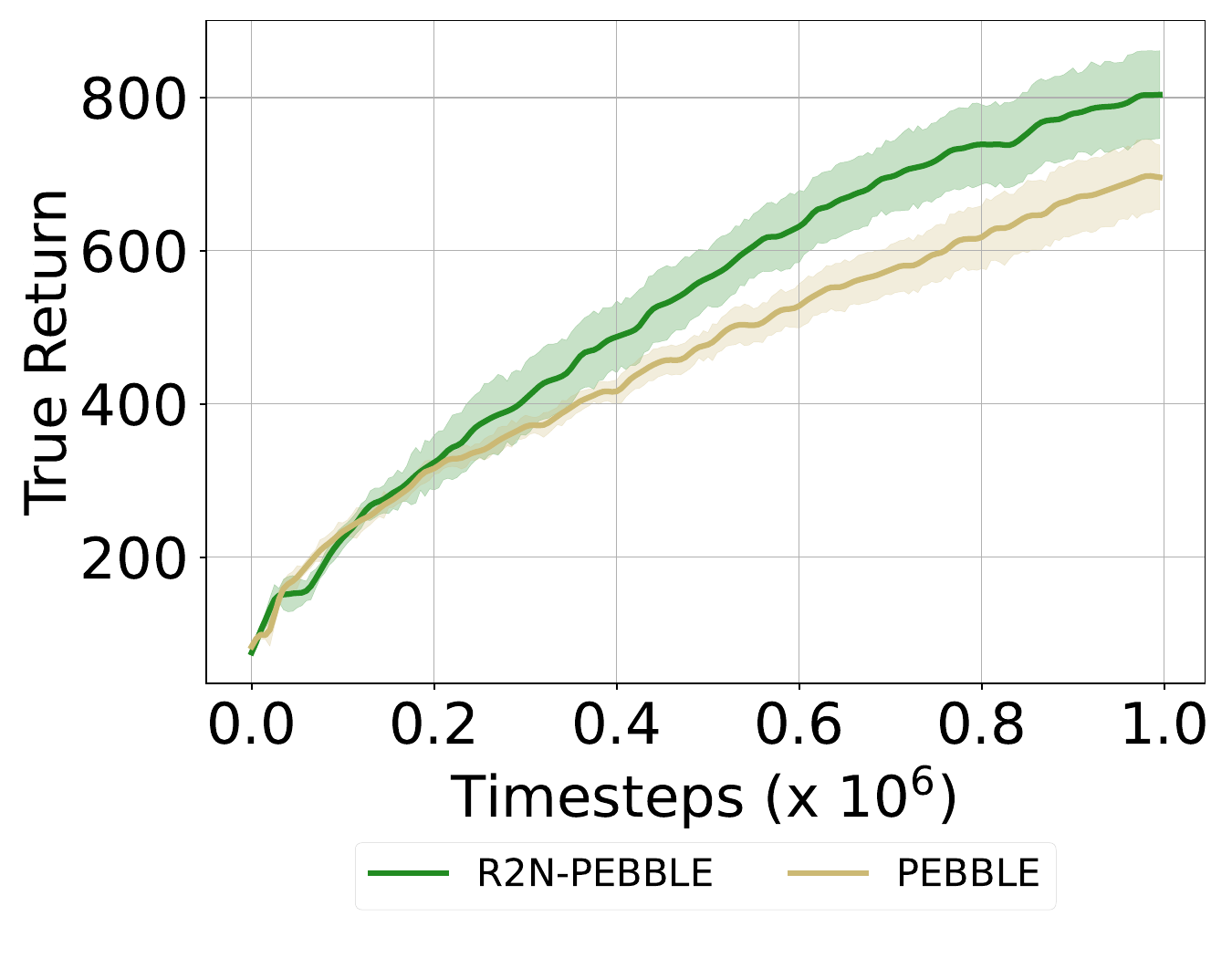}\label{fig:walker_walk_50_400}}
  \hfill
  \subfloat[Feedback = 1000]{\includegraphics[width=0.33\textwidth]{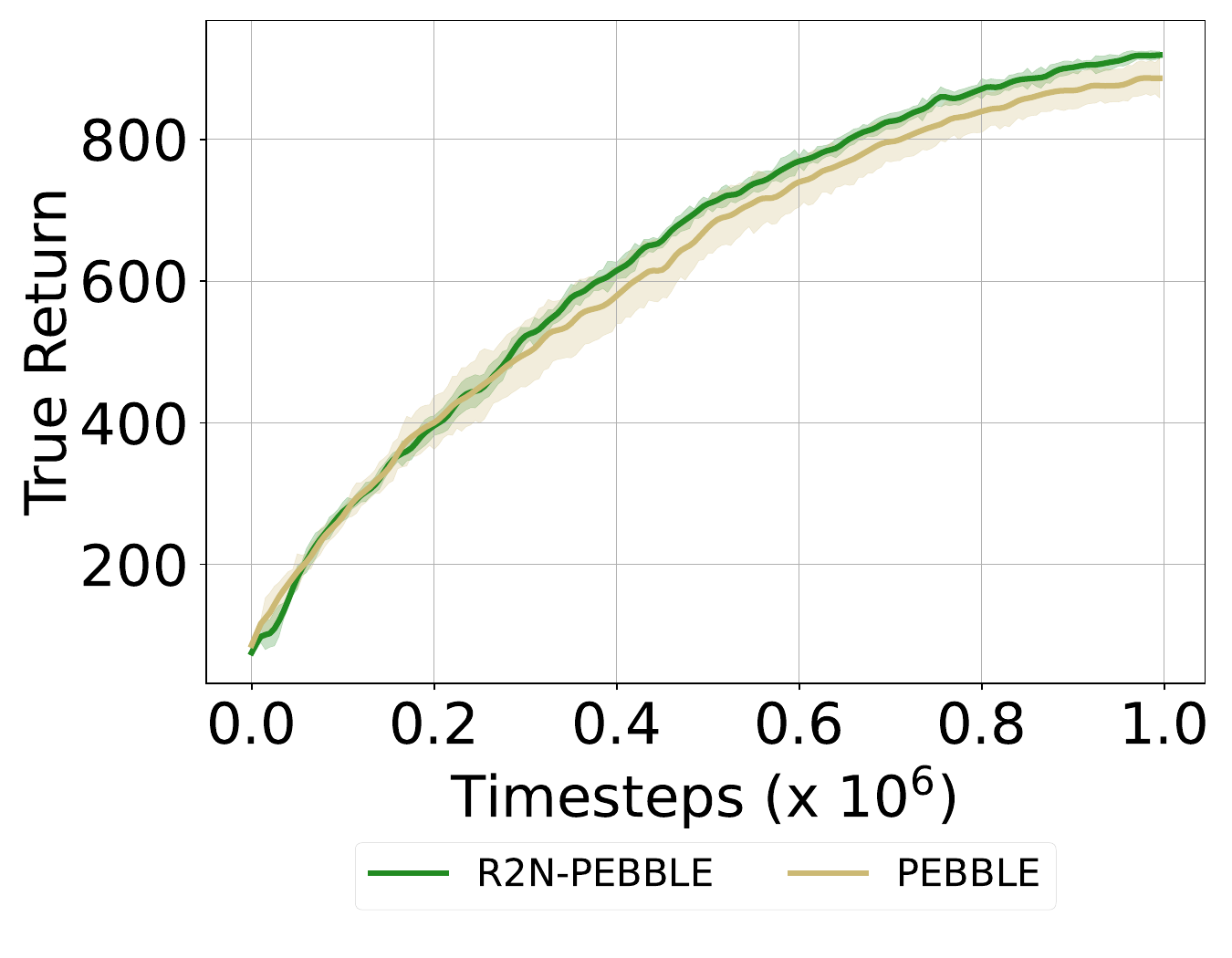}\label{fig:walker_walk_50_1000}}
    \hfill
  \subfloat[Feedback = 2000]{\includegraphics[width=0.33\textwidth]{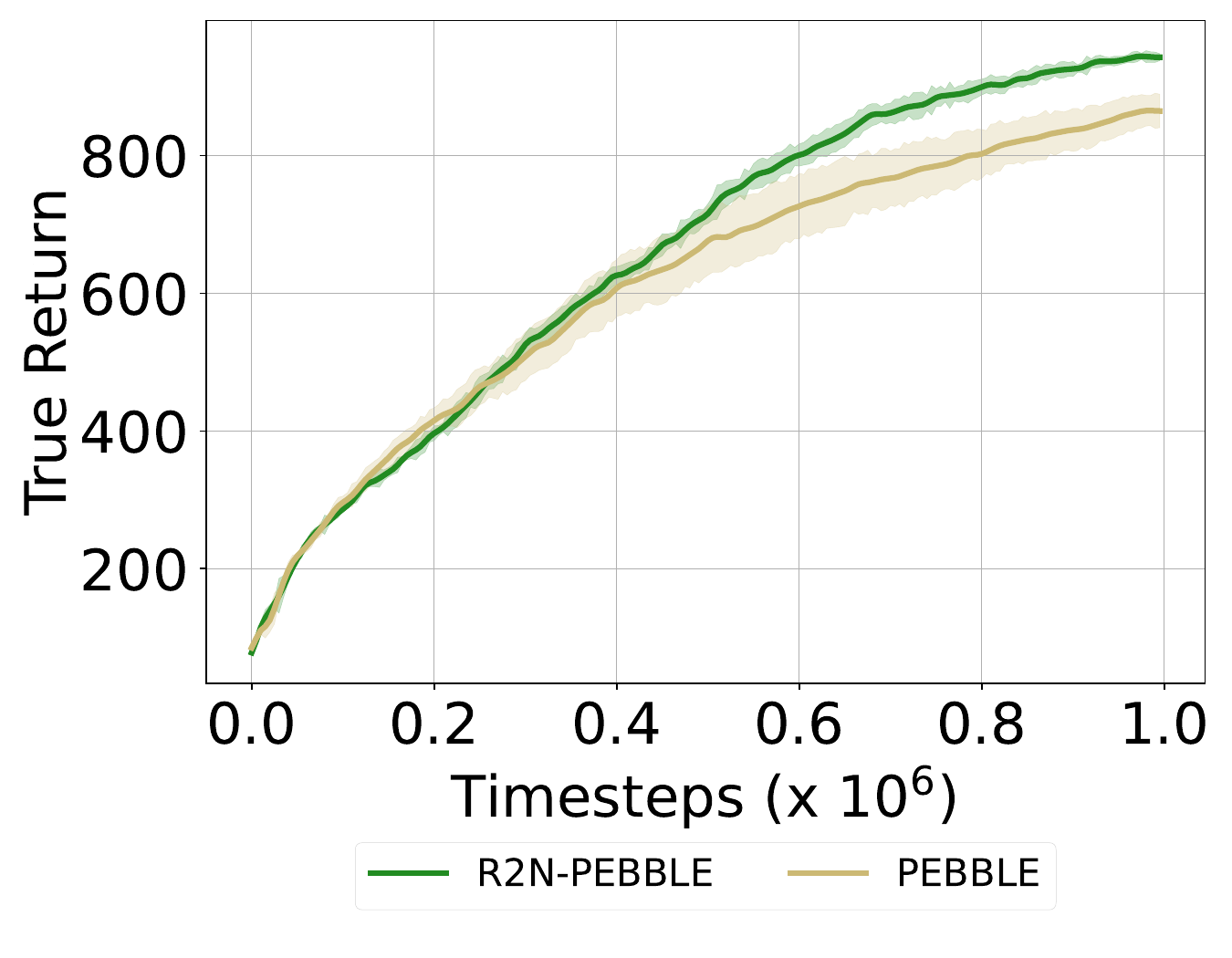}\label{fig:walker_walk_50_2000}}
    \hfill
  \subfloat[Feedback = 4000]{\includegraphics[width=0.33\textwidth]{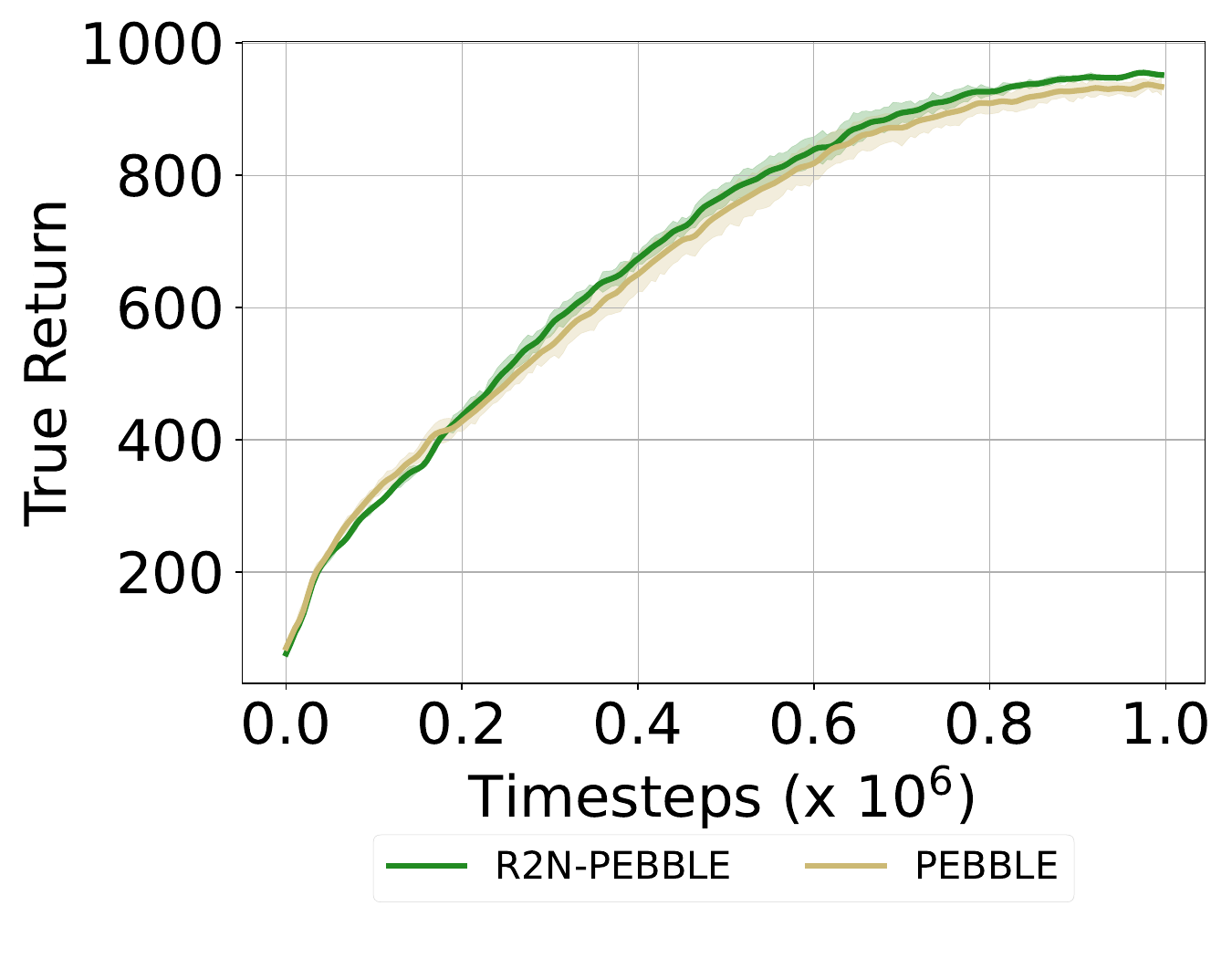}\label{fig:walker_walk_50_4000}}
    \hfill
  \subfloat[Feedback = 10000]{\includegraphics[width=0.33\textwidth]{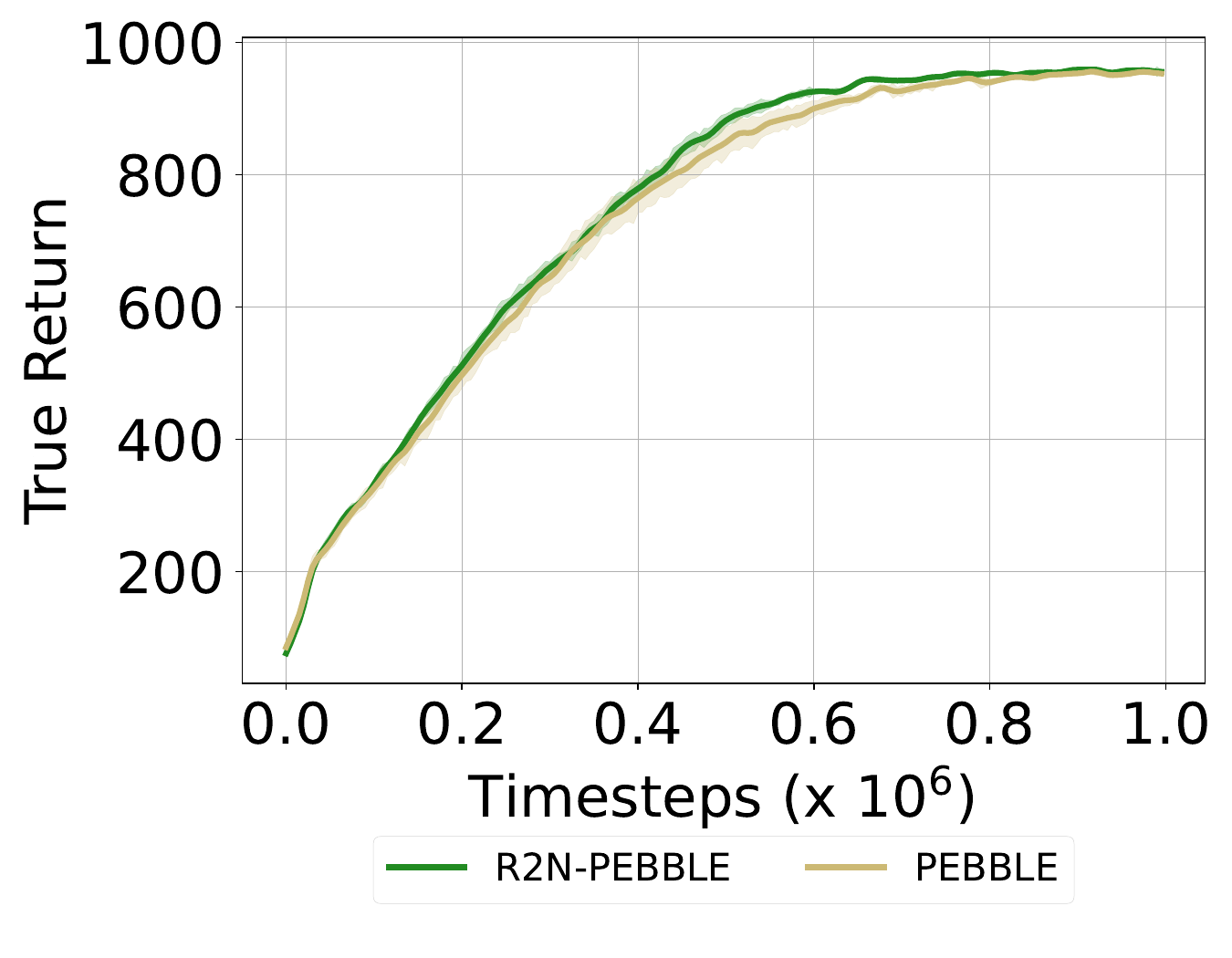}\label{fig:walker_walk_50_10000}}
   
  \caption{Walker-walk, Feedback Ablation, Noise = 50\%}\label{fig:walker_walk_fb_ablation_noise50}
\end{figure*}

\begin{figure*}[h!]
  \centering
 \subfloat[Feedback = 100]{\includegraphics[width=0.33\textwidth]{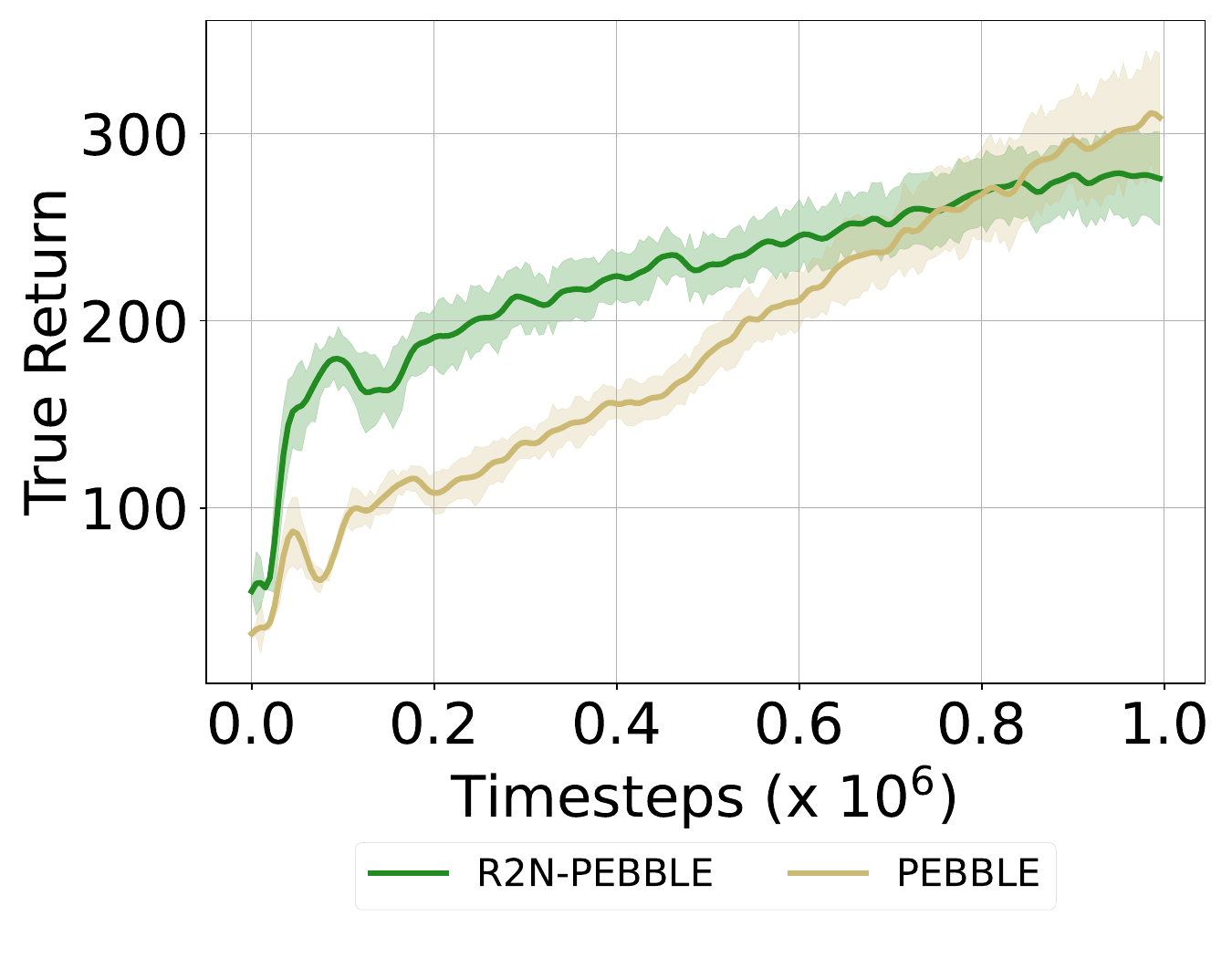}\label{fig:walker_walk_90_100}}
  \hfill
\subfloat[Feedback = 200]{\includegraphics[width=0.33\textwidth]{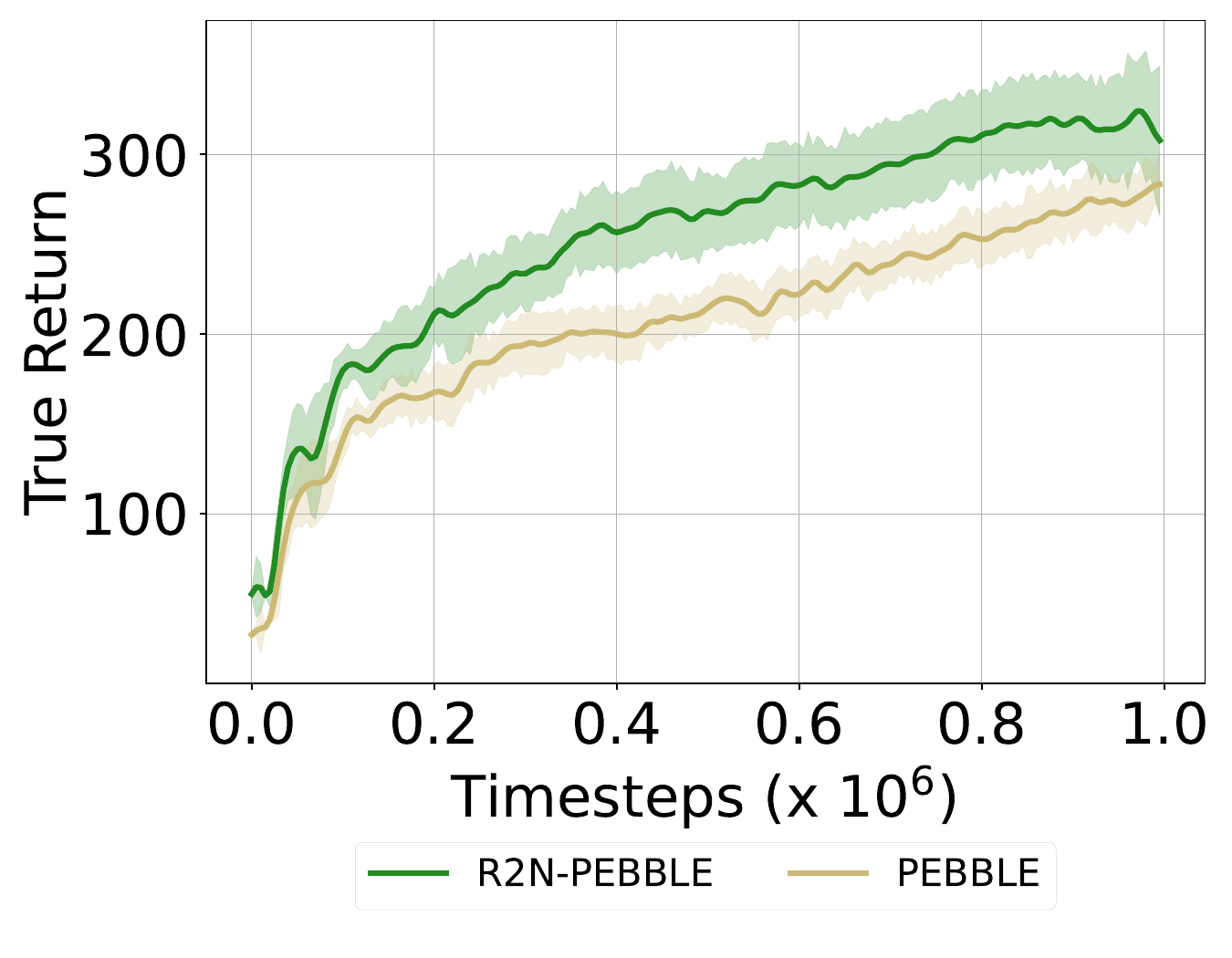}\label{fig:walker_walk_90_200}}
  \hfill
\subfloat[Feedback = 400]
{\includegraphics[width=0.33\textwidth]{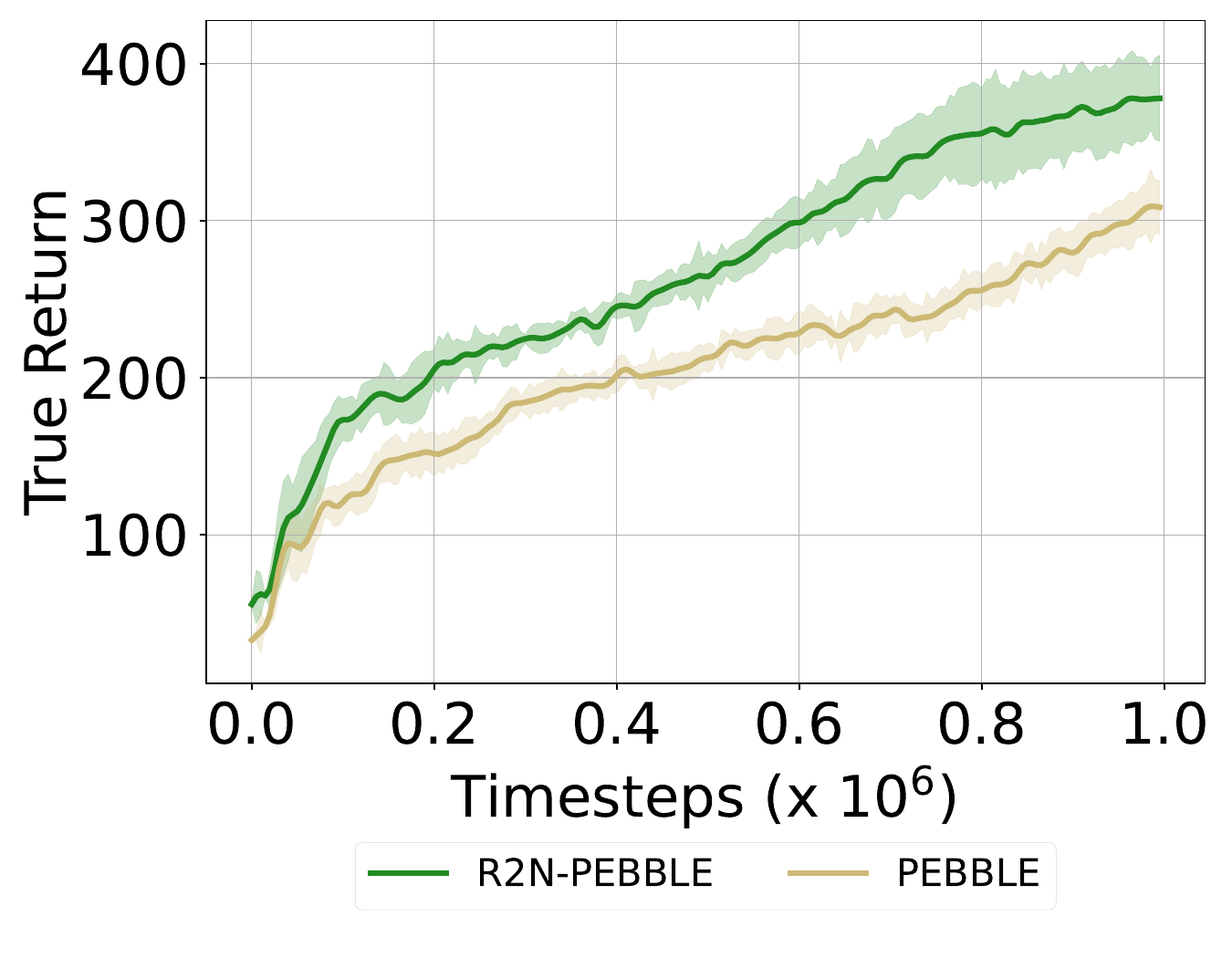}\label{fig:walker_walk_90_400}}
  \hfill
  \subfloat[Feedback = 1000]{\includegraphics[width=0.33\textwidth]{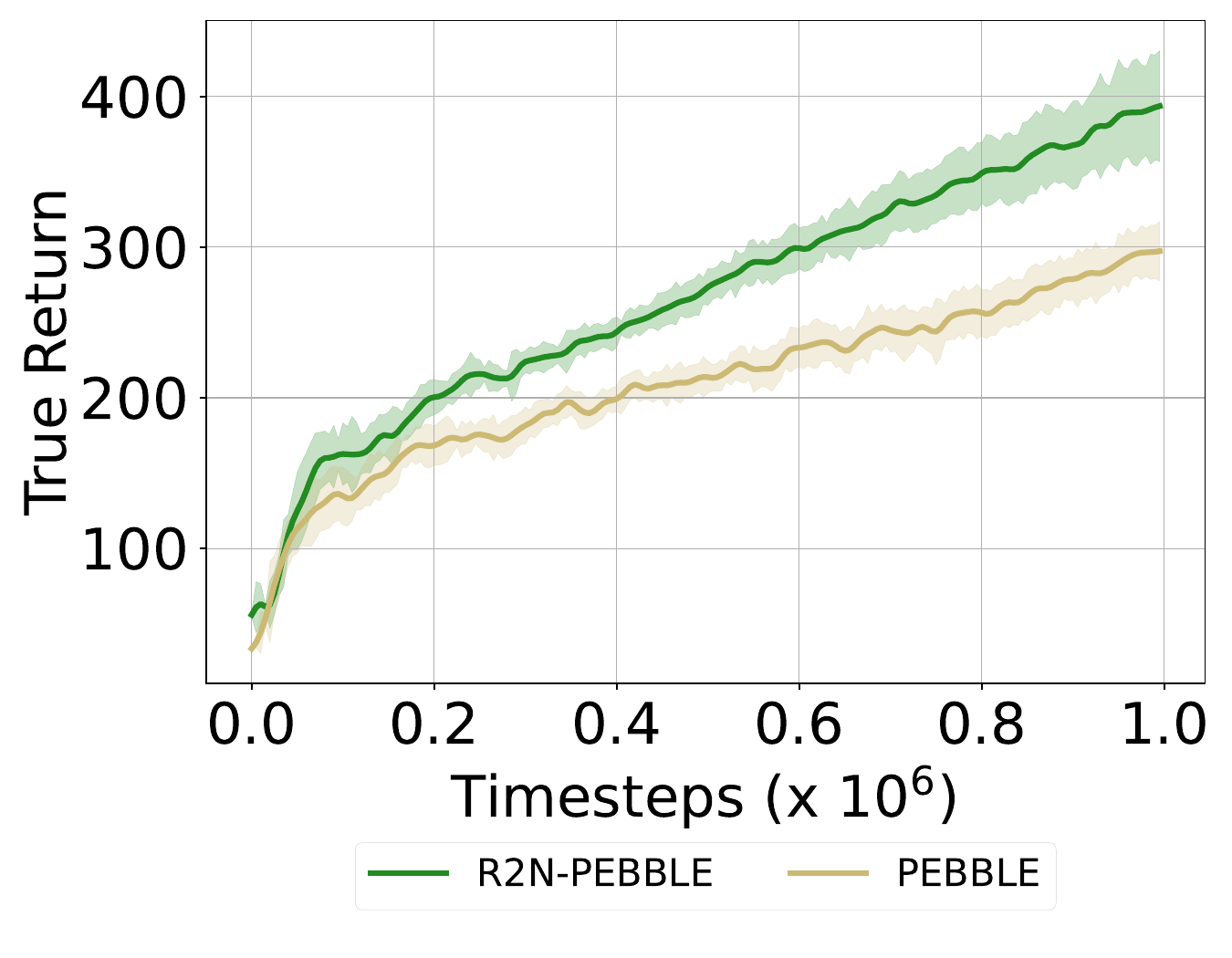}\label{fig:walker_walk_90_1000}}
    \hfill
  \subfloat[Feedback = 2000]{\includegraphics[width=0.33\textwidth]{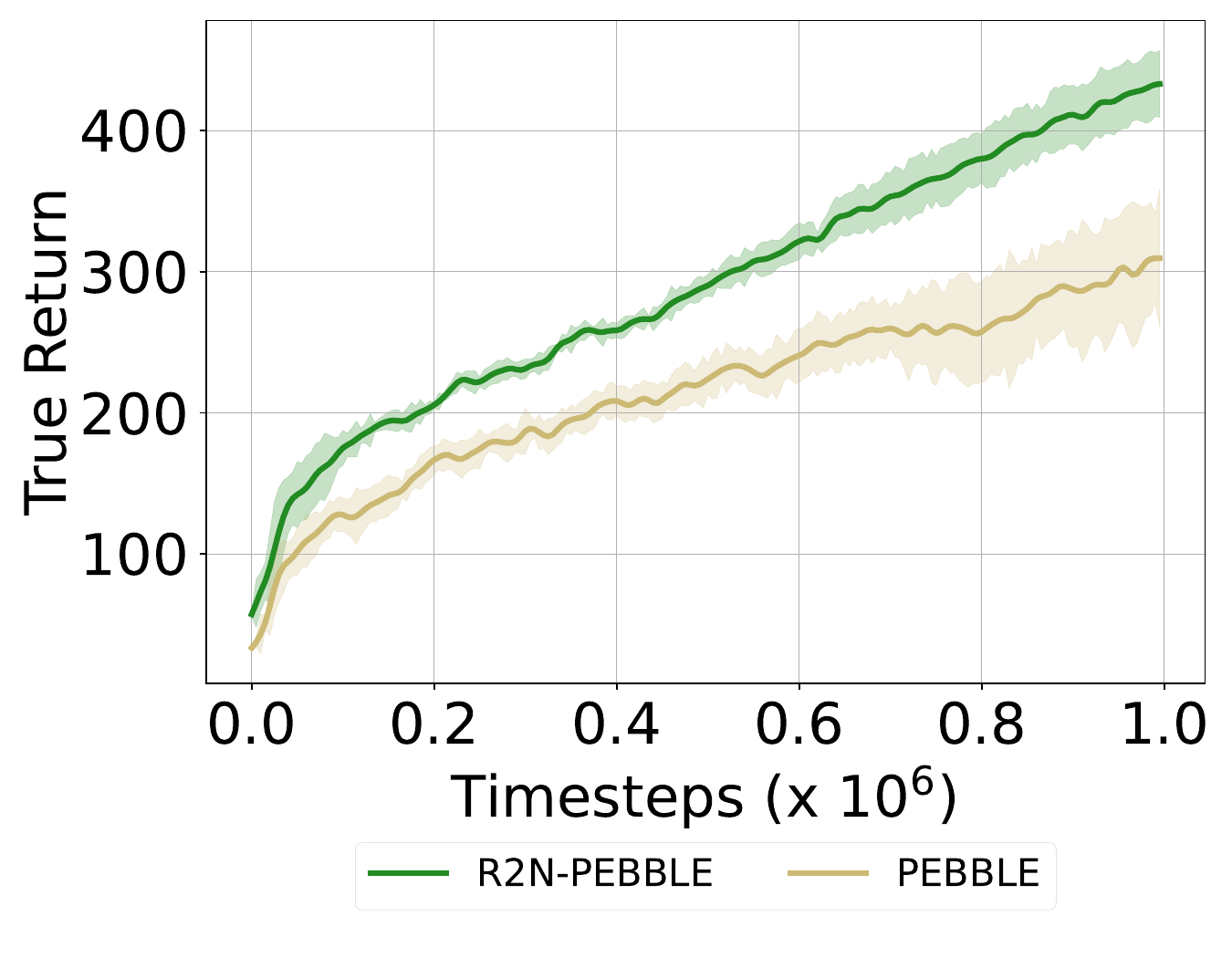}\label{fig:walker_walk_90_2000}}
    \hfill
  \subfloat[Feedback = 4000]{\includegraphics[width=0.33\textwidth]{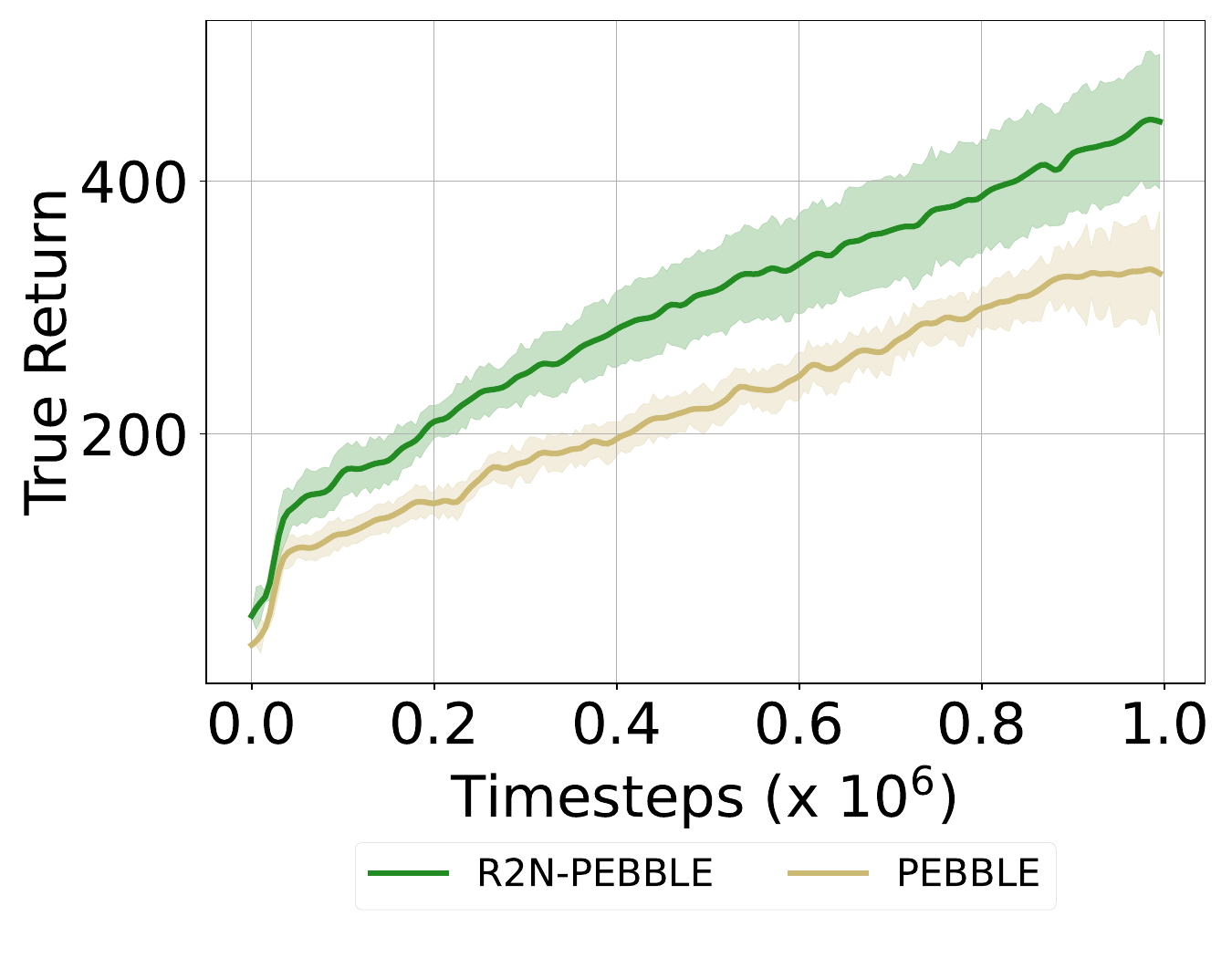}\label{fig:walker_walk_90_4000}}
    \hfill
  \subfloat[Feedback = 10000]{\includegraphics[width=0.33\textwidth]{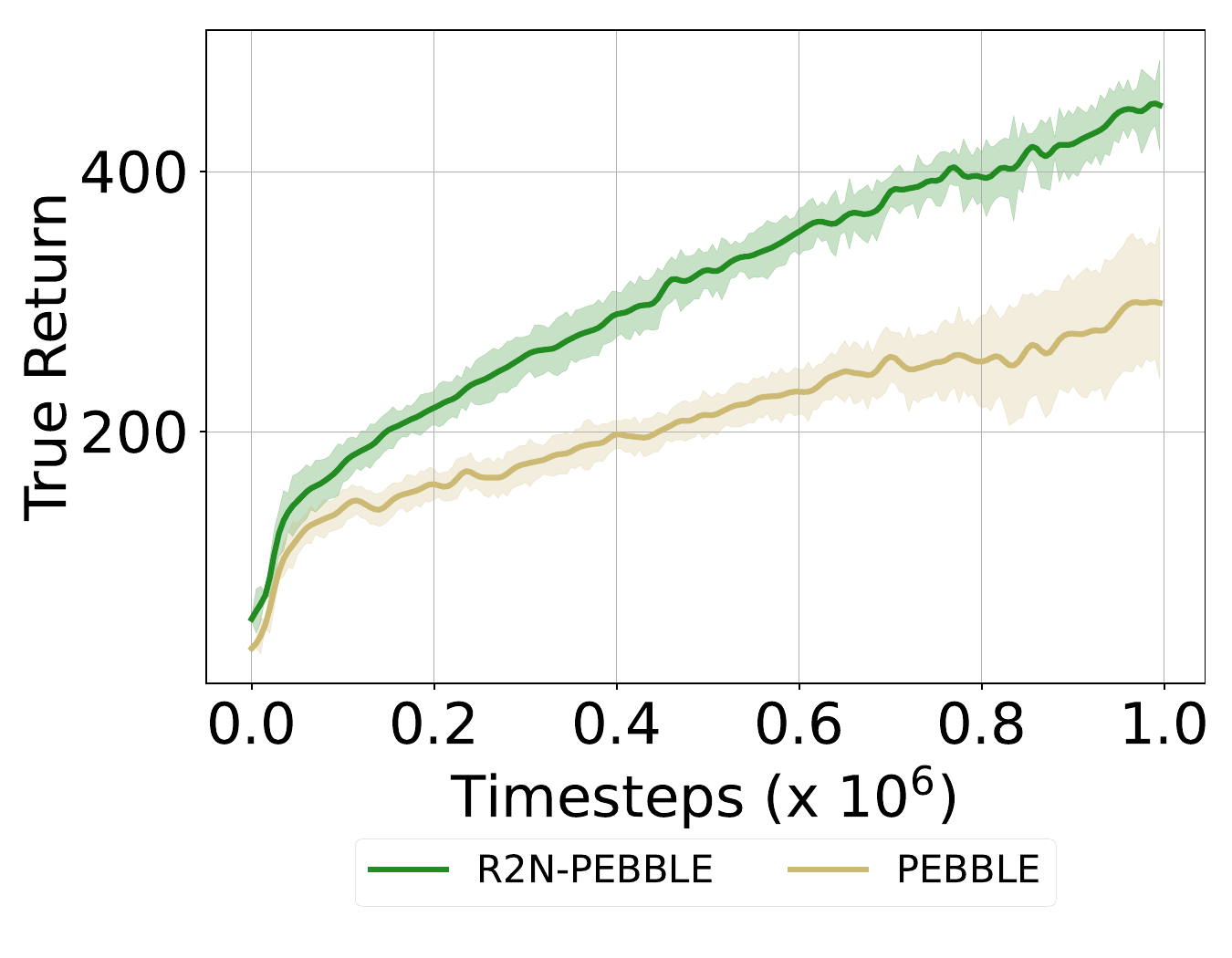}\label{fig:walker_walk_90_10000}}
   
  \caption{Walker-walk, Feedback Ablation, Noise = 90\%}\label{fig:walker_walk_fb_ablation_noise90}
\end{figure*}

\begin{figure*}[h!]
  \centering
 \subfloat[Feedback = 100]{\includegraphics[width=0.33\textwidth]{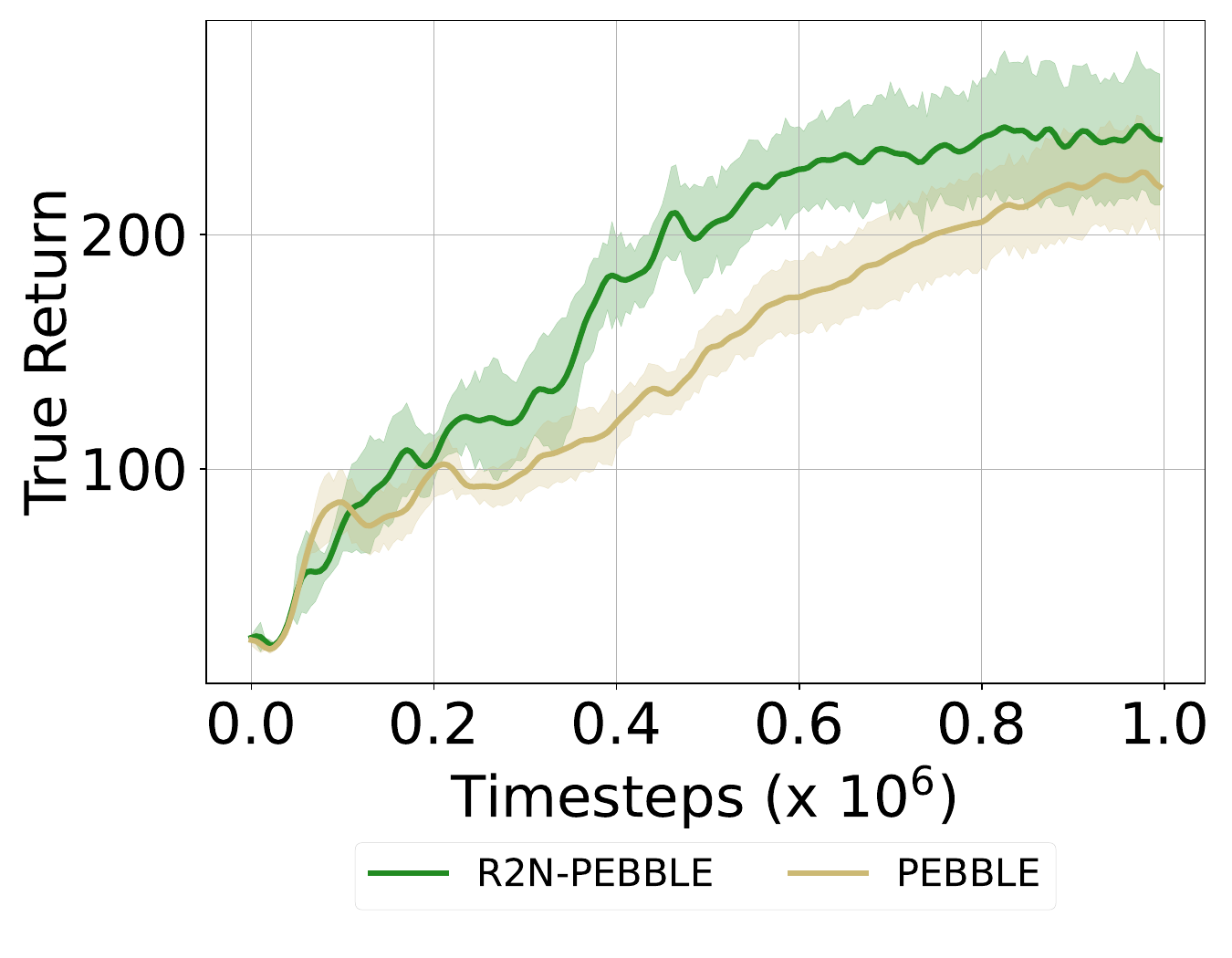}\label{fig:walker_walk_95_100}}
  \hfill
\subfloat[Feedback = 200]{\includegraphics[width=0.33\textwidth]{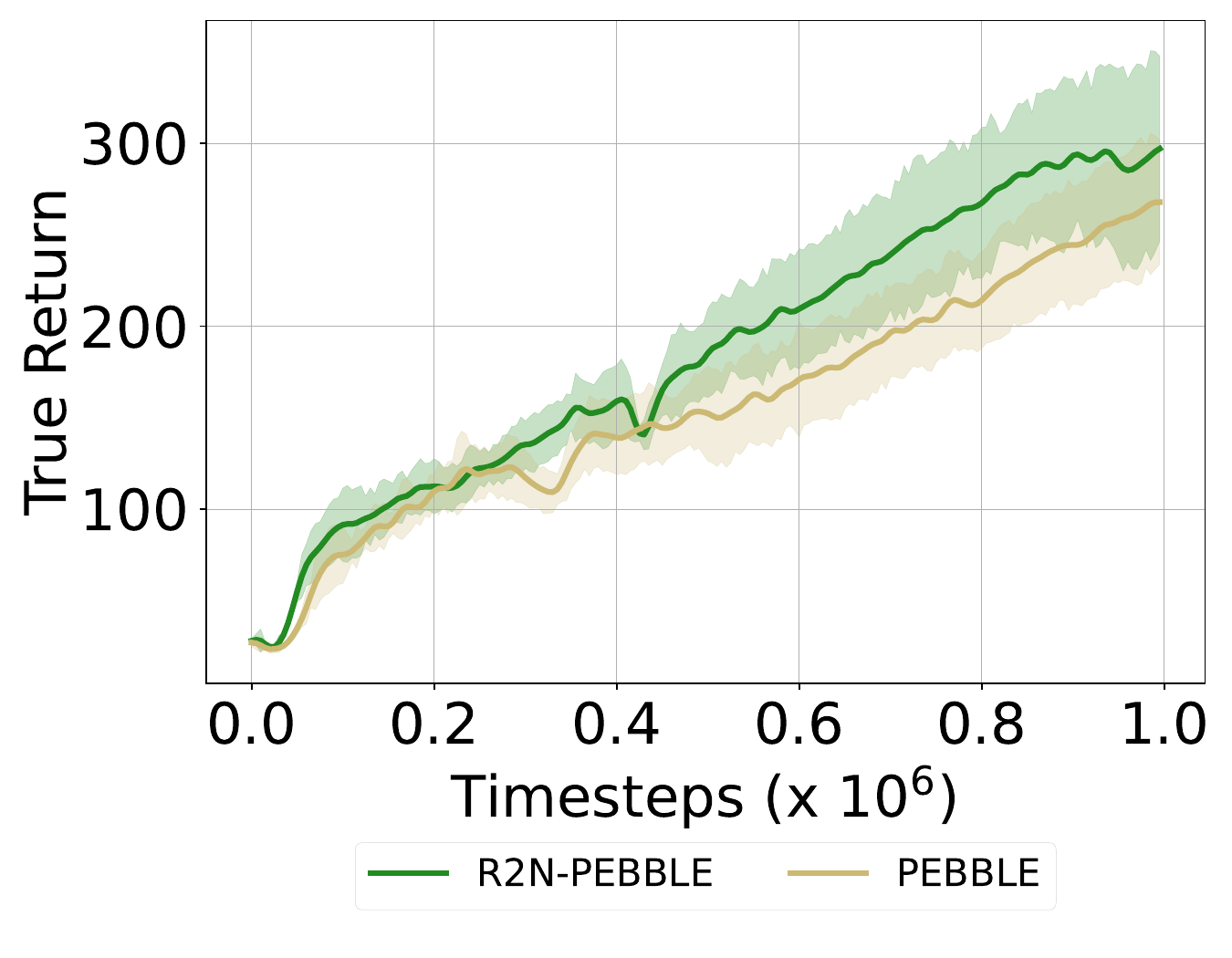}\label{fig:walker_walk_95_200}}
  \hfill
\subfloat[Feedback = 400]
{\includegraphics[width=0.33\textwidth]{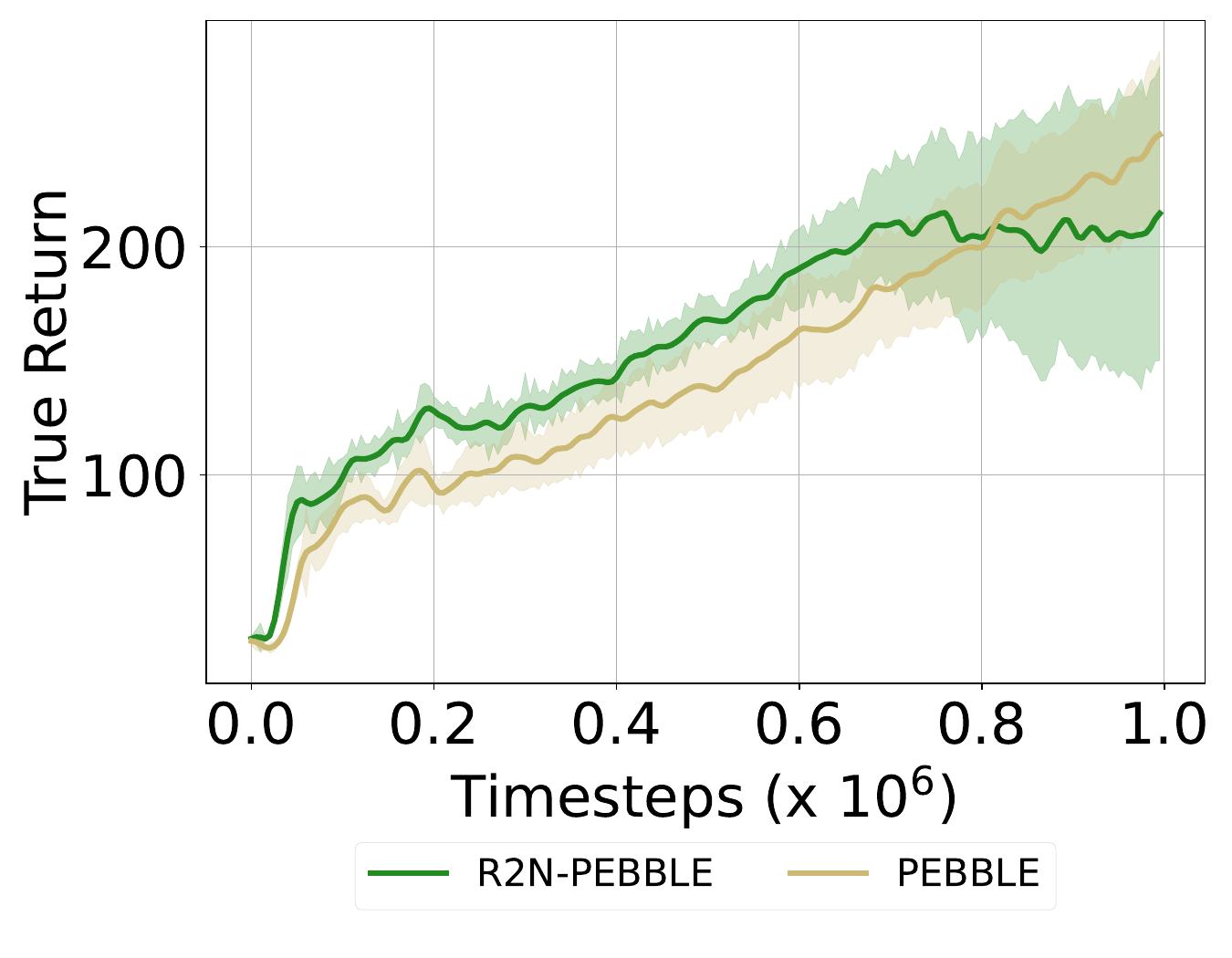}\label{fig:walker_walk_95_400}}
  \hfill
  \subfloat[Feedback = 1000]{\includegraphics[width=0.33\textwidth]{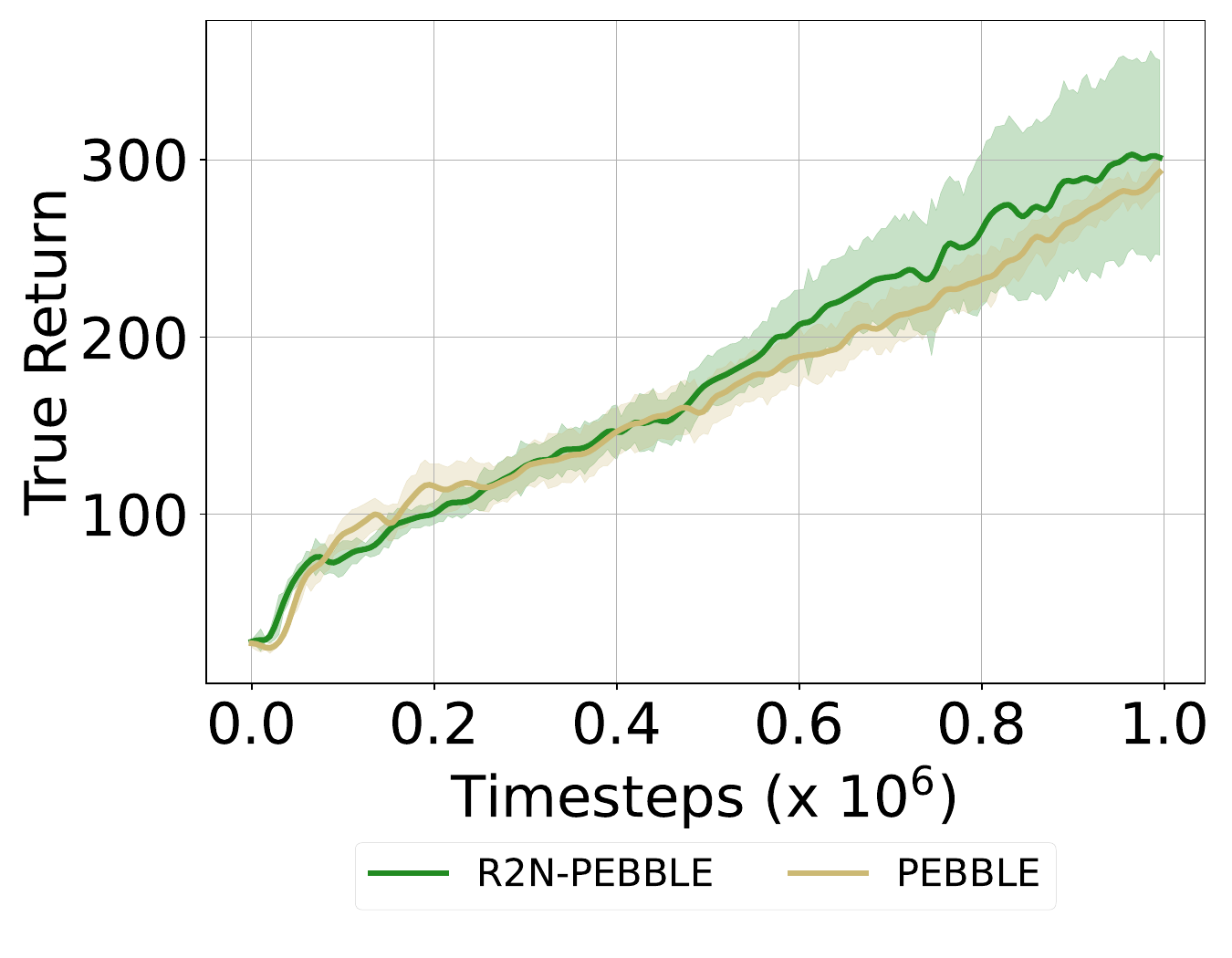}\label{fig:walker_walk_95_1000}}
    \hfill
  \subfloat[Feedback = 2000]{\includegraphics[width=0.33\textwidth]{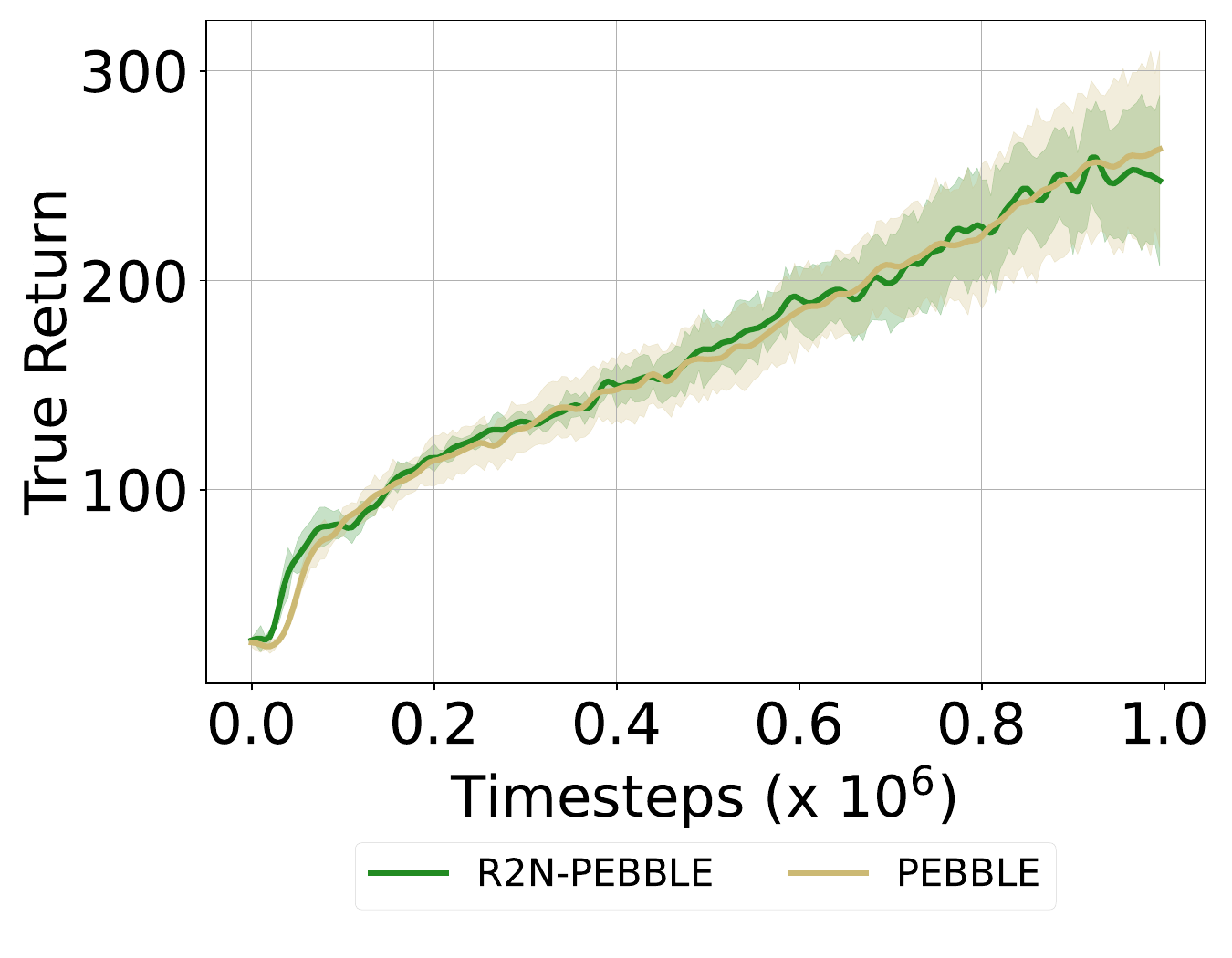}\label{fig:walker_walk_95_2000}}
    \hfill
  \subfloat[Feedback = 4000]{\includegraphics[width=0.33\textwidth]{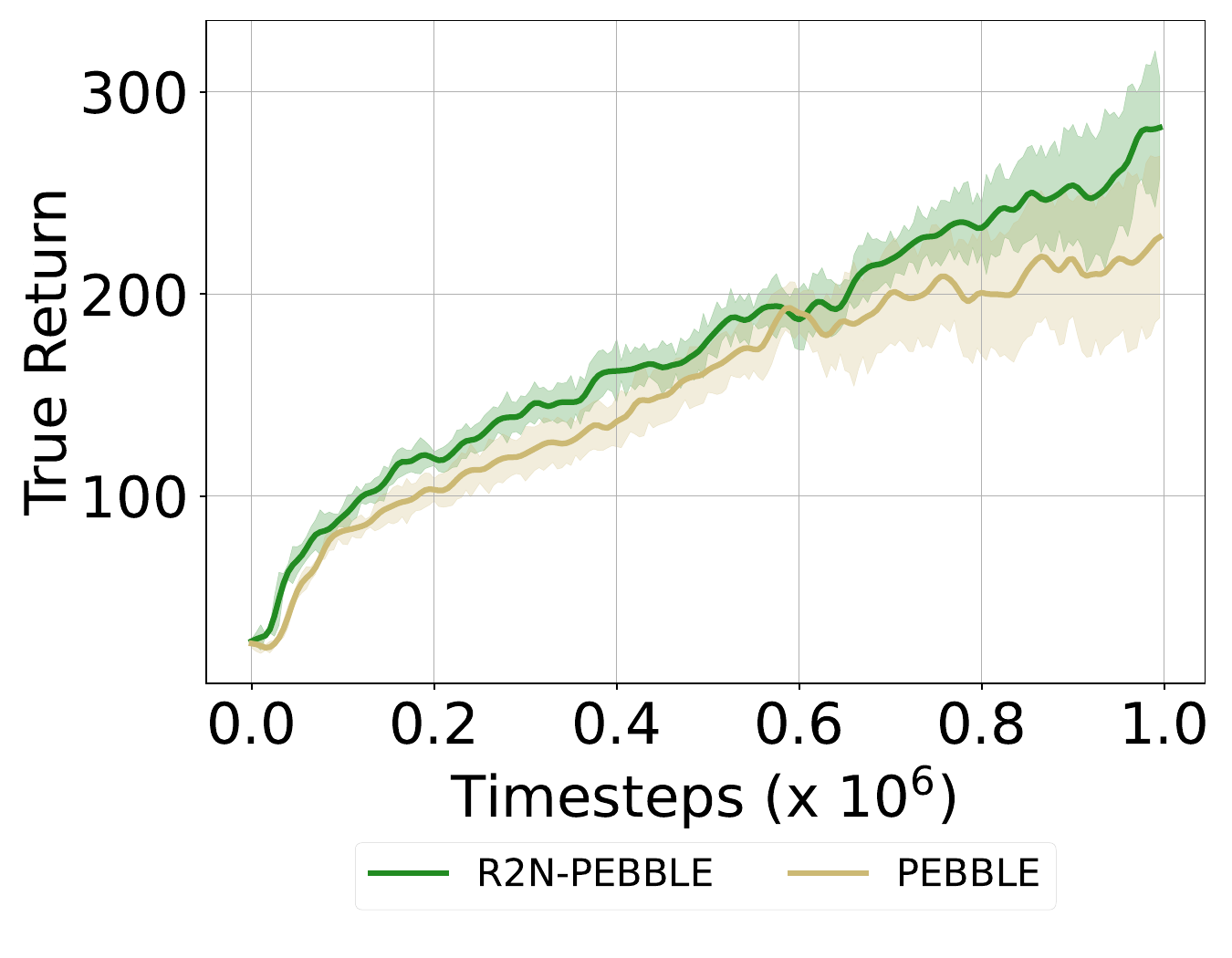}\label{fig:walker_walk_95_4000}}
    \hfill
  \subfloat[Feedback = 10000]{\includegraphics[width=0.33\textwidth]{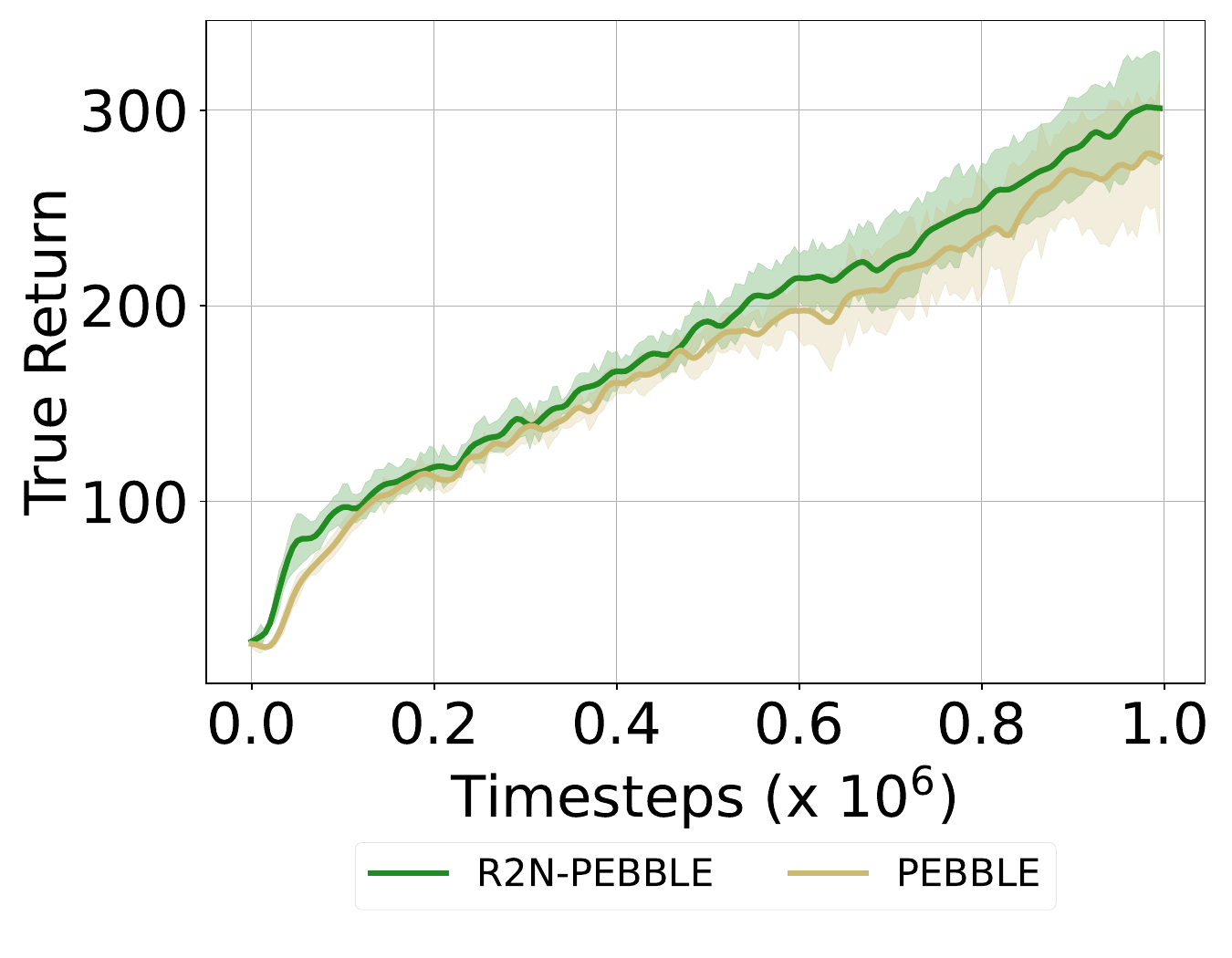}\label{fig:walker_walk_95_10000}}
   
  \caption{Walker-walk, Feedback Ablation, Noise = 95\%}\label{fig:walker_walk_fb_ablation_noise95}
\end{figure*}

In this section, we present additional experimental results ablating the noise levels and preference budgets for the Walker-walk environment.  Overall, in Figures \ref{fig:walker_walk_fb_ablation_noise20} through \ref{fig:walker_walk_fb_ablation_noise95}, we find that R2N-PEBBLE (green curves) is generally more robust than PEBBLE (yellow curves) in varying noise levels and feedback amounts.

\subsection{Comparison of R2N with Traditional RL Algorithms}
\label{sec:r2n_vs_rl_algs}

\begin{figure*}[h!]
  \centering
 \subfloat[Walker-walk, Noise=90\%]{\includegraphics[width=0.33\textwidth]{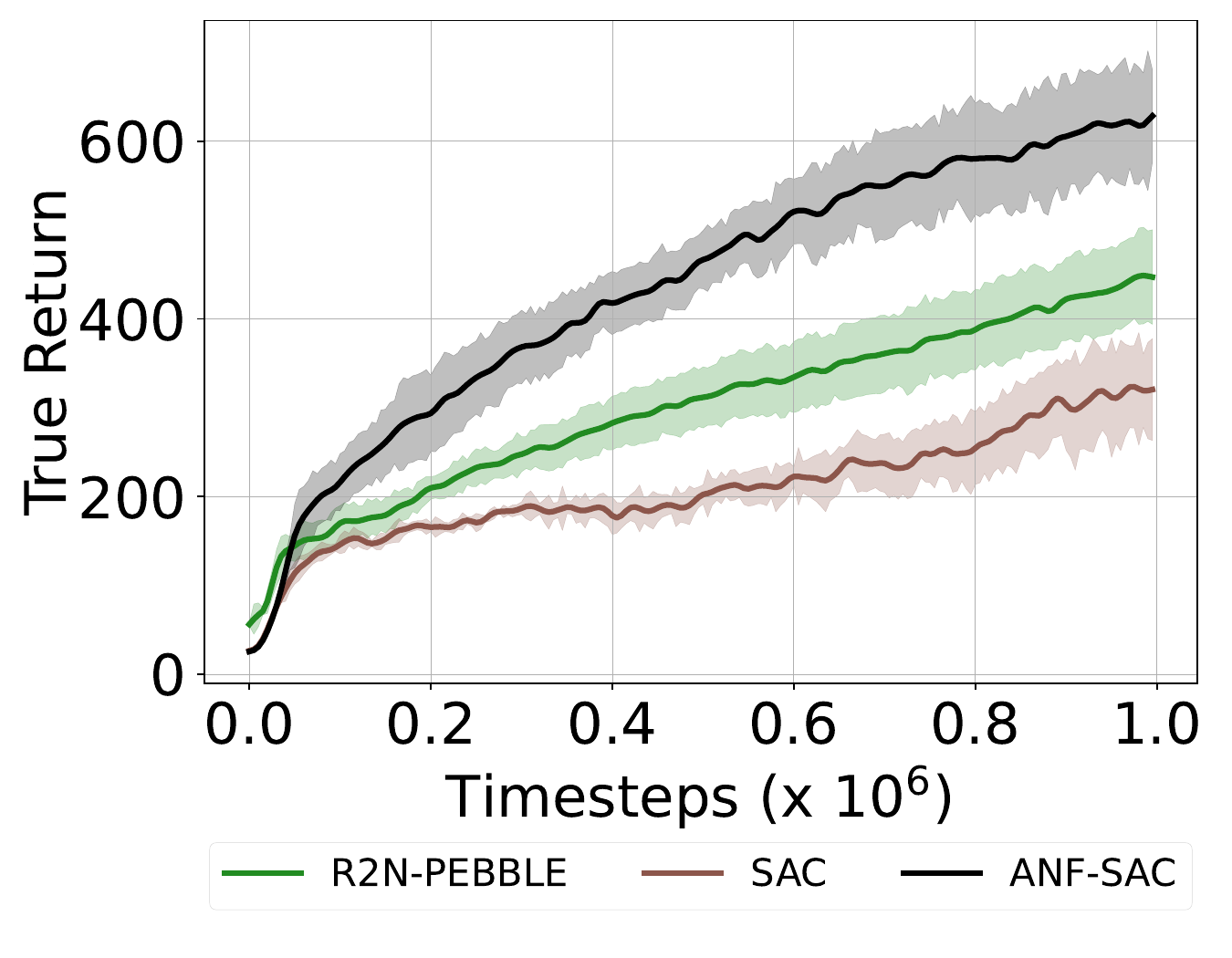}\label{fig:walker_walk_env_reward}}
  \hfill   
    \centering
 \subfloat[Cheetah-run, Noise=90\%]{\includegraphics[width=0.33\textwidth]{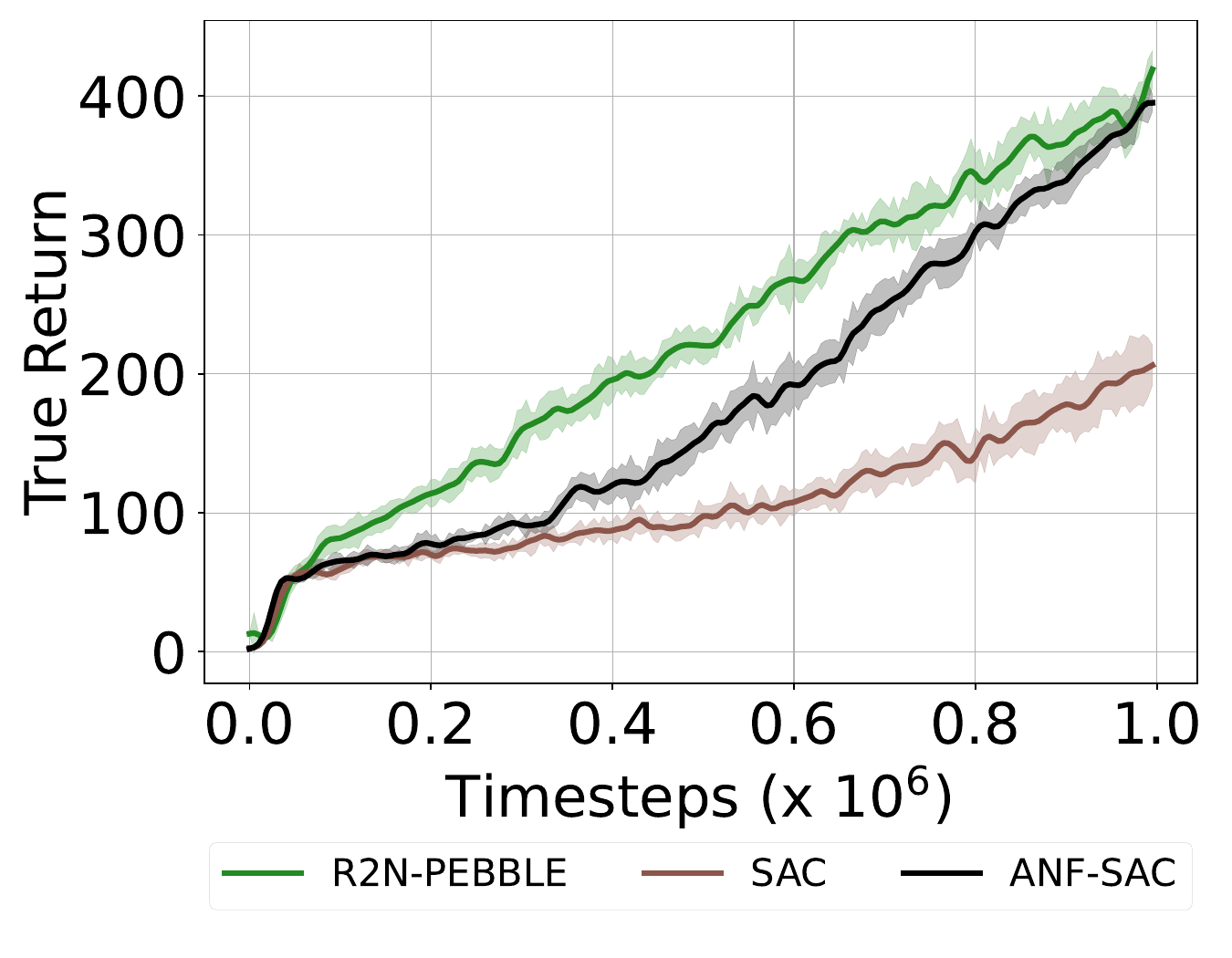}\label{fig:cheetah_run_env_reward}}
  \hfill   
    \centering
 \subfloat[Cartpole-swingup, Noise=90\%]{\includegraphics[width=0.33\textwidth]{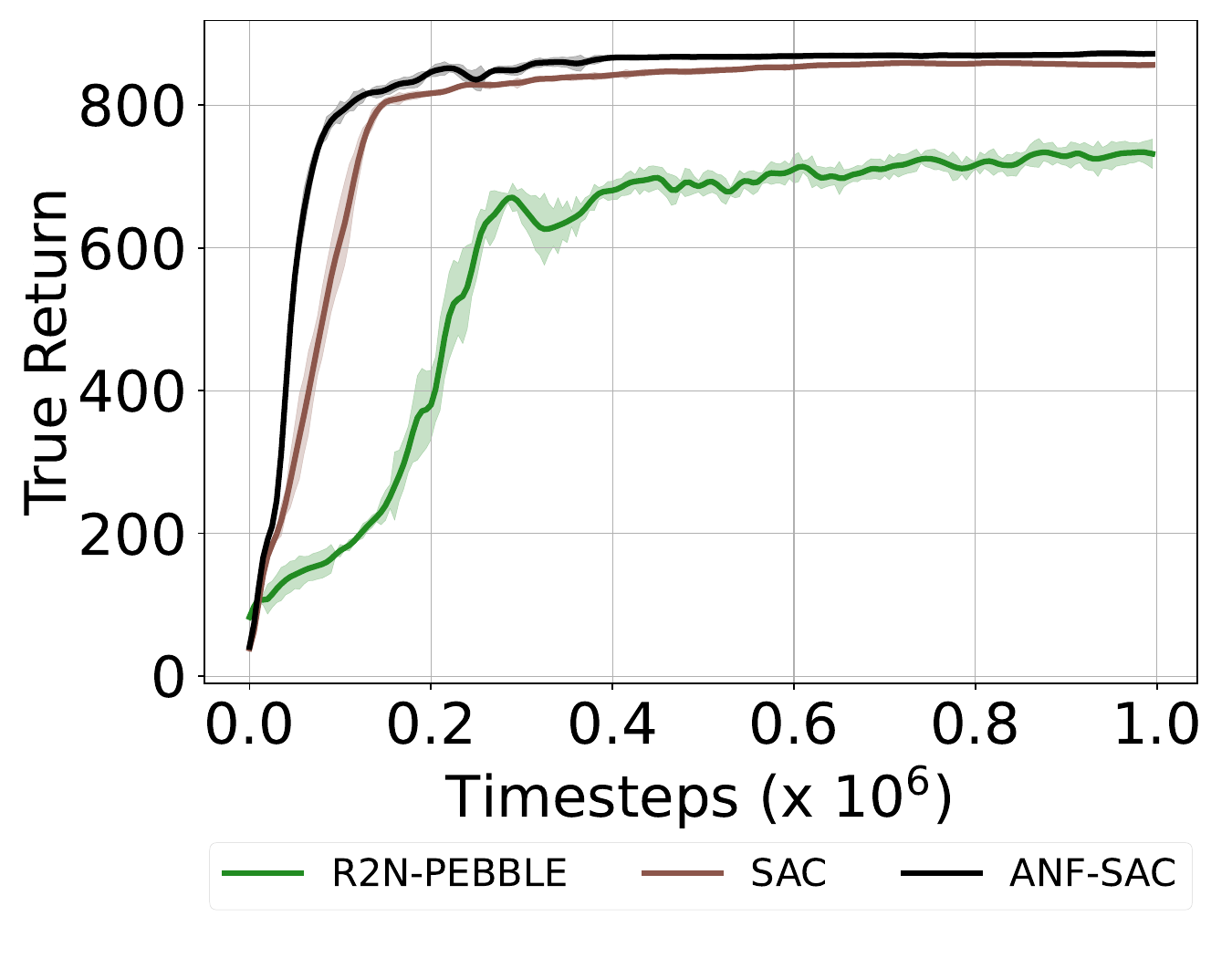}\label{fig:cartpole_env_reward}}
  \hfill   
      \centering
 \subfloat[Humanoid-stand, Noise=70\%]{\includegraphics[width=0.33\textwidth]{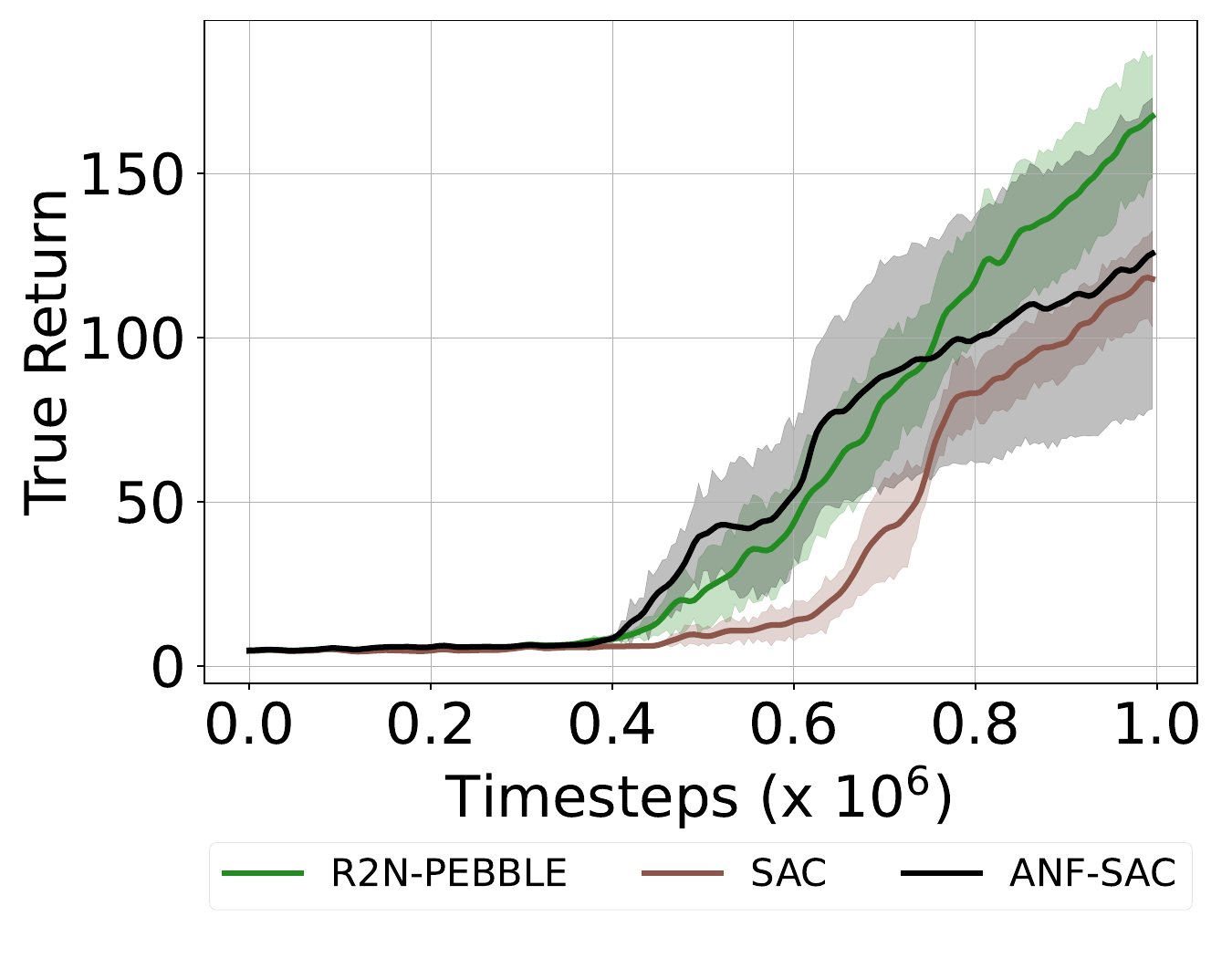}\label{fig:humanoid_stand_env_reward}}
  \hspace{2em}
        \centering
 \subfloat[Quadruped-walk, Noise=70\%]{\includegraphics[width=0.33\textwidth]{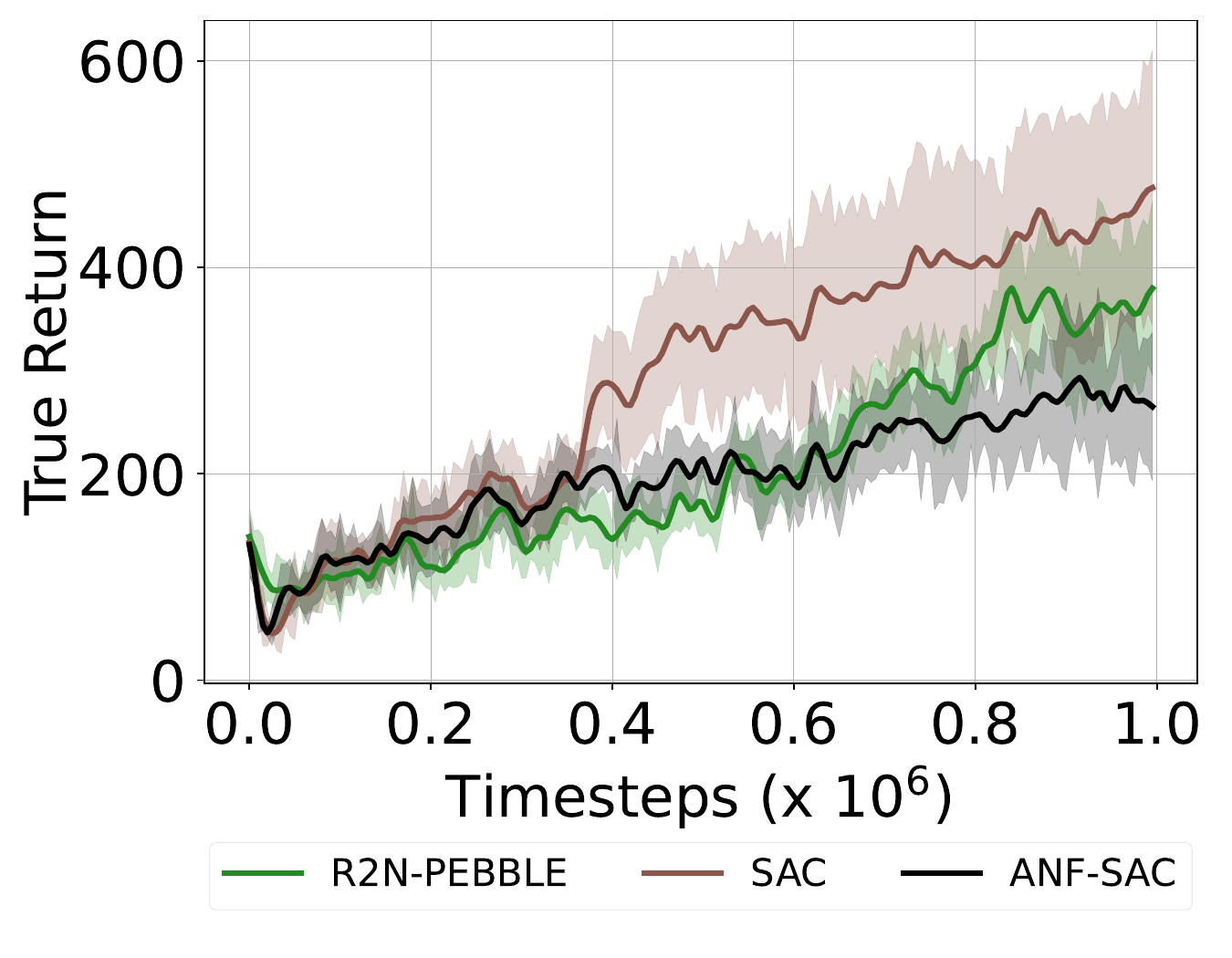}\label{fig:quadruped_walk_env_reward}}
  \hfill   
  \caption{These plots compare R2N-PEBBLE with SAC and ANF-SAC, algorithms that have access to the ground truth reward function.}\label{fig:env_reward_baselines}
\end{figure*}

In this section, we show additional results comparing R2N-PEBBLE (green curve) with two RL algorithms, SAC (brown curve) and ANF-SAC (black curve). We note that both these algorithms learn while accessing the true environmental reward which is unavailable to R2N. Impressively, in Figure \ref{fig:env_reward_baselines}, we find that in three out of the five tested DMControl environments, R2N can achieve comparable results with SAC and ANF-SAC.

\vspace{1em}

\subsection{Imitating Noise Analysis}
\label{sec:imitating_noise_study}

\begin{figure*}[h!]
 \subfloat[Cheetah-run, Noise=90\%, Feedback=10000]{\includegraphics[width=0.5\textwidth]{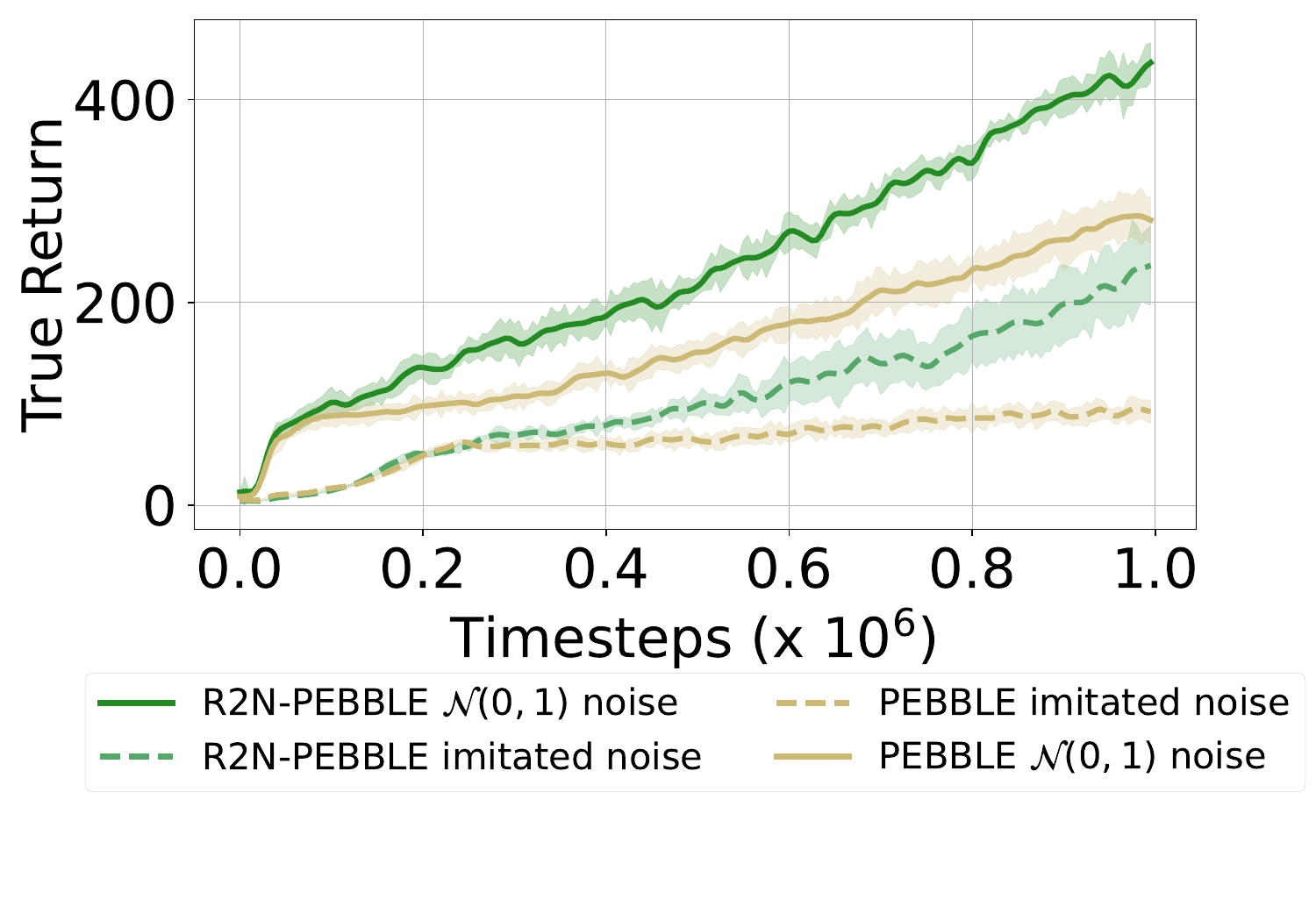}\label{fig:cheetah_run_imiate_noise_10k}}
  \hfill   
 \subfloat[Walker-walk, Noise=90\%, Feedback=4000]{\includegraphics[width=0.5\textwidth]{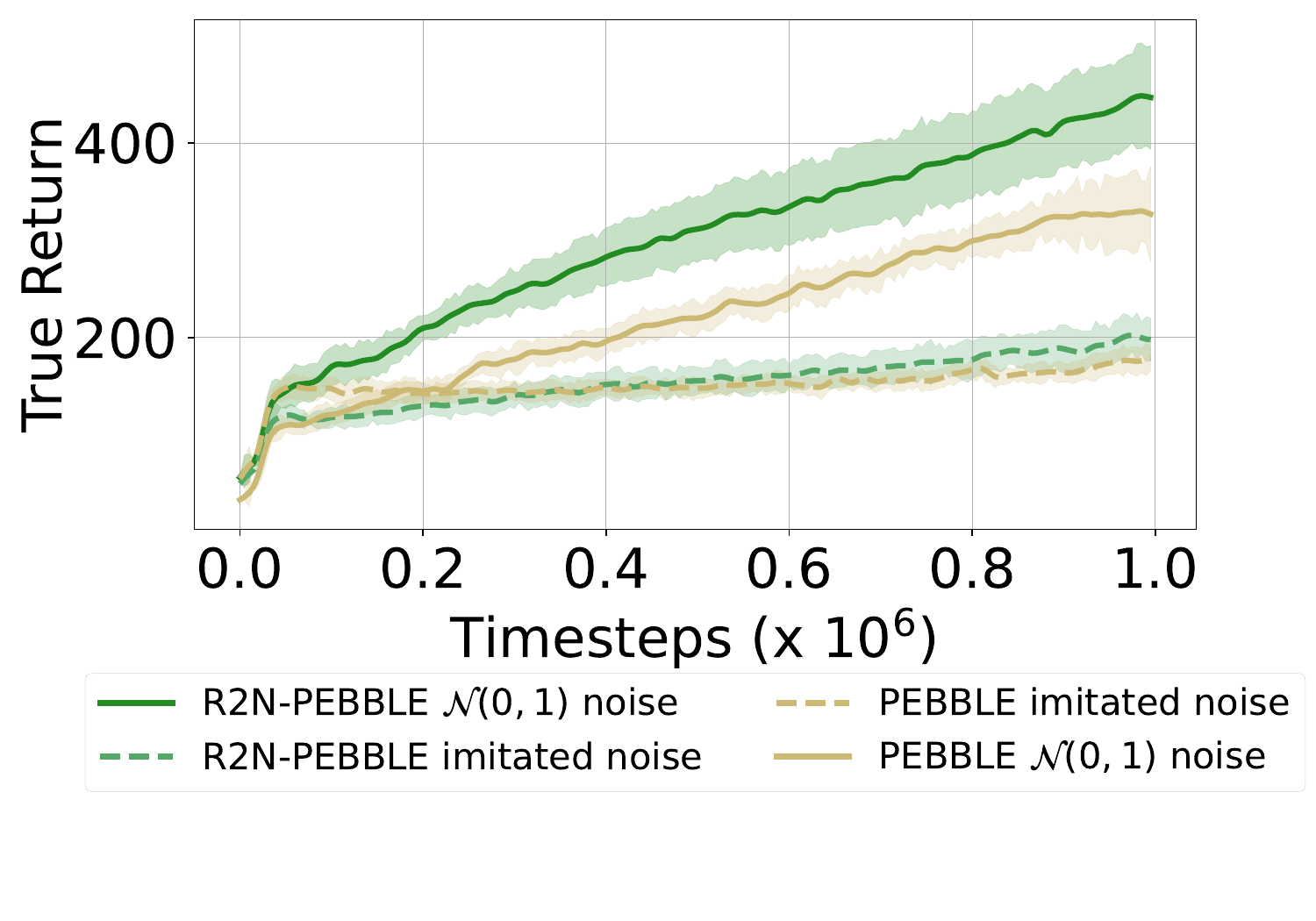}\label{fig:walker_walk_imiate_noise_4k}}
  \hfill   
  \centering
 \subfloat[Walker-walk, Noise=90\%, Feedback=10000]{\includegraphics[width=0.5\textwidth]{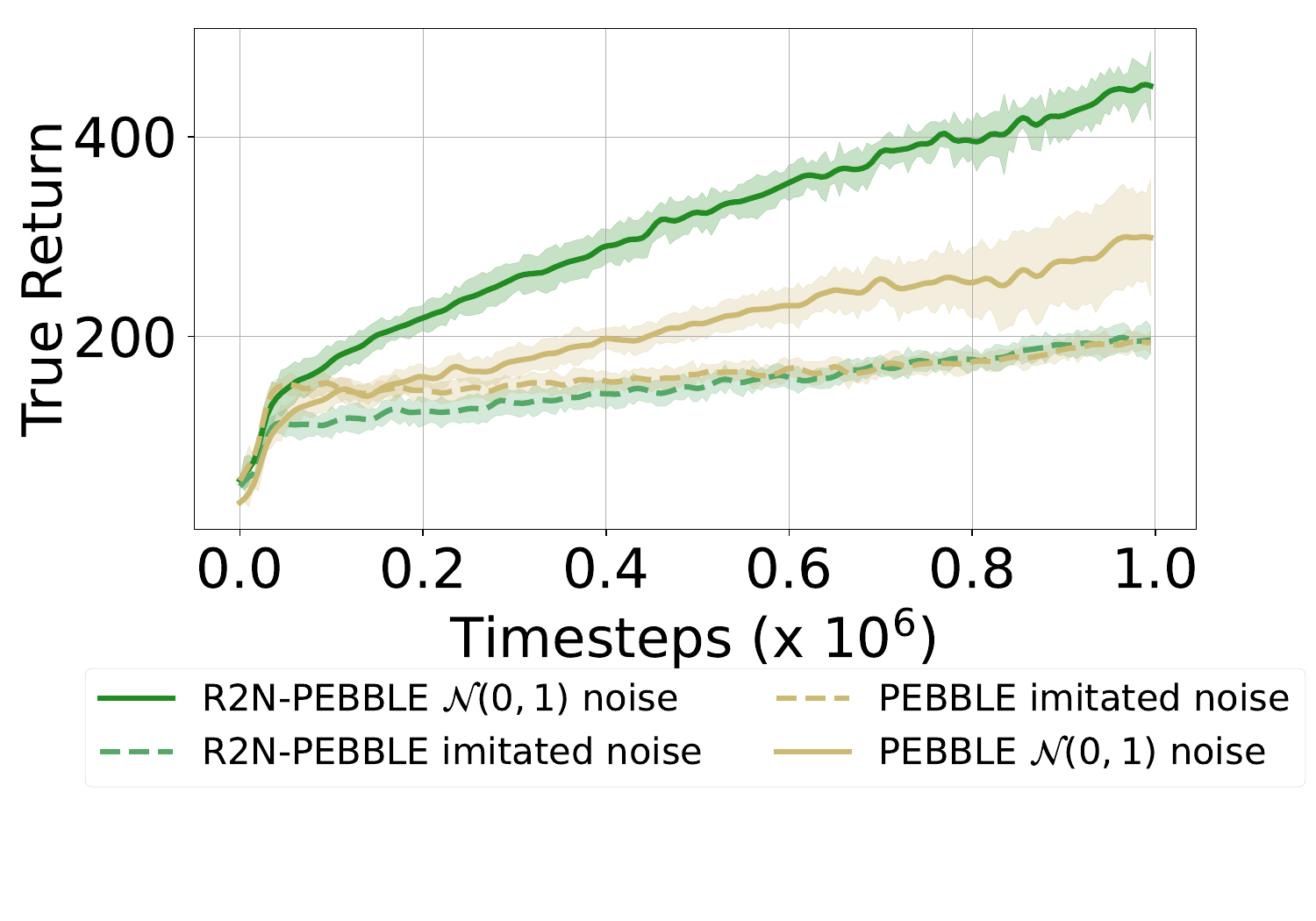}\label{fig:walker_walk_imiate_noise_10k}}
  \par 
  \captionsetup{skip=-0.2em}
  \caption{These plots compare the performance of R2N-PEBBLE and PEBBLE in the ENE setting where the noise features imitate the task-relevant features.}\label{fig:imitate noise}
\end{figure*}

In this section, we show additional results of the experiment where noise features imitate the distribution of real features, as described in Section \ref{sec:imitation}. We find that in the Cheetah-run environment, R2N-PEBBLE (green dotted curve) maintains performance gains over PEBBLE (yellow dotted curve) when we increase the preference budget to 10000 (see Figure \ref{fig:cheetah_run_imiate_noise_10k}). However, we find that for Walker-walk, R2N-PEBBLE performs comparably to PEBBLE for preference budgets of 4000 and 10000 (see Figures \ref{fig:walker_walk_imiate_noise_4k} and \ref{fig:walker_walk_imiate_noise_10k}).

\vspace{1em}

\subsection{DST Component Analyisis}\label{sec:dst_comp_study}
\begin{figure*}[h!]
  \centering
 \includegraphics[width=0.49\textwidth]{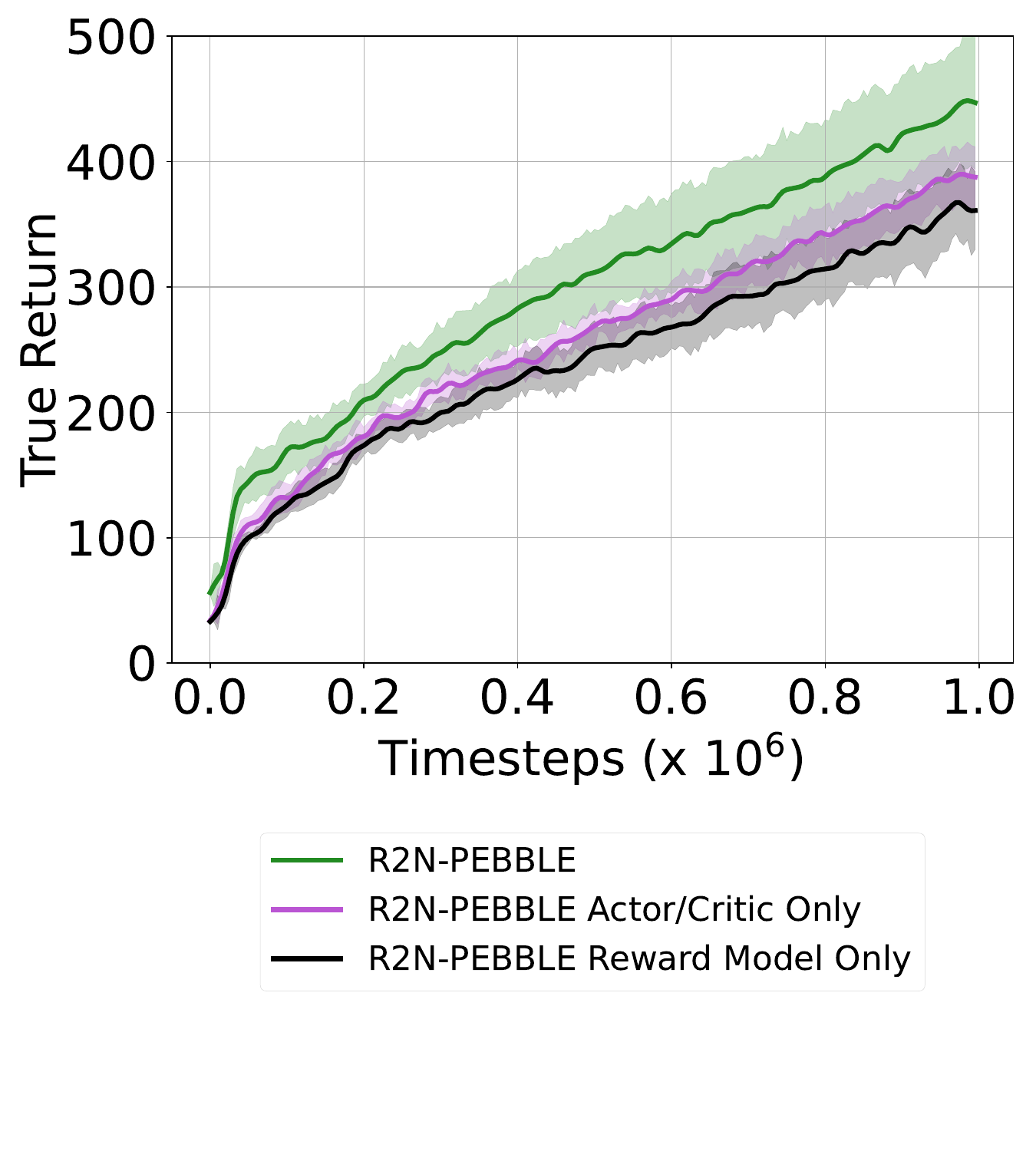}
  \captionsetup{width=0.8\textwidth, skip=-2em}
  \caption{This plot shows the performance of R2N-PEBBLE in Walker-walk (noise = 90\%, feedback = 4000) with R2N variants that only apply DST to either the RL modules (actor/critic) or the reward learning module.}\label{fig:walker_walk_dst}
\end{figure*}

In this section, we aim to understand the importance of dynamic sparse training on both reward learning and RL modules. In Figure \ref{fig:walker_walk_dst}, we find that in Walker-walk, full R2N which applies DST to both learning modules is superior to R2N variants that only apply DST to one learning module. 

\subsection{R2N on Zero Noise Environments}\label{sec:zero_noise_exp}
\begin{figure*}[h!]
  \centering
 \subfloat[Cheetah-run, Feedback=1000]{\includegraphics[width=0.5\textwidth]{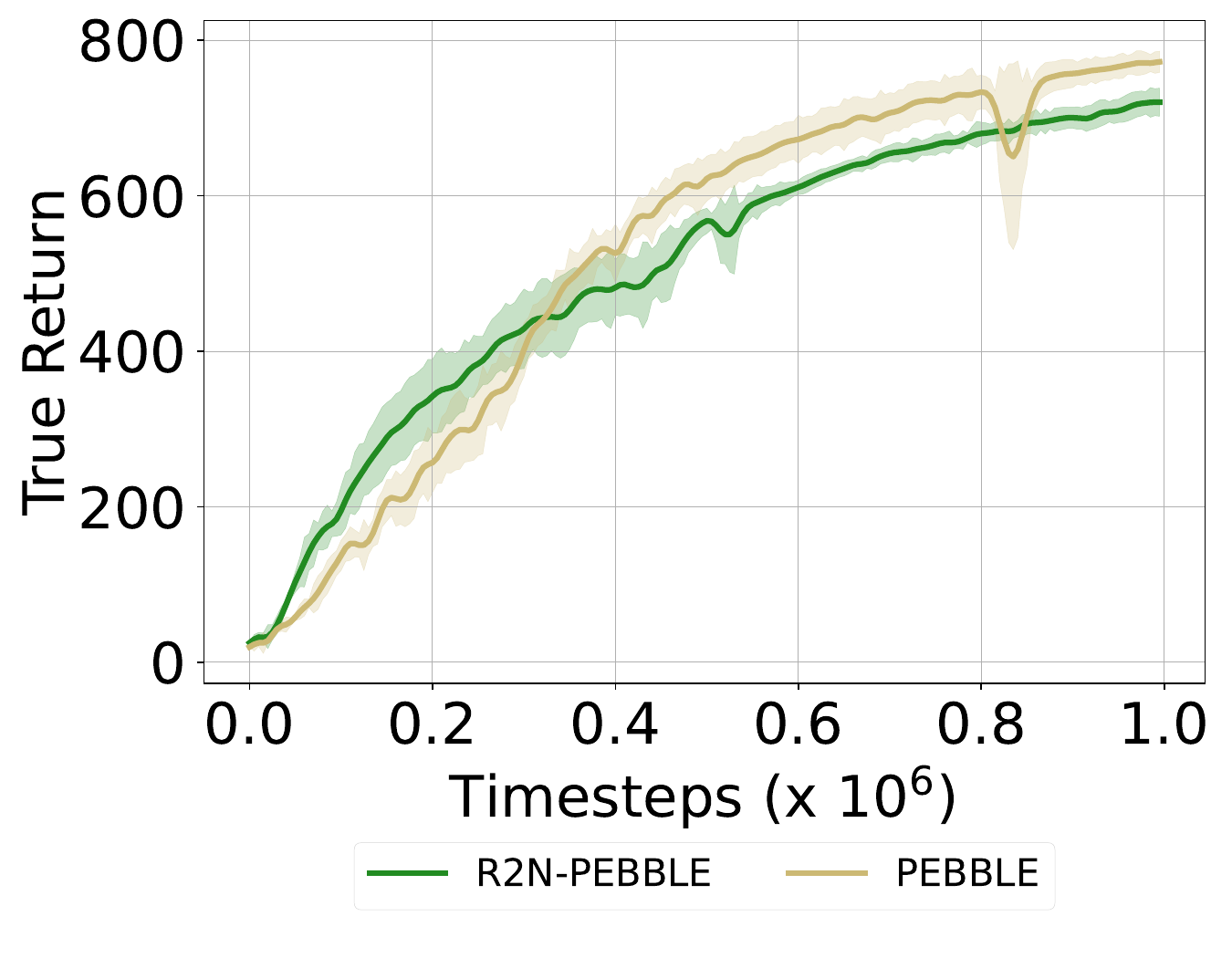}\label{fig:cheetah_run_0noise}}
  \hfill   
  \centering
 \subfloat[Walker-walk, Feedback=4000]{\includegraphics[width=0.5\textwidth]{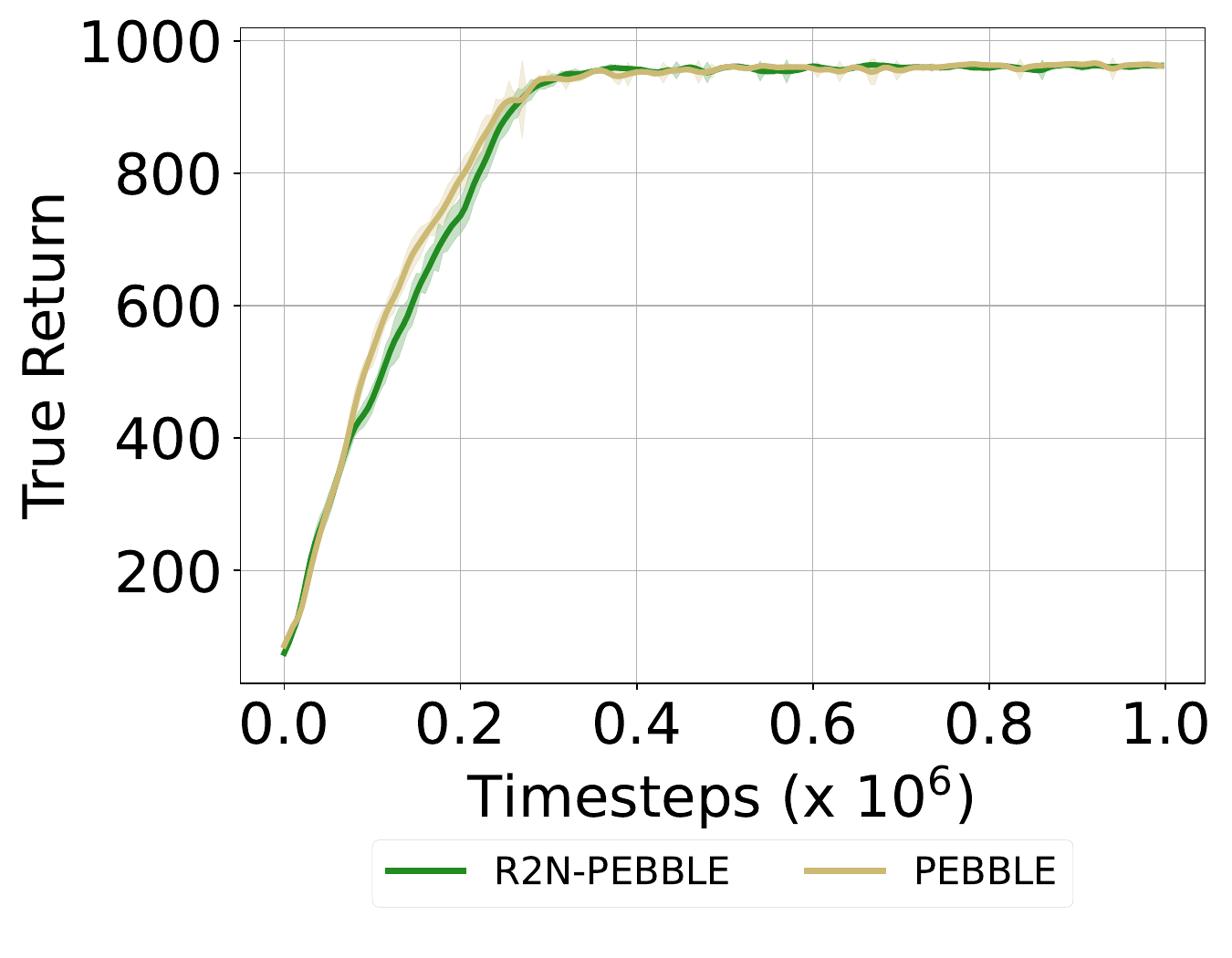}\label{fig:walker_walk_0noise}}
  \hfill   
  \centering
 \subfloat[Humanoid-stand, Feedback=10000]{\includegraphics[width=0.5\textwidth]{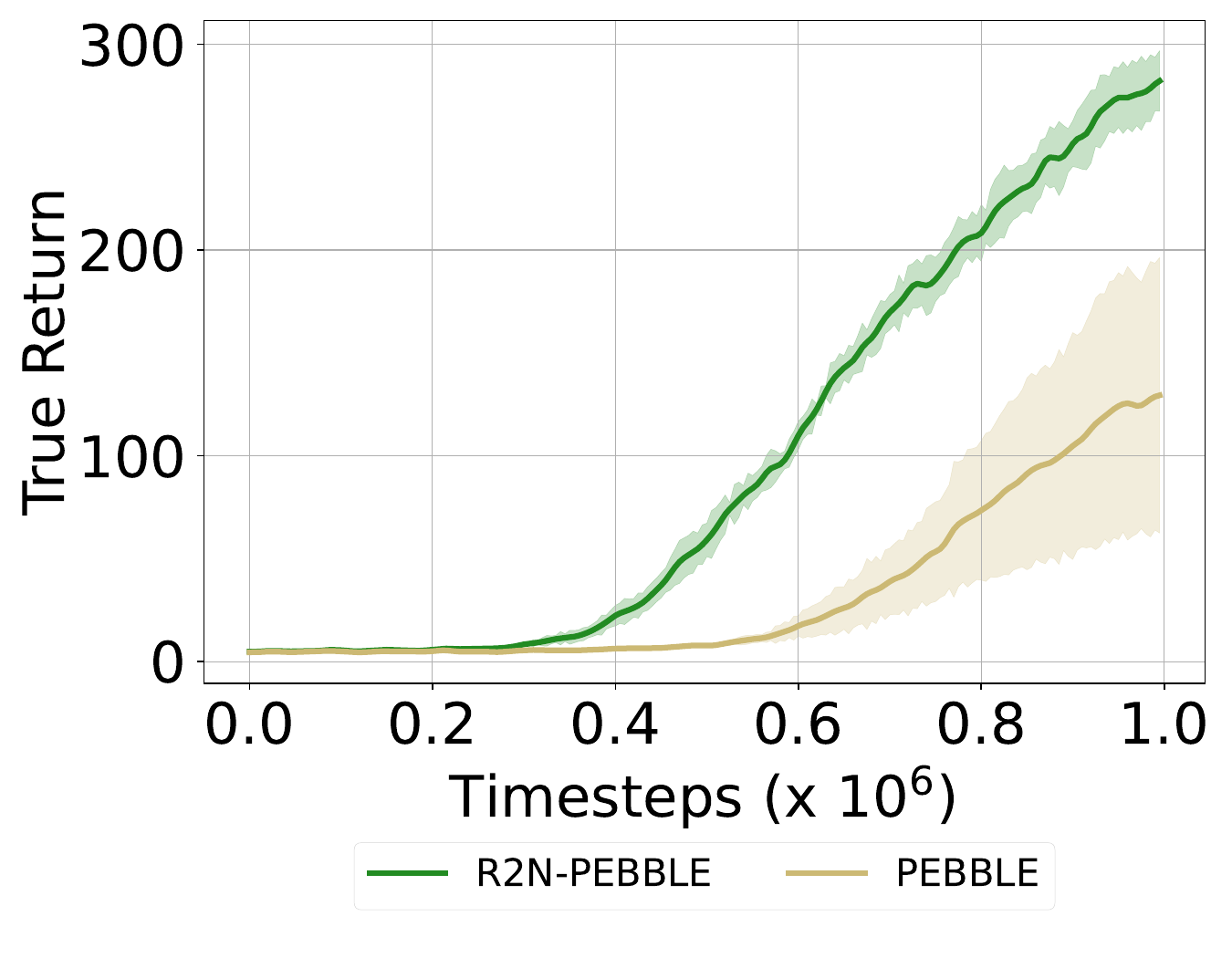}\label{fig:humanoid_stand_0noise}}
  \par   
  \captionsetup{skip=-0.2em}
  \caption{These plots compare the performance of R2N-PEBBLE and PEBBLE in traditional RL environments (i.e., $0\%$ noise setting).}\label{fig:0noise}
\end{figure*}

In this section, we show additional results comparing R2N-PEBBLE (green curve) with PEBBLE (yellow curve) in traditional RL environments without any added noise features. Overall in Figure \ref{fig:0noise}, we find that R2N-PEBBLE can achieve performance gains in some environments (Humanoid-stand), however, in others, it performs comparable or slightly worse. In this set of experiments, we kept the feedback budgets the same as the experiments with high added noise. Therefore, it might be the case that in $0\%$ noise environments, R2N-PEBBLE can achieve greater performance gains in lower feedback regimes. 

\subsection{Tests for Statistical Significance: R2N versus Sparse Training Baselines}\label{sec:table_results_rq1}
This section provides a full summary of all experiments under RQ 1. 
Tables \ref{tab:AUC_comparing_baselines} and \ref{tab:final_return_comparing_baselines} show the average AUC and final return for R2N and the four sparse training baselines in the tested DMControl experiments.
We also include the outcome of a Welch's $t$-test (equal variances not assumed). We use a p-value significance threshold of $0.05$. 
\begin{table*}[]
\caption{This table shows the average area under the curve (AUC) +/- standard error across all methods and environments. * indicates a significant difference between R2N and the other sparse training baselines.}
\label{tab:AUC_comparing_baselines}
\begin{tabular}{@{}lllll@{}}
\toprule
Task             & Feedback / Noise & Method             & AUC                   & P Value \\ \midrule
                 &                  & R2N-PEBBLE         & 117315.62 +/- 2335.79 &         \\
                 &                  & Static-R2N-PEBBLE  & 116461.45 +/- 2334.99 & 0.411   \\
Cartpole-swingup & 400 / 0.9        & SET-R2N-PEBBLE     & 112016.81 +/- 6987.0  & 0.269   \\
                 &                  & DropConnect-PEBBLE & 117025.07 +/- 2073.38 & 0.4678  \\
                 &                  & L1Reg-PEBBLE       & 110815.92 +/- 4545.69 & 0.144   \\ \midrule
                 &                  & R2N-PEBBLE         & 59633.26 +/- 6314.67  &         \\
                 &                  & Static-R2N-PEBBLE  & 54004.32 +/- 4094.93  & 0.261   \\
Walker-walk      & 4000 / 0.9       & SET-R2N-PEBBLE     & 47873.41 +/- 1013.76  & 0.069   \\
                 &                  & DropConnect-PEBBLE & 38794.09 +/- 3707.38  & 0.017*  \\
                 &                  & L1Reg-PEBBLE       & 43978.68 +/- 2085.58  & 0.034*  \\ \midrule
                 &                  & R2N-PEBBLE         & 45055.91 +/- 1383.87  &         \\
                 &                  & Static-R2N-PEBBLE  & 39629.12 +/- 1361.03  & 0.018*  \\
Cheetah-run      & 1000 / 0.9       & SET-R2N-PEBBLE     & 41051.56 +/- 2109.16  & 0.096   \\
                 &                  & DropConnect-PEBBLE & 24896.96 +/- 2316.76  & 0.000*  \\
                 &                  & L1Reg-PEBBLE       & 26874.27 +/- 1625.5   & 0.000*  \\ \midrule
                 &                  & R2N-PEBBLE         & 41388.91 +/- 3133.91  &         \\
                 &                  & Static-R2N-PEBBLE  & 41030.21 +/- 1388.97  & 0.463   \\
Quadruped-walk   & 4000 / 0.7       & SET-R2N-PEBBLE     & 37897.23 +/- 2112.66  & 0.216   \\
                 &                  & DropConnect-PEBBLE & 29980.62 +/- 1000.13  & 0.007*  \\
                 &                  & L1Reg-PEBBLE       & 28905.57 +/- 2408.5   & 0.011*  \\ \midrule
                 &                  & R2N-PEBBLE         & 10231.46 +/- 1680.98  &         \\
                 &                  & Static-R2N-PEBBLE  & 4323.92 +/- 997.25    & 0.013*  \\
Humanoid-stand   & 10000 / 0.7      & SET-R2N-PEBBLE     & 2310.3 +/- 668.14     & 0.002*  \\
                 &                  & DropConnect-PEBBLE & 1109.44 +/- 69.78     & 0.000*  \\
                 &                  & L1Reg-PEBBLE       & 1170.72 +/- 56.38     & 0.000*  \\ \bottomrule
\end{tabular}
\end{table*}

\begin{table*}[]
 {%
\begin{tabular}{@{}lllll@{}}
\toprule
\textbf{Task}    & \textbf{Feedback / Noise} & \textbf{Method}    & \textbf{Final Performnce} & P Value \\ \midrule
Cartpole-swingup & 400 / 0.9                 & R2N-PEBBLE         & 729.95 +/- 14.26          &         \\
                 &                           & Static-R2N-PEBBLE  & 737.59 +/- 16.23          & 0.620   \\
                 &                           & SET-PEBBLE         & 728.63 +/- 20.24          & 0.481   \\
                 &                           & DropConnect-PEBBLE & 719.91 +/- 7.62           & 0.296   \\
                 &                           & L1Reg-PEBBLE       & 690.3 +/- 17.57           & 0.077   \\ \midrule
Walker-walk      & 4000 / 0.9                & R2N-PEBBLE         & 434.85 +/- 48.74          &         \\
                 &                           & Static-R2N-PEBBLE  & 367.56 +/- 30.18          & 0.162   \\
                 &                           & SET-PEBBLE         & 329.17 +/- 11.53          & 0.047*  \\
                 &                           & DropConnect-PEBBLE & 252.73 +/- 34.83          & 0.013*  \\
                 &                           & L1Reg-PEBBLE       & 340.34 +/- 33.39          & 0.095   \\ \midrule
Cheetah-run      & 1000 / 0.9                & R2N-PEBBLE         & 384.8 +/- 12.53           &         \\
                 &                           & Static-R2N-PEBBLE  & 377.23 +/- 13.53          & 0.361   \\
                 &                           & SET-PEBBLE         & 404.03 +/- 13.36          & 0.812   \\
                 &                           & DropConnect-PEBBLE & 219.24 +/- 19.39          & 0.000*  \\
                 &                           & L1Reg-PEBBLE       & 239.53 +/- 14.52          & 0.000*  \\ \midrule
Quadruped-walk   & 4000 / 0.7                & R2N-PEBBLE         & 355.62 +/- 73.09          &         \\
                 &                           & Static-R2N-PEBBLE  & 283.55 +/- 20.56          & 0.210   \\
                 &                           & SET-PEBBLE         & 245.51 +/- 12.33          & 0.110   \\
                 &                           & DropConnect-PEBBLE & 207.51 +/- 14.16          & 0.056   \\
                 &                           & L1Reg-PEBBLE       & 218.3 +/- 34.69           & 0.083   \\ \midrule
Humanoid-stand   & 10000 / 0.7               & R2N-PEBBLE         & 154.63 +/- 20.46          &         \\
                 &                           & Static-R2N-PEBBLE  & 81.42 +/- 19.54           & 0.024*  \\
                 &                           & SET-PEBBLE         & 43.65 +/- 19.9            & 0.004*  \\
                 &                           & DropConnect-PEBBLE & 7.93 +/- 1.64             & 0.000*  \\
                 &                           & L1Reg-PEBBLE       & 8.34 +/- 1.22             & 0.000*  \\ \bottomrule
\end{tabular}%
}
\caption{This table shows the average area under the curve (AUC) +/- standard error across all methods and environments. * indicates a significant difference between R2N and the other sparse training baselines.}
\label{tab:final_return_comparing_baselines}
\end{table*}
\subsection{Tests for Statistical Significance for: Effectiveness of R2N across PbRL Algorithms}\label{sec:table_results_rq2}

This section provides a full summary of all experiments under RQ 2. Table~\ref{tab:final_return} and Table~\ref{tab:AUC} show the average final return and average area under the curve respectively over 14 seeds. We perform statistical tests between R2N and each baseline it is paired with, showing the outcome of Welch's $t$-test (equal variances not assumed). We use a p-value significance threshold of $0.05$. Note that in all 15 cases, R2N significantly increases both the final performance and the area under the curve (AUC) of its baseline.

\begin{table*}[h!]

\resizebox{\textwidth}{!}{%
\begin{tabular}{@{}llll@{}}
\toprule

  & \multicolumn{3}{c}{\textbf{\textsc{Percent Improvement of R2N}}}             \\ \midrule
\textbf{\textsc{Task / Feedback / Noise Fraction}} & \textsc{R2N-PEBBLE v. PEBBLE} & \textsc{R2N-RUNE v. RUNE} & \textsc{R2N-SURF v. SURF} \\ \midrule
\textsc{Cartpole-swingup / 400 / 0.90}                    & 8.540\%                         & 8.21\%             & 7.90\%             \\
\textsc{Walker-walk / 4000 / 0.90}                        & 33.60\%                         & 36.59\%            & 44.98\%            \\
\textsc{Cheetah-run / 1000 / 0.90}                        & 70.26\%                         & 103.06\%           & 55.74\%            \\
\textsc{Quadruped-walk / 4000 / 0.70}                     & 66.83\%                         & 39.67\%            & 39.55\%            \\
\textsc{Humanoid-stand / 10000 / 0.70}                    & 1942.97\%                       & 2165.27\%          & 1266.18\%          \\ \bottomrule
\end{tabular}%
}
\vspace{0.25cm}
\caption{This table shows the percent improvement in the average final return of the original PbRL algorithms once R2N is applied to it. This highlights that in all tested cases, R2N can boost the performance of the base PbRL algorithm.}
\label{tab:percent_increase}
\end{table*}

\begin{table*}[h]
\centering
\captionsetup{width=0.95\textwidth}
\caption{This table shows the average final return +/- standard error across all methods and environments. * indicates a significant difference between R2N and the original PbRL algorithm. }
\label{tab:final_return}
\begin{tabular}{@{}lllll@{}}
\toprule
\textbf{Task}    & \textbf{Feedback / Noise} & \textbf{Method}                                                                                 & \textbf{Final Return}                                                                                                                                      & \textbf{P Value}                                                    \\ \midrule
Cartpole-swingup & 400 / 0.9                 & \begin{tabular}[c]{@{}l@{}}\textbf{R2N-PEBBLE}\\ PEBBLE\\ \textbf{R2N-RUNE}\\ RUNE\\ \textbf{R2N-SURF}\\ SURF\end{tabular} & \begin{tabular}[c]{@{}l@{}}\textbf{716.89 +/- 9.75*}\\ 660.48 +/- 11.47\\ \textbf{731.78 +/- 8.69*}\\ 674.94 +/- 5.52\\ \textbf{734.73 +/- 10.83*}\\ 680.90 +/- 13.79\end{tabular}    & \begin{tabular}[c]{@{}l@{}}0.001\\ \\ 0.000\\ \\ 0.003\end{tabular}  \\ \midrule
Walker-walk      & 4000 / 0.9                & \begin{tabular}[c]{@{}l@{}}\textbf{R2N-PEBBLE}\\ PEBBLE\\ \textbf{R2N-RUNE}\\ RUNE\\ \textbf{R2N-SURF}\\ SURF\end{tabular} & \begin{tabular}[c]{@{}l@{}}\textbf{455.22 +/- 21.62*}\\ 340.72 +/- 16.29\\ \textbf{462.98 +/- 20.32*}\\ 338.94 +/- 24.78\\ \textbf{411.19 +/- 23.52*}\\ 283.60 +/- 11.33\end{tabular}  & \begin{tabular}[c]{@{}l@{}}0.000\\ \\ 0.000\\ \\ 0.000\end{tabular} \\ \midrule
Cheetah-run      & 1000 / 0.9                & \begin{tabular}[c]{@{}l@{}}\textbf{R2N-PEBBLE}\\ PEBBLE\\ \textbf{R2N-RUNE}\\ RUNE\\ \textbf{R2N-SURF}\\ SURF\end{tabular} & \begin{tabular}[c]{@{}l@{}}\textbf{404.59 +/- 17.12*}\\ 237.62 +/- 14.43\\ \textbf{410.94 +/- 19.18*}\\ 202.37 +/- 15.26\\ \textbf{405.68 +/- 18.38*}\\ 260.47 +/- 10.74\end{tabular} & \begin{tabular}[c]{@{}l@{}}0.000\\ \\ 0.000\\ \\ 0.000\end{tabular} \\ \midrule
Quadruped-walk   & 4000 / 0.7                & \begin{tabular}[c]{@{}l@{}}\textbf{R2N-PEBBLE}\\ PEBBLE\\ \textbf{R2N-RUNE}\\ RUNE\\ \textbf{R2N-SURF}\\ SURF\end{tabular} & \begin{tabular}[c]{@{}l@{}}\textbf{309.33 +/- 29.36*}\\ 185.41 +/- 8.88\\ \textbf{266.06 +/- 24.87*}\\ 190.48 +/- 9.30\\ \textbf{274.78 +/- 13.77*}\\ 196.90 +/- 24.05\end{tabular}  & \begin{tabular}[c]{@{}l@{}}0.000\\ \\ 0.005\\ \\ 0.006\end{tabular} \\ \midrule
Humanoid-stand   & 10000 / 0.7               & \begin{tabular}[c]{@{}l@{}}\textbf{R2N-PEBBLE}\\ PEBBLE\\ \textbf{R2N-RUNE}\\ RUNE\\ \textbf{R2N-SURF}\\ SURF\end{tabular} & \begin{tabular}[c]{@{}l@{}}\textbf{127.89 +/- 20.31*}\\ 6.26 +/- 0.18\\ \textbf{140.90 +/- 21.94*}\\ 6.22 +/- 0.19\\ \textbf{74.73 +/- 18.66*} \\ 5.47 +/- 0.19\end{tabular}                           & \begin{tabular}[c]{@{}l@{}}0.000\\ \\ 0.000\\\\ 0.000\end{tabular}            \\ \bottomrule
\end{tabular}%
\end{table*}

\begin{table*}[h!]
\centering
\captionsetup{width=\textwidth}
\caption{This table shows the average area under the curve (AUC) +/- standard error across all methods and environments. * indicates a significant difference between R2N and the original PbRL algorithm. }
\label{tab:AUC}
\resizebox{\textwidth}{!}{%
\begin{tabular}{@{}lllll@{}}
\toprule
\textbf{Task}    & \textbf{Feedback / Noise} & \textbf{Method}                                                                                 & \textbf{AUC}                                                                                                                                                                            & \textbf{P Value}                                                    \\ \midrule
Cartpole-swingup & 400 / 0.9                 & \begin{tabular}[c]{@{}l@{}}\textbf{R2N-PEBBLE}\\ PEBBLE\\ \textbf{R2N-RUNE}\\ RUNE\\ \textbf{R2N-SURF}\\ SURF\end{tabular} & \begin{tabular}[c]{@{}l@{}}\textbf{114986.66 +/- 2597.90*}\\ 104113.11 +/- 2612.11\\ \textbf{114836.05 +/- 2165.07*}\\  106055.52 +/- 1370.12\\ \textbf{117204.59 +/- 3351.06*}\\ 107394.51 +/- 2908.52\end{tabular} & \begin{tabular}[c]{@{}l@{}}0.004\\ \\ 0.001\\ \\ 0.021\end{tabular} \\ \midrule
Walker-walk      & 4000 / 0.9                & \begin{tabular}[c]{@{}l@{}}\textbf{R2N-PEBBLE}\\ PEBBLE\\ \textbf{R2N-RUNE}\\ RUNE\\ \textbf{R2N-SURF}\\ SURF\end{tabular} & \begin{tabular}[c]{@{}l@{}}\textbf{61814.89 +/- 2902.65*}\\ 45987.92 +/- 1716.72\\ \textbf{64458.89 +/- 2444.11*}\\ 45895.60 +/-  2579.91\\ \textbf{ 51750.05 +/- 2031.67*}\\ 41111.81 +/- 1387.92\end{tabular}      & \begin{tabular}[c]{@{}l@{}}0.000\\ \\ 0.000\\ \\ 0.000\end{tabular} \\ \midrule
Cheetah-run      & 1000 / 0.9                & \begin{tabular}[c]{@{}l@{}}\textbf{R2N-PEBBLE}\\ PEBBLE\\ \textbf{R2N-RUNE}\\ RUNE\\ \textbf{R2N-SURF}\\ SURF\end{tabular} & \begin{tabular}[c]{@{}l@{}}\textbf{43970.86 +/- 1898.99*}\\ 27585.24 +/- 1743.22\\ \textbf{42316.32 +/- 1626.45*}\\22849.97 +/- 894.95\\ \textbf{45172.59 +/- 2385.34*}\\ 28496.79 +/- 1338.12\end{tabular}       & \begin{tabular}[c]{@{}l@{}}0.000\\ \\ 0.000\\ \\ 0.000\end{tabular} \\ \midrule
Quadruped-walk   & 4000 / 0.7                & \begin{tabular}[c]{@{}l@{}}\textbf{R2N-PEBBLE}\\ PEBBLE\\ \textbf{R2N-RUNE}\\ RUNE\\ \textbf{R2N-SURF}\\ SURF\end{tabular} & \begin{tabular}[c]{@{}l@{}}\textbf{39639.90 +/- 1498.20*}\\ 29227.14 +/- 1320.89\\ \textbf{36803.72 +/- 1287.16*}\\ 29744.02 +/- 1331.92\\ \textbf{37356.69 +/- 1390.05*}\\ 28689.95 +/- 1460.15\end{tabular}        & \begin{tabular}[c]{@{}l@{}}0.000\\ \\ 0.001\\ \\ 0.000\end{tabular} \\ \midrule
Humanoid-stand   & 10000 / 0.7               & \begin{tabular}[c]{@{}l@{}}\textbf{R2N-PEBBLE}\\ PEBBLE\\ \textbf{R2N-RUNE}\\ RUNE\\ \textbf{R2N-SURF}\\ SURF\end{tabular} & \begin{tabular}[c]{@{}l@{}}\textbf{8508.52 +/- 1357.08*}\\ 1058.61 +/- 13.09\\ \textbf{9112.00 +/- 1421.13*}\\ 1059.46 +/- 13.33\\ \textbf{4987.07 +/- 1251.07*}\\ 985.91 +/- 17.57\end{tabular}                                     & \begin{tabular}[c]{@{}l@{}}0.000\\ \\ 0.000\\\\ 0.002\end{tabular}            \\ \bottomrule
\end{tabular}%
}
\end{table*}

\subsection{Tests for Statistical Significance for Sensitivity and Ablation Studies }\label{sec:ablation_tables}

This section provides a full summary of all experiments presented in Section \ref{sec:ablations}. 
Tables~\ref{tab:noise_ablation_table}--\ref{tab:imitating_noise_table} show the average final return and average area under the curve respectively over five seeds for the various sensitivity and ablation studies. 
We perform statistical tests between R2N and each baseline it is paired with, showing the outcome of Welch's $t$-test (equal variances not assumed). We use a p-value significance threshold of $0.05$. 

\begin{table*}[h]
\caption{This plot shows the average area under the curve (AUC) and final return +/- standard error for the noise ablation in Cheetah-run. * indicates a significant difference between R2N-PEBBLE and PEBBLE.}
\label{tab:noise_ablation_table}
\resizebox{\textwidth}{!}{%
\begin{tabular}{llllll}
\hline
\textbf{Noise Level} & \textbf{Method}                                             & \textbf{AUC}                                                                           & \textbf{P Value} & \textbf{Final Return}                                                        & \textbf{P Value} \\ \hline
0                    & \begin{tabular}[c]{@{}l@{}}R2N-PEBBLE\\ PEBBLE\end{tabular} & \begin{tabular}[c]{@{}l@{}}100172.87 +/- 3499.24\\ 103557.90 +/-  3922.01\end{tabular} & 0.710            & \begin{tabular}[c]{@{}l@{}}709.93 +/- 15.80\\ 765.43 +/-  14.62\end{tabular} & 0.975            \\ \hline
20                   & \begin{tabular}[c]{@{}l@{}}R2N-PEBBLE\\ PEBBLE\end{tabular} & \begin{tabular}[c]{@{}l@{}}107612.78 +/- 5146.40\\ 99416.24 +/- 7195.32\end{tabular}   & 0.216            & \begin{tabular}[c]{@{}l@{}}757.15 +/- 28.88\\ 736.78 +/- 36.13\end{tabular}  & 0.352            \\ \hline
50                   & \begin{tabular}[c]{@{}l@{}}\textbf{R2N-PEBBLE}\\ PEBBLE\end{tabular} & \begin{tabular}[c]{@{}l@{}}\textbf{102043.73 +/- 4298.56*}\\ 82705.98 +/-  4826.48\end{tabular} & 0.014             & \begin{tabular}[c]{@{}l@{}}737.09 +/- 21.96\\ 678.45 +/- 31.12\end{tabular}  & 0.103            \\ \hline
70                   & \begin{tabular}[c]{@{}l@{}}\textbf{R2N-PEBBLE}\\ PEBBLE\end{tabular} & \begin{tabular}[c]{@{}l@{}}\textbf{69993.54 +/- 2737.56*}\\ 55111.51 +/- 3559.99\end{tabular}   & 0.009            & \begin{tabular}[c]{@{}l@{}}\textbf{598.03 +/- 28.87*}\\ 473.89 +/- 20.99\end{tabular} & 0.007            \\ \hline
90                   & \begin{tabular}[c]{@{}l@{}}\textbf{R2N-PEBBLE}\\ PEBBLE\end{tabular} & \begin{tabular}[c]{@{}l@{}}\textbf{45055.91 +/- 1383.87*}\\ 26506.93 +/- 1444.38\end{tabular}    & 0.000            & \begin{tabular}[c]{@{}l@{}}\textbf{384.80 +/- 12.53*}\\ 238.24 +/- 23.91\end{tabular}  & 0.001            \\ \hline
95                   & \begin{tabular}[c]{@{}l@{}}\textbf{R2N-PEBBLE}\\ PEBBLE\end{tabular} & \begin{tabular}[c]{@{}l@{}}\textbf{27014.93 +/- 491.60}\\ 15493.66 +/- 2258.69\end{tabular}     & 0.001            & \begin{tabular}[c]{@{}l@{}}\textbf{279.80 +/- 18.30*}\\ 109.26 +/- 16.56\end{tabular}  & 0.000            \\ \hline
\end{tabular}%
}
\end{table*}

\begin{table*}[h]
\caption{This table shows the average area under the curve (AUC) and final return +/- standard error for the feedback study in Cheetah-run. * indicates a significant difference between R2N-PEBBLE and PEBBLE.}
\label{tab:fb_ablation_table}
\resizebox{\textwidth}{!}{%
\begin{tabular}{llllll}
\hline
\textbf{Feedback Amount} & \textbf{Method}                                             & \textbf{AUC}                                                                         & \textbf{P Value} & \textbf{Final Return}                                                        & \textbf{P Value} \\ \hline
100                      & \begin{tabular}[c]{@{}l@{}}\textbf{R2N-PEBBLE}\\ PEBBLE\end{tabular} & \begin{tabular}[c]{@{}l@{}}\textbf{37114.41 +/- 3084.44*}\\ 19610.52 +/- 2607.00\end{tabular} & 0.002            & \begin{tabular}[c]{@{}l@{}}\textbf{372.74 +/- 30.01*}\\ 176.43 +/- 28.41\end{tabular} & 0.001            \\ \hline
200                      & \begin{tabular}[c]{@{}l@{}}\textbf{R2N-PEBBLE}\\ PEBBLE\end{tabular} & \begin{tabular}[c]{@{}l@{}}\textbf{47562.73 +/- 2197.19*}\\ 23088.88 +/- 2265.25\end{tabular} & 0.000            & \begin{tabular}[c]{@{}l@{}}\textbf{498.50 +/- 18.10*}\\ 186.83 +/- 19.61\end{tabular} & 0.000            \\ \hline
400                      & \begin{tabular}[c]{@{}l@{}}\textbf{R2N-PEBBLE}\\ PEBBLE\end{tabular} & \begin{tabular}[c]{@{}l@{}}\textbf{41644.86 +/- 3360.05*}\\ 27966.80 +/- 2221.47\end{tabular} & 0.008            & \begin{tabular}[c]{@{}l@{}}\textbf{380.38 +/- 29.99*}\\ 227.58 +/- 15.53\end{tabular} & 0.001            \\ \hline
1000                     & \begin{tabular}[c]{@{}l@{}}\textbf{R2N-PEBBLE}\\ PEBBLE\end{tabular} & \begin{tabular}[c]{@{}l@{}}\textbf{45055.91 +/- 1383.87*}\\ 26506.94 +/- 1444.38\end{tabular} & 0.000            & \begin{tabular}[c]{@{}l@{}}\textbf{384.80 +/- 12.53*}\\ 238.24 +/- 23.91\end{tabular} & 0.001            \\ \hline
2000                     & \begin{tabular}[c]{@{}l@{}}\textbf{R2N-PEBBLE}\\ PEBBLE\end{tabular} & \begin{tabular}[c]{@{}l@{}}\textbf{45139.38 +/- 1645.85*}\\ 32659.90 +/- 3447.91\end{tabular} & 0.009            & \begin{tabular}[c]{@{}l@{}}\textbf{414.54 +/- 18.11*}\\ 266.72 +/- 23.28\end{tabular} & 0.001            \\ \hline
4000                     & \begin{tabular}[c]{@{}l@{}}\textbf{R2N-PEBBLE}\\ PEBBLE\end{tabular} & \begin{tabular}[c]{@{}l@{}}\textbf{46260.24 +/- 2147.41*}\\ 28900.51 +/- 1154.58\end{tabular} & 0.000            & \begin{tabular}[c]{@{}l@{}}\textbf{433.98 +/- 26.61*}\\ 263.39 +/- 9.58\end{tabular}  & 0.000            \\ \hline
10000                    & \begin{tabular}[c]{@{}l@{}}\textbf{R2N-PEBBLE}\\ PEBBLE\end{tabular} & \begin{tabular}[c]{@{}l@{}}\textbf{46704.99 +/- 1681.03*}\\ 31946.62 +/- 2314.48\end{tabular} & 0.001            & \begin{tabular}[c]{@{}l@{}}\textbf{415.72 +/- 16.04*}\\ 276.67 +/- 22.17\end{tabular} & 0.001            \\ \hline
\end{tabular}%
}
\end{table*}








\begin{table*}[h]
\caption{This table shows the average area under the curve (AUC) and final return +/- standard error for the DST ablation in Cheetah-run. ** indicates a significant difference between R2N-PEBBLE and both R2N-Actor/Critic Only and R2N-Reward Model Only .}
\label{tab:dst_ablation_table}
\resizebox{\textwidth}{!}{%
\begin{tabular}{lllll}
\hline
\textbf{Method}       & \textbf{AUC}           & \textbf{P Value} & \textbf{Final Return} & \textbf{P Value} \\ \hline
\textbf{R2N-PEBBLE}            & \textbf{45055.91 +/- 1383.87**} &                  & \textbf{384.80 +/- 12.53**}    &                  \\ 
R2N-Actor/Critic Only & 33159.39 +/- 1813.11   & 0.001            & 327.33 +/- 17.82      & 0.022             \\ 
R2N-Reward Model Only & 28643.04 +/- 2679.96   & 0.001            & 265.27 +/- 22.51      & 0.002            \\ \hline
\end{tabular}%
}
\end{table*}

\begin{table*}[h]
\caption{This table shows the average area under the curve (AUC) and final return +/- standard error for the imitating noise study in Cheetah-run. * indicates a significant difference between R2N-PEBBLE and PEBBLE .}
\label{tab:imitating_noise_table}
\resizebox{\textwidth}{!}{%
\begin{tabular}{lllll}
\hline
\textbf{Method} & \textbf{AUC}           & \textbf{P Value} & \textbf{Final Return} & \textbf{P Value} \\ \hline
\textbf{R2N-PEBBLE imitated noise}     & \textbf{18081.23 +/- 1611.21*} &                  & \textbf{220.19 +/- 26.47*}     &                  \\
PEBBLE imitated noise         & 11356.92 +/- 660.51    & 0.004            & 93.92 +/- 50.70       & 0.001            \\ \hline
\end{tabular}%
}
\end{table*}


\end{document}